\RequirePackage{fix-cm}
\documentclass[twocolumn,runningheads]{svjour3}
\PassOptionsToPackage{colorlinks,citecolor=blue,urlcolor=blue,linkcolor=blue}{hyperref}

\smartqed
\journalname{International Journal on Document Analysis and Recognition (IJDAR)}

\usepackage{amsmath}
\usepackage{newtxtext,newtxmath}
\DeclareMathAlphabet{\mathcal}{OMS}{cmsy}{m}{n}

\usepackage{graphicx}
\usepackage{multirow}
\usepackage{appendix}
\usepackage{xcolor}
\usepackage{textcomp}

\makeatletter
\renewcommand{\email}[1]{#1}
\makeatother
\usepackage{marvosym}

\newlength{\Ijdarmarkwidth}
\newcommand{\Ijdarcorrmark}{\makebox[\Ijdarmarkwidth][l]{\normalfont\Letter}}
\newcommand{\Ijdaraffilmark}{\makebox[\Ijdarmarkwidth][l]{\textsuperscript{1}}}
\newcommand{\Ijdarindent}{\hspace*{\Ijdarmarkwidth}}
\usepackage{booktabs}
\usepackage{array}
\usepackage{placeins}
\usepackage{algorithm}
\usepackage{algorithmicx}
\usepackage{algpseudocode}
\usepackage{listings}
\usepackage{epstopdf}
\usepackage{float}
\usepackage{cuted}%

\makeatletter

\preCutedStrip{\vskip-0.75\baselineskip}%
\makeatother
\usepackage{subcaption}

\usepackage[numbers,sort&compress]{natbib}
\usepackage[english]{babel}
\usepackage{arabtex}
\usepackage{utf8}
\setcode{utf8}

\AtBeginDocument{\setcounter{secnumdepth}{3}}
\usepackage{etoolbox}
\pretocmd\document{\endgroup}{}{}
\newcommand{\artext}[1]{%
  {\fontsize{8pt}{11pt}\selectfont \raisebox{0pt}[0pt][0pt]{\RL{#1}}}%
}
\usepackage[autostyle, english = american]{csquotes}
\MakeOuterQuote{"}
\usepackage{silence}
\usepackage[babel=true]{microtype}
\makeatletter
\renewcommand\labelitemi{\normalfont$\m@th\bullet$}
\renewcommand\labelitemii{\normalfont$\m@th\circ$}
\makeatother

\newcommand*{\MNVL}{MobileNetV3-\allowbreak{}Large }

\makeatletter
\@ifundefined{orcidlogo}{}{%
  
}
\makeatother
\usepackage{xurl}
\usepackage{orcidlink}

\newcommand{\Ijdardoi}{https://doi.org/10.1007/s10032-026-xxxxx-x}
\makeatletter
\def\makeheadbox{{%
  \hbox to0pt{\vbox{%
    \kern2pt
    \hbox to\hsize{\footnotesize\normalfont\@journalname\hfil}%
    \nointerlineskip
    \kern2pt
    \hbox to\hsize{\footnotesize\normalfont\Ijdardoi\hfil}%
    \kern4pt
  }\hss}}}%
\makeatother

\makeatletter
\renewenvironment{abstract}{%
  \topsep\z@\partopsep\z@\parsep\z@\itemsep\z@
  \list{}{%
    \leftmargin\z@
    \rightmargin\z@
    \listparindent\z@
    \itemindent\z@
    \parsep\parskip
    \topsep\z@
  }%
  \item\relax
  {\bfseries\abstractname}\par\nobreak
  \vskip\smallskipamount
  \noindent\ignorespaces
}{%
  \endlist
  \addvspace{3mm}%
}%
\makeatother

\newlength{\Ijdarmetablock}
\makeatletter
\if@twocolumn
  \global\headerboxheight=\z@
\fi
\patchcmd{\@maketitle}{\noindent\ignorespaces\@author\vskip7.23pt}%
  {\noindent\ignorespaces\@author\vskip\Ijdarmetablock}{}{}%
\patchcmd{\@maketitle}{\noindent{\small\@date\if@twocolumn\vskip 7.2mm\else\vskip 5.2mm\fi}}%
  {\noindent{\small\@date\if@twocolumn\vskip\Ijdarmetablock\else\vskip 5.2mm\fi}}{}{}%

\patchcmd{\@maketitle}{\global\@minipagetrue
 \global\everypar{\global\@minipagefalse\global\everypar{}}}{\global\@minipagefalse}{}{}%

\patchcmd{\@maketitle}{\rightskip43mm plus 4cm minus 3mm}{\rightskip=\z@ plus 4cm\relax}{}{}%
\makeatother

\makeatletter
\def\ps@Ijdarfirstpage{%
  \let\@oddhead\@empty\let\@evenhead\@empty
  \def\@oddfoot{\small\itshape Article accepted for publication in IJDAR\hfil}%
  \def\@evenfoot{\small\itshape Article accepted for publication in IJDAR\hfil}%
}
\makeatother


\makeatletter
\renewcommand{\subsection}{%
  \@startsection{subsection}{2}{\z@}%
  {-21dd plus-8pt minus-4pt}{10.5dd}%
  {\normalfont\normalsize\bfseries\boldmath}}%

\renewcommand{\subsubsection}{%
  \@startsection{subsubsection}{3}{\z@}%
  {-13dd plus-8pt minus-4pt}{10.5dd}%
  {\normalfont\normalsize\bfseries\boldmath}}%
\makeatother

\begin{document}

\title{Different Strokes for Different Folks: Writer Identification for Historical Arabic Manuscripts%
\thanks{Hamza A. Abushahla, Ariel Justine N. Panopio, and Layth Al-Khairulla: These authors contributed equally to this work.}}

\titlerunning{Writer Identification for Historical Arabic Manuscripts}

\author{%
Hamza A. Abushahla$^{1}$\orcidlink{0009-0003-6247-6776}
\and
Ariel Justine N. Panopio$^{1}$\orcidlink{0009-0000-8168-6478}
\and
Layth Al-Khairulla$^{1}$\orcidlink{0009-0009-5273-5969}
\and
Mohamed I. AlHajri$^{1}$\orcidlink{0000-0003-1480-8054}}

\authorrunning{Abushahla et al.}

\institute{%
\Ijdarcorrmark Mohamed I. AlHajri \at
\Ijdarindent\email{mialhajri@aus.edu}
\and
\Ijdarindent Hamza A. Abushahla \at
\Ijdarindent\email{b00090279@aus.edu}
\and
\Ijdarindent Ariel Justine N. Panopio \at
\Ijdarindent\email{b00088568@aus.edu}
\and
\Ijdarindent Layth Al-Khairulla \at
\Ijdarindent\email{b00087225@aus.edu}\\[3pt]
\Ijdaraffilmark\parbox[t]{\dimexpr\hsize-\Ijdarmarkwidth\relax}{Department of Computer Science and Engineering, American University of Sharjah, Sharjah 26666, United Arab Emirates}}

\date{Received: 11 August 2025 / Revised: 20 February 2026 / Accepted: 27 April 2026}
\maketitle
\thispagestyle{Ijdarfirstpage}

\begin{strip}
\vspace*{-4\baselineskip}
\begin{abstract}

Handwritten Arabic manuscripts preserve the Arab world’s intellectual and cultural heritage, and writer identification supports provenance, authenticity verification, and historical analysis. Using the Muharaf dataset of historical Arabic manuscripts, we evaluate writer identification from individual line images and, to the best of our knowledge, provide the first baselines reported under both line-level and page-disjoint evaluation protocols. Since the dataset is only partially labeled for writer identification, we manually verified and expanded writer labels in the public portion from 6,858 (28.00\%) to 21,249 lines (86.75\%) out of 24,495 line images, correcting inconsistencies and removing non-handwritten text. After further filtering, we retained 18,987 lines (77.51\%). We propose a Convolutional Neural Network (CNN)-based model with attention mechanisms for closed-set writer identification, including rare two-writer lines modeled as composite writer-pair classes. We benchmark fourteen configurations and conduct ablations across different feature extractors and training regimes. To assess generalization to unseen pages, the page-disjoint protocol assigns all lines from each page to a single split. Under the line-level protocol, a fine-tuned DenseNet201 with attention achieves \textbf{99.05\%} Top-1 accuracy, \textbf{99.73\%} Top-5 accuracy, and \textbf{97.44\%} F1-score. Under the more challenging page-disjoint protocol, the best observed results are \textbf{78.61\%} Top-1 accuracy, \textbf{87.79\%} Top-5 accuracy, and \textbf{66.55\%} F1-score, thus quantifying the impact of page-level cues. By expanding the Muharaf dataset’s labeled subset and reporting both protocols, we provide a clearer benchmark and a practical resource for historians and linguists engaged with culturally and historically significant documents. The code and implementation details are available on \href{https://github.com/7abushahla/Muharaf-Writer-Identification}{GitHub}.

\end{abstract}
\end{strip}

\section{Introduction}
\label{sec:introduction}
The Arabic language has played a vital role in preserving the intellectual and cultural heritage of the Arab world for over a millennium. Handwritten manuscripts document various subjects, including science, literature, personal letters, and religious texts \citep{boyi2024evolution}. These manuscripts capture the evolution of the Arabic script and provide insights into historical and cultural contexts. Identifying the authors helps researchers trace the origins of these manuscripts, verify their authenticity, and understand their historical significance. This task also preserves the legacy of individual authors and supports a deeper historical analysis \citep{asi2017}.

Writer identification holds critical importance for historians, linguists, and manuscript collectors. For decades, historians have worked to attribute historical documents to their original writers, as knowing the writer’s identity helps determine the text’s era, typography, and provenance, which are often uncertain \cite{rehman2019writer}. Many manuscripts lack essential metadata, such as the writer’s name or creation date, making identification especially challenging \cite{chammas2020}. While some manuscripts include colophons, which are brief notes at the end providing the writer’s name or date, these are often damaged, lost, or deliberately removed, leaving the text disconnected from its origins \cite{chammas2024}. As a result, experts rely heavily on comparing the writing style to that of other known works by the same writer.

Furthermore, identifying writers in historical Arabic manuscripts presents unique challenges. The Arabic script is inherently cursive, connecting letters fluidly within words and creating stylistic variations that complicate analysis. Contextual letter shapes, diacritical marks, known as "\artext{حركات}" (harakat), and diverse regional writing styles further increase this complexity. Additionally, handwriting reflects the author’s era, region, and personal preferences, contributing to significant variability.

This identification process requires substantial time, effort, and expertise. Recent advances in deep learning (DL) can accelerate and simplify this task. DL models have shown strong performance in extracting handwriting features and supporting accurate writer identification \citep{ngo2021vlad}, although challenges remain, particularly in handling degraded texts and varied script origins.

DL methods rely on large, high-quality labeled datasets to learn robust representations. In the domain of Arabic writer identification, several datasets exist, each varying in focus, size, and availability. Muharaf \cite{saeed2024muharaf} is currently the largest publicly available Arabic dataset, offering a larger sample size and greater diversity compared to the widely used WAHD dataset \cite{abdelhaleem2017wahd}. It is also publicly accessible, unlike the Balamand \cite{chammas2020} and KHATT~\cite{mahmoud2014khatt} datasets. Released in June 2024, Muharaf has so far been only used for handwriting recognition tasks \citep{saeed2024muharaf, chan2024hatformer}, as the dataset is only partially labeled for writer identification.

{In this study, we introduce the first application of the Muharaf dataset to \emph{closed-set} line-level writer identification and, to the best of our knowledge, report the first benchmarks under both line-level and page-disjoint evaluation protocols.} Given a line image, the goal is to assign it to one of a fixed set of writers observed during training (i.e., discrimination among known writers rather than open-set identification of unseen writers). To accomplish this, we leverage an end-to-end convolutional neural network (CNN)-based DL system augmented with attention modules. {In addition to the standard line-level evaluation, we evaluate a stricter page-disjoint protocol that assigns all lines from each page to a single split to assess generalization to unseen pages and reduce page-level leakage.} Our contributions are summarized as follows:

\begin{itemize}

\item { We benchmark writer identification on the Muharaf dataset under both line-level and page-disjoint splits, and quantify the performance gap between the two protocols.}

\item {We substantially expand the Muharaf dataset’s publicly labeled subset by manually verifying and increasing the labeled lines from 6{,}858 (28.00\%) to 21{,}249 (86.75\%), improving its usability for supervised writer identification.}

\item We propose an end-to-end DL pipeline for \emph{closed-set} line-level writer identification in historical Arabic manuscripts, combining a CNN backbone with an encoding/aggregation module and attention mechanisms for robust feature learning.

\item We retain instances of rare two-writer line images by modeling each observed writer-pair as a composite class under single-label classification, and we note that some composite classes lack standalone samples from both constituent writers.

\item {We analyze various transfer learning strategies, showing that fine-tuning pre-trained feature extractors can match or surpass non-fine-tuned and from-scratch training while significantly reducing training time, and we study the optimal number of layers to unfreeze to identify an effective fine-tuning depth.}

\item We highlight the challenges and potential of leveraging partially annotated datasets, such as Muharaf, for writer identification, offering valuable insights for future research in writer identification and related domains.

\end{itemize}

The rest of this paper is organized as follows. Section~\ref{sec:related} presents the related works in this field. Section~\ref{sec:data} illustrates the dataset preparation methodology. Section~\ref{sec:architecture} introduces the proposed end-to-end system. Section~\ref{sec:results} lists and discusses the empirical results. Section~\ref{sec:limitations} explores the study's limitations and outlines future work, while Section~\ref{sec:conclusion} concludes the paper.

\section{Related Work}
\label{sec:related}

In this section, we focus on DL-based and end-to-end approaches to writer identification that learn feature representations directly from data, unlike traditional handcrafted methods. Within this area, CNNs have proven especially effective at capturing both low-level features, such as {handwriting} strokes and textures, and high-level stylistic cues. Architectures like ResNet \citep{Resnet} and VGG-Net \citep{VGG} have achieved state-of-the-art performance across various datasets \citep{cilia2020end, cilia2020experimental}. A more specialized advancement comes from Arandjelovic et al. \citep{arandjelovic2016netvlad}, who introduced an end-to-end CNN-based image recognition system centered around the NetVLAD layer, a trainable version of the VLAD\footnote{Vector of Locally Aggregated Descriptors} layer. NetVLAD provides an encoding pipeline that aggregates local descriptors into a global representation and can be integrated into any CNN-based architecture. This innovation has paved the way for many later works.

Recently, Chammas et al. \citep{chammas2024} proposed an end-to-end CNN system combining ResNet50 as a local feature extractor with a NetVLAD aggregation layer via the Deep-TEN framework \citep{zhang2017deep}, achieving 99.2\% accuracy on the Balamand dataset. Moreover, Srivastava et al. \citep{srivastava2021exploiting} enhanced CNNs through multi-scale fusion, spatial attention, and patch interaction. DeepWriter \citep{xing2016deepwriter}, a multi-stream deep CNN, leveraged multilingual datasets (e.g., English and Chinese), demonstrating improved generalization. Similarly, DeepWINet \citep{chahi2023effective} was evaluated across eight languages as well as mixed-language datasets (e.g., the CERUG-MIXED dataset \citep{he2017writer} of English and Chinese). Their model achieved an accuracy of 99.27\% on the IFN/ENIT Arabic dataset~\citep{pechwitz2002ifn} and 94.28\% on the CERUG-MIXED dataset, showcasing robustness in multilingual, mixed-script settings.

Attention mechanisms~\citep{vaswani2017attention}, which enable models to focus on relevant handwriting regions while filtering out noise, are being increasingly integrated into writer identification frameworks. For instance, on word-level benchmarks (e.g., IAM~\citep{IAM_dataset}, CVL~\citep{CVL_dataset}, and CERUG-EN~\citep{he2015junction}), Kumar et al.~\citep{kumar2024attention} propose a fragment-driven dual-stream CNN with an attention mechanism, Okawa~\citep{okawa2025multistage} introduces a multistage CNN with a deformable attention module, and Majithia et al.~\citep{majithia2025integrated} propose a hybrid convolutional--transformer encoder (CTE) that integrates VGG-style convolutional blocks with transformer layers to capture both local and longer-range handwriting dependencies. Moreover, A-VLAD \citep{ngo2021vlad} combines attention modules with NetVLAD for effective feature aggregation, enhancing performance on historical documents. Additionally, Koepf et al. \citep{koepf2022writer} used vision transformers (ViTs) \citep{ViT} for writer identification, reporting strong results on CVL and ICDAR2013~\citep{louloudis2013icdar}. In a related effort, Fatnassi et al. \citep{fatnassi2025st} introduced ST-WID, a self-supervised ViT-based framework tailored for Arabic writer identification, reporting accuracies of 99.63\% and 89.15\% on the IFN/ENIT~\citep{pechwitz2002ifn} and AHTID/MW~\citep{mezghani2012database} datasets, respectively.

Further innovations address settings beyond standard supervised identification. For instance, Briber and Chibani \citep{briber2024open} propose a lightweight CNN trained on text fragments and paired with a distance-based classifier, allowing the model to generalize to new writers without retraining and achieving competitive results on IFN/ENIT. Additionally, Yang et al. \citep{yang2024dt2f} introduce DT2F-TLNet, combining deep fuzzy-logic modeling with transfer learning to improve robustness across multilingual and Arabic-script datasets under uncertainty. Moreover, Maitra et al.~\citep{maitra2025decorrelation} explore decorrelation-based self-supervised representation learning with a ResNet50~\citep{Resnet} encoder and evaluate on Arabic and English benchmarks, including AHAWP~\citep{AHAWP_dataset}, IAM, and CVL. Finally, Khalaif et al.~\citep{khalaif2025enhanced} combine corner detection with a CNN backbone, and further extend the task to writer verification by testing whether a query sample matches a claimed writer, reporting results on KHATT~\citep{mahmoud2014khatt}, AHAWP, and IAM.

{In summary, prior work has evolved from strong CNN backbones and aggregation layers toward attention-augmented CNNs and hybrid CNN--transformer designs to better capture local handwriting strokes and longer-range style cues. However, most gains are demonstrated on relatively clean, well-established benchmarks, and it remains unclear how well these gains transfer to historical Arabic manuscripts, which are typically noisier and more difficult to evaluate consistently. Moreover, while self-supervised approaches reduce dependence on labels during representation learning, reliable closed-set evaluation still requires ground truth and protocol rigor, which are precisely the bottlenecks in historical settings. Motivated by these gaps, we study \emph{closed-set} writer identification on Muharaf~\citep{saeed2024muharaf} using an end-to-end attention-based architecture, and we evaluate it under two complementary protocols with extensive ablations.}

\section{Data Labeling and Preparation}
\label{sec:data}

The Muharaf dataset contains fully annotated and transcribed historical Arabic manuscripts at the text-line level. It comprises 1,644 pages (1,216 public and 428 restricted), spanning from the early 19\textsuperscript{th} to the early 21\textsuperscript{st} century, with a total of 36,311 text lines. However, only 24,495 text lines are publicly available. Each line-level PNG image was generated using line-warping software to ensure consistent horizontal alignment. The dataset includes extensive metadata, such as \texttt{writer} tags that identify the author or scribe of each page. However, not all public samples contain this writer metadata.

\begin{figure}[htbp]
    \centering
    \includegraphics[width=0.9\linewidth]{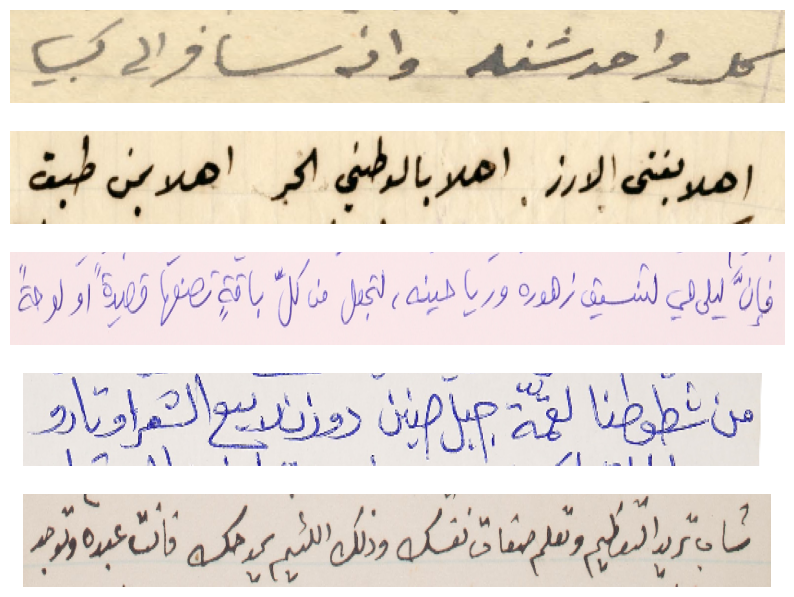}
    \caption{Sample line images from the Muharaf dataset.}
    \label{fig:line_examples}
\end{figure}

Figure~\ref{fig:line_examples} presents sample line-level images from the dataset, showcasing the wide stylistic and structural variation in handwriting across time periods and individuals. Table~\ref{tab:writer_distribution} summarizes the distribution of writer metadata within the public portion of the dataset.

\vspace{-20pt}
\begin{table}[htbp]
\caption{Writer Status in the Muharaf Dataset (Public)} 
\renewcommand{\arraystretch}{1.2} 
\label{tab:writer_distribution}
\centering
\footnotesize 
\setlength{\tabcolsep}{2pt} 
\begin{tabular*}{\columnwidth}{@{\extracolsep{\fill}}lllll} 
\toprule 
\textbf{Writer Status} & \textbf{No. Pages} & \textbf{\% Pages} & \textbf{No. Lines} & \textbf{\% Lines} \\
\midrule 
Labeled & 309 & 25.41\% & 6,858 & 28.0\% \\ 
Unlabeled & 907 & 74.59\% & 17,637 & 72.0\% \\ 
\textbf{Total} & \textbf{1,216} & \textbf{100\%} & \textbf{24,495} & \textbf{100\%} \\ 
\bottomrule 
\end{tabular*}
\end{table}

The dataset was originally annotated and transcribed by three of its creators, who are Arabic speakers with expertise in digital archives of Arabic manuscripts. The metadata does not include explicit information on writer gender, and, since gender cannot be reliably inferred from writer names, gender statistics are not reported.

\subsection{Manual Labeling}
\label{sec:manual-labeling}

We undertook a comprehensive manual labeling effort to address a significant portion of the unlabeled data. First, we organized metadata, linking each image to the corresponding writer’s name in Arabic and English. {Throughout this process, we assigned labels to the best of our ability based on explicit on-page attribution, such as names stated in the opening lines, headers, addresses, or signatures, supported by available document context.}

\begin{figure}[htbp]
\centering

\begin{subfigure}[b]{0.9\linewidth}
    \centering
    \includegraphics[width=\linewidth]{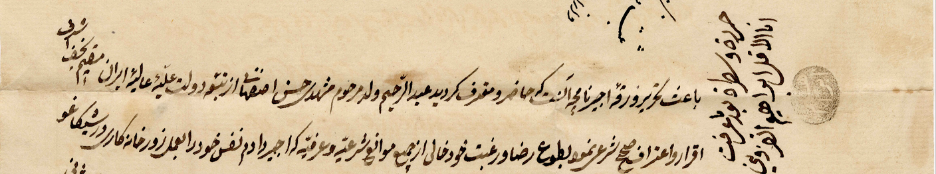}
    \caption{{Ottoman Turkish.}}
    \label{fig:ottoman-lines}
\end{subfigure}

\vspace{0.35cm}

\begin{subfigure}[b]{0.9\linewidth}
    \centering
    \includegraphics[width=\linewidth]{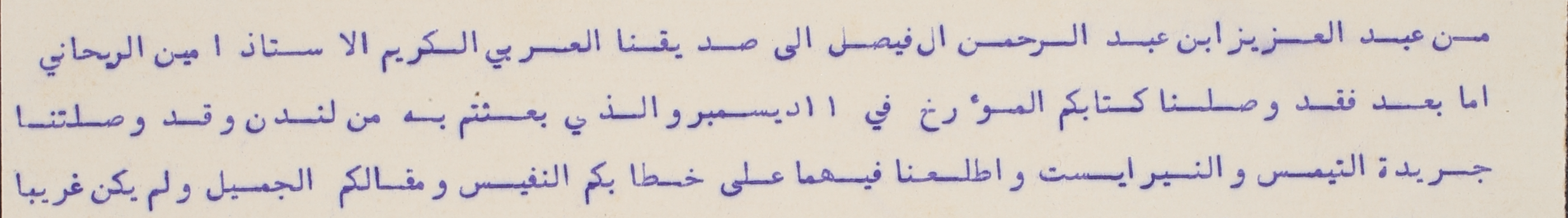}
    \caption{{Typewritten Arabic.}}
    \label{fig:typewritten-lines}
\end{subfigure}

\vspace{0.35cm}

\begin{subfigure}[b]{0.9\linewidth}
    \centering
    \includegraphics[width=\linewidth]{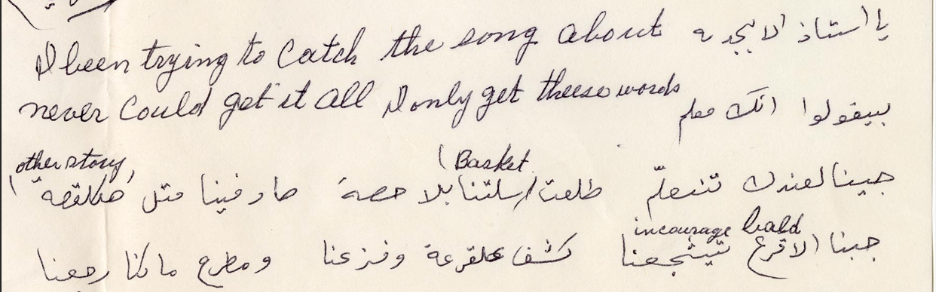}
    \caption{{Arabic and English on the same line.}}
    \label{fig:mixed-lines}
\end{subfigure}

\vspace{0.35cm}

\begin{subfigure}[b]{0.9\linewidth}
    \centering
    \includegraphics[width=\linewidth]{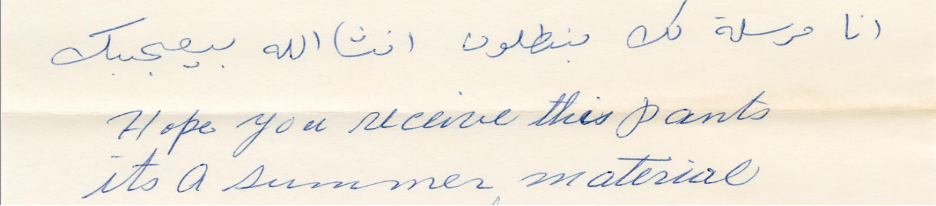}
    \caption{{Arabic and English on separate lines.}}
    \label{fig:arabic-english-lines}
\end{subfigure}

\caption{{Special cases observed in the dataset: (a) Ottoman Turkish receipt written in Perso-Arabic script, (b) typewritten Arabic letter, (c) Arabic and English on the same line, and (d) Arabic and English on separate lines.}}
\label{fig:special-cases}
\end{figure}

However, the dataset presented some notable challenges. Among its diverse contents were 21 Ottoman Turkish pages written in Arabic script, as seen in Figure~\ref{fig:ottoman-lines}, for which no writer could be identified---these were excluded from the analysis. Additionally, we encountered typewritten Arabic pages (see Figure~\ref{fig:typewritten-lines}), which were treated as non-handwritten content and removed during filtering. We also observed handwritten mixed-language samples where Arabic and English appear together, sometimes on the same line (see Figure~\ref{fig:mixed-lines}) and sometimes on separate lines (see Figure~\ref{fig:arabic-english-lines}). Despite the apparent differences between Arabic and English scripts, we retained these cross-script handwritten samples based on empirical evidence that suggests a person’s handwriting remains recognizable across scripts \citep{srihari2002individuality, djeddi2013text, bertolini2016multi}. While this decision may introduce additional complexity and potential challenges during inference, we believe it enriches the dataset and improves its overall representativeness.

For the predominantly Arabic material, we transliterated each writer’s name into English using reliable sources and cross-referenced them with the previously labeled portion of the dataset for consistency. In some cases, direct mapping was possible when writers signed their names or addresses in English. Where explicit attribution was missing, we used collection-level and historical context, together with coarse handwriting similarity, to support attribution where possible; nevertheless, not all such cases could be resolved, resulting in a portion of the dataset remaining unlabeled. In instances where relational identifiers were used, such as ``Your nephew'' or ``Your son,'' these were preserved without further disambiguation.

{Beyond ambiguous attribution, we also encountered rare pages in which two writers are explicitly indicated. Some letters include two signatures on the same page and were labeled using the observed writer pair as-is, such as \textit{Father Youssef Hanna \& Father Botros Hasan} and \textit{Mkhayel \& Ibrahim Daher}. In rarer cases, parts of the same letter or page were written by two distinct individuals; these were likewise retained and labeled as composite writer-pairs, such as \textit{Yousef Hobeiche \& Angele Ellis}, and treated as composite classes in our closed-set formulation.}

Historical and literary context also played an important role. For example, letters signed as ``May'' in the \textit{Amin Rihani} collection were attributed to \textit{May Ziadeh}, a well-known female Arab poet and author, based on historical correspondence between her and Rihani. Similarly, manuscripts in the \textit{Elias Abu Shabaki} collection were confirmed via online poetry archives. Moreover, scripts in the \textit{Salah Tizani} collection were labeled by matching character names from his TV and theater works.

In family collections, such as the Ellis family collection, we relied on familial relationships and document context to label letters and postcards, supported by visual handwriting cues and additional family tree research. {Finally, we note that, across the dataset, some pages may have been written by a scribe but signed by an author; since our labels follow the attribution present on the page, such cases may introduce intra-class variability.}

\subsection{Label Verification}

Following the manual labeling process, we undertook a careful verification process to ensure consistency. This included aligning Arabic-English transliterations between the newly labeled entries and those in the existing labeled portion. This was essential to avoid discrepancies or variations in spellings and formatting.

This was followed by a rigorous refinement process to detect potential duplicate writer names. We applied fuzzy string matching using the \textit{thefuzz}\footnote{\url{https://github.com/seatgeek/thefuzz}} library, which relies on Levenshtein distance \citep{levenshtein1966binary} to calculate string similarity based on the minimum number of single-character edits. Scores range from 0 (no similarity) to 100 (identical). We initially set a 90\% similarity threshold to flag likely duplicates, such as “Botros Hassan” vs. “Boutros Hassan” (98\%) and “Botros Hassan” vs. “Botros Hasan” (97\%). Thresholds were then dynamically adjusted between 85--95\% to capture less obvious or highly confident matches. Notably, all flagged pairs underwent careful manual review, during which we assessed context, handwriting evidence, and naming patterns to ensure the accurate consolidation of writer identities.

\subsection{Error Corrections}
\begin{figure}[htbp]
\centering

\begin{subfigure}[b]{0.9\linewidth}
    \centering
    \includegraphics[width=0.8\linewidth]{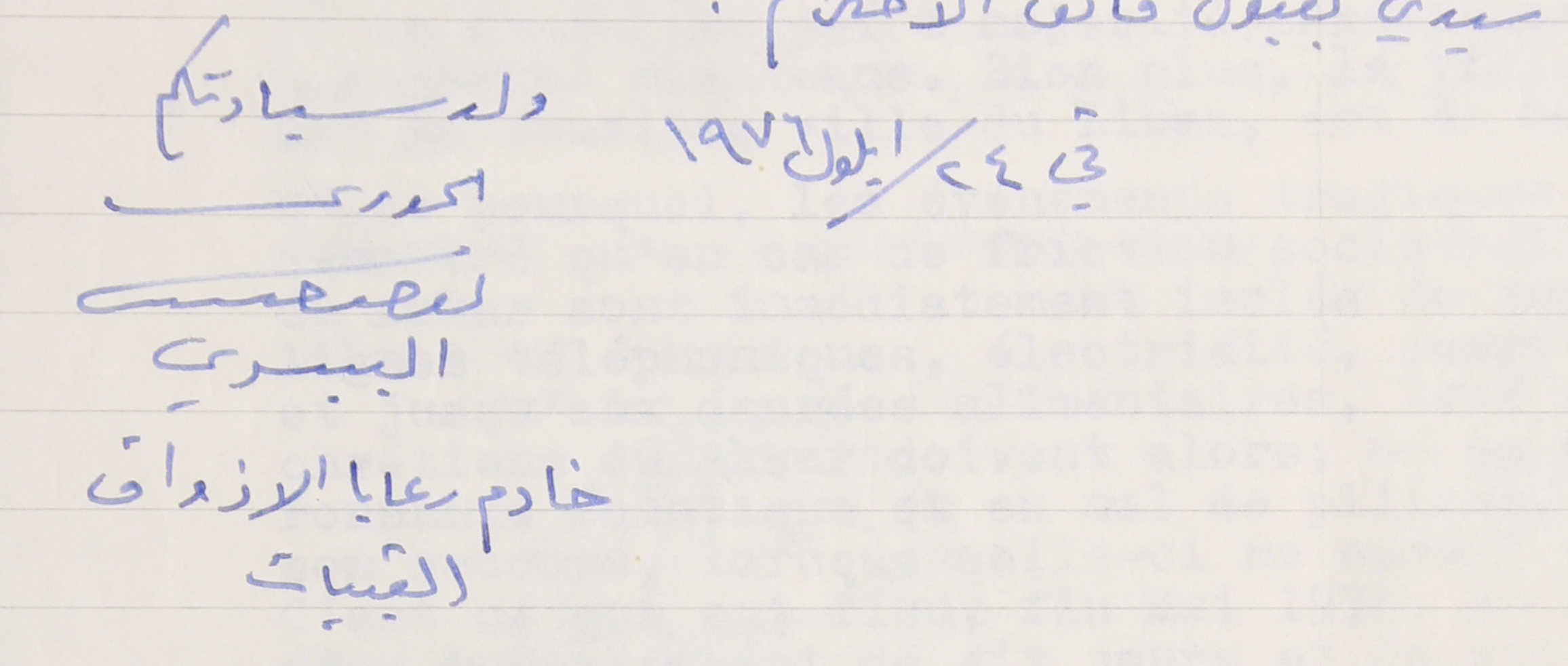}
    \caption{{Signature text in Arabic that reads: ``\artext{الخوري يوحنا حبيب البيسري خادم رعايا الأذواق، القبيات}''~(Al-Khouri Youhanna Habib Al-Baissary, servant of the parishes of Al-Azwaq, Al-Qoubaiyat).}}
    \label{fig:signature_mismatch_labeled}
\end{subfigure}


\begin{subfigure}[b]{0.9\linewidth}
    \centering
    \includegraphics[width=0.8\linewidth]{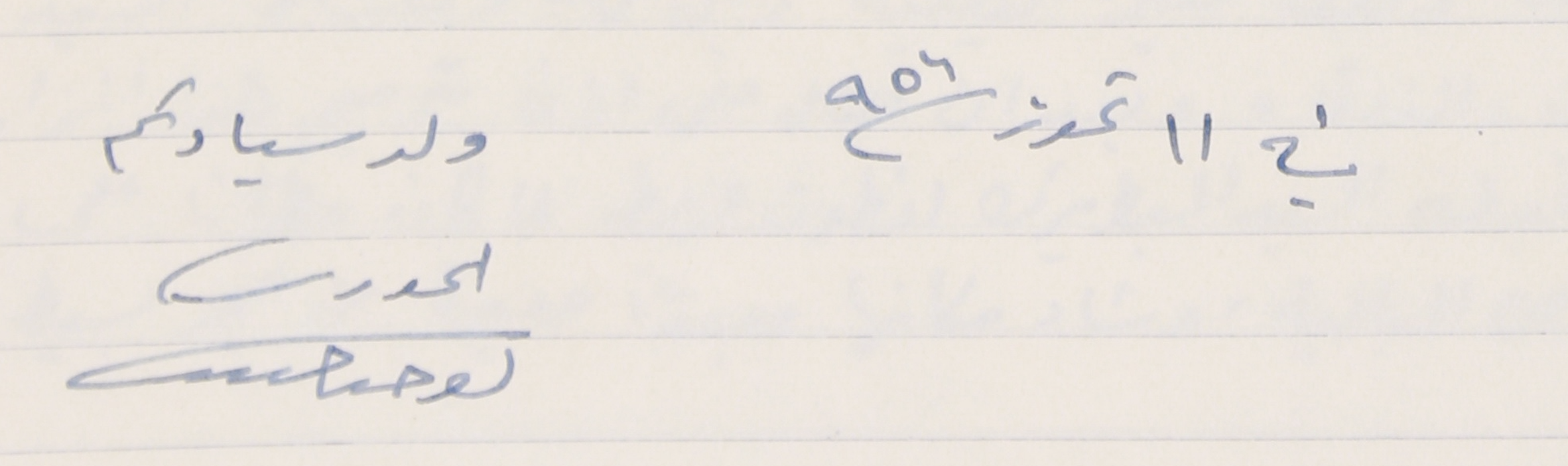}
    \caption{{Similar signature transcribed as ``Al-Khouri Youhanna Habib''.}}
    \label{fig:signature_similar_unlabeled}
\end{subfigure}

\vspace{0.35cm}

\begin{subfigure}[b]{0.9\linewidth}
    \centering
    \includegraphics[width=0.8\linewidth]{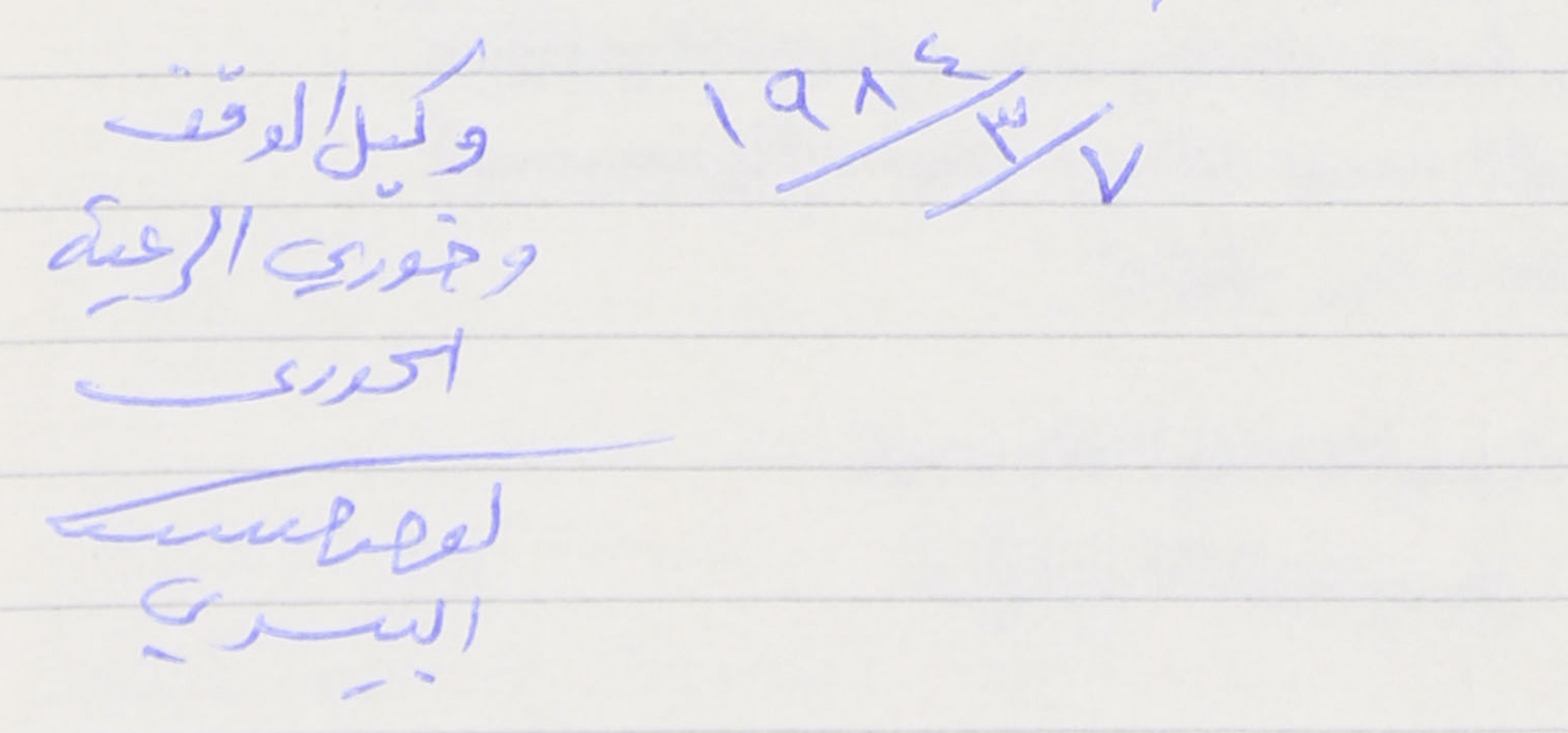}
    \caption{{Full name and title transcribed as ``Al-Khouri Youhanna Habib Al-Baissary''.}}
    \label{fig:signature_fullname_unlabeled}
\end{subfigure}

\caption{{Examples illustrating inconsistent or partial name transcriptions for the same writer across the labeled and unlabeled portions of the dataset: (a) labeled portion with a mismatched label (``Father Youssef Baissary''), (b) unlabeled portion with a partial transcription, and (c) unlabeled portion showing the full name and title.}}
\label{fig:signature_examples}
\end{figure}

While comparing the labeled and newly labeled portions, we identified several instances of mislabeling in the original dataset. One prominent example involved a page tagged as written by "Father Youssef Baissary," shown in Figure~\ref{fig:signature_mismatch_labeled}. The Arabic title \artext{الخوري} (Al-Khouri), meaning “Father,” refers specifically to a Christian priest. However, the original page transcription only recorded “\artext{الخوري البيسري}” (Al-Khouri Al-Baissary), omitting the handwritten text between the title and surname. When cross-referencing other pages, we found matching handwriting and signatures. For instance, Figure~\ref{fig:signature_similar_unlabeled} shows the name as “\artext{الخوري يوحنا حبيب}” (Al-Khouri Youhanna Habib). At the same time, Figure~\ref{fig:signature_fullname_unlabeled} records the full name and title “\artext{الخوري يوحنا حبيب البيسري}” (Al-Khouri Youhanna Habib Al-Baissary), confirming all these pages belonged to the same individual. The label was then corrected to "Father Youhanna Habib Baissary" to guarantee consistency with the initially labeled portion of the dataset.

\subsection{Dataset Preparation}
\label{sec:data-preparation}

Afterward, the dataset’s line-level images were mapped to the identified writers. Of the 24,495 public text lines in the dataset, 21,249 lines were successfully labeled, increasing the number of identified writers from 94 to 179. These labeled lines were then filtered to remove non-handwritten content, such as stamps, page numbers, and printed text, resulting in 18,987 usable lines for the dataset. A summary of the writer-label coverage before and after manual labeling is presented in Table~\ref{tab:before_filtering}, while Table~\ref{tab:filtered_training} illustrates the filtered dataset.

\begin{table}[htbp]
\centering
\caption{{Writer Status Before and After Manual Labeling}}
\label{tab:before_filtering}
\footnotesize
\setlength{\tabcolsep}{2pt}
\renewcommand{\arraystretch}{1.2}
{
\begin{tabular*}{\columnwidth}
{@{\extracolsep{\fill}}lllll}
\toprule
\textbf{Writer Status} & \textbf{No. Pages} & \textbf{\% Pages} & \textbf{No. Lines} & \textbf{\% Lines} \\
\midrule
\multicolumn{5}{c}{Before manual labeling} \\
\midrule
Labeled   & 309  & 25.41\% & 6,858  & 28.0\% \\
Unlabeled & 907  & 74.59\% & 17,637 & 72.0\% \\
\textbf{Total} & \textbf{1,216} & \textbf{100\%} & \textbf{24,495} & \textbf{100\%} \\
\midrule
\multicolumn{5}{c}{After manual labeling} \\
\midrule
Labeled   & 1,015 & 83.5\%  & 21,249 & 86.75\% \\
Unlabeled & 201   & 16.5\%  & 3,246  & 13.25\% \\
\textbf{Total} & \textbf{1,216} & \textbf{100\%} & \textbf{24,495} & \textbf{100\%} \\
\bottomrule
\end{tabular*}
}
\vspace{-0.3em}
\end{table}


\begin{table}[htbp]
\caption{Filtered Dataset After Manual Labeling and Excluding Non-Handwritten Content}
\renewcommand{\arraystretch}{1.2}
\label{tab:filtered_training}
\centering
\footnotesize
\setlength{\tabcolsep}{5pt}
\begin{tabular*}{\columnwidth}{@{\extracolsep{\fill}}ll}
\toprule
\textbf{Metric} & \textbf{Value} \\
\midrule
Total writers (classes) & 179 \\

{Total pages used} & {1,015} \\
{Total pages unused} & {201} \\
{\% of pages used (out of the original 1,216)} & {83.50\%} \\

{Maximum pages per writer} & {68} \\
{Minimum pages per writer} & {1} \\
{Mean pages per writer} & {5.67} \\
{Standard deviation (pages per writer)} & {10.012} \\

\midrule
Total lines used & 18,987 \\
Total lines unused & 2,262 \\
\% of lines used (out of the original 24,495) & 77.51\% \\
Maximum lines per writer & 949 \\
Minimum lines per writer & 10 \\
Mean lines per writer & 106.07 \\
Standard deviation (lines per writer) & 183.29 \\
\bottomrule
\end{tabular*}
\end{table}

\begin{figure*}[t]
\centering
\includegraphics[width=1.0\textwidth]{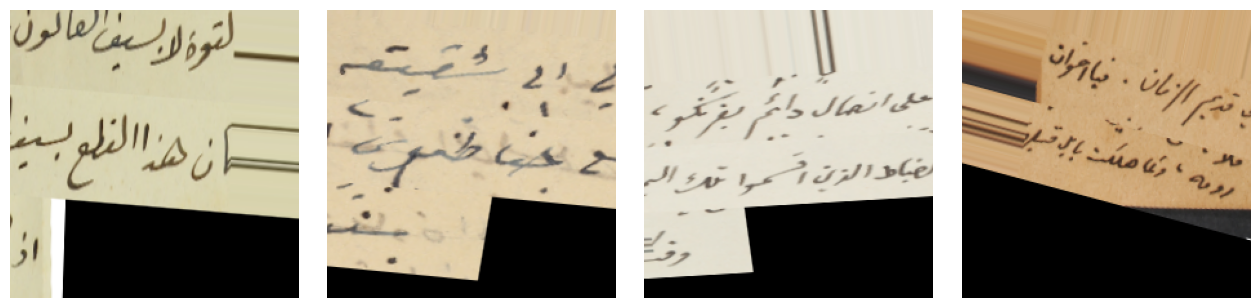}
\caption{Examples of training data augmentations: rotation, zoom, shear, shifts, and nearest fill mode.}
\label{fig:augmentations}
\end{figure*}

This manual labeling process significantly increased the usability of the Muharaf dataset for writer identification. However, the dataset remains highly imbalanced and exhibits an extreme long-tail class distribution, with a small number of writers contributing a disproportionately larger number of labeled samples compared to others. For example, the top three classes include \textit{Ameen Rihani} with 949 lines, \textit{Hanna Ghayth} with 934 lines, and \textit{Hanna Moussa} with 876 lines. Conversely, the lowest classes include \textit{Nehme Elias Mikhail}, \textit{Shibli Barakat Witnesses}, and \textit{Father Elias} with 12, 11, and 10 line images, respectively. A visualization of the highly skewed distribution of labeled samples is shown in Appendix~\ref{app:protocol-a-dist}.

\begin{figure}[htbp]
\centering
\includegraphics[width=0.75\linewidth]{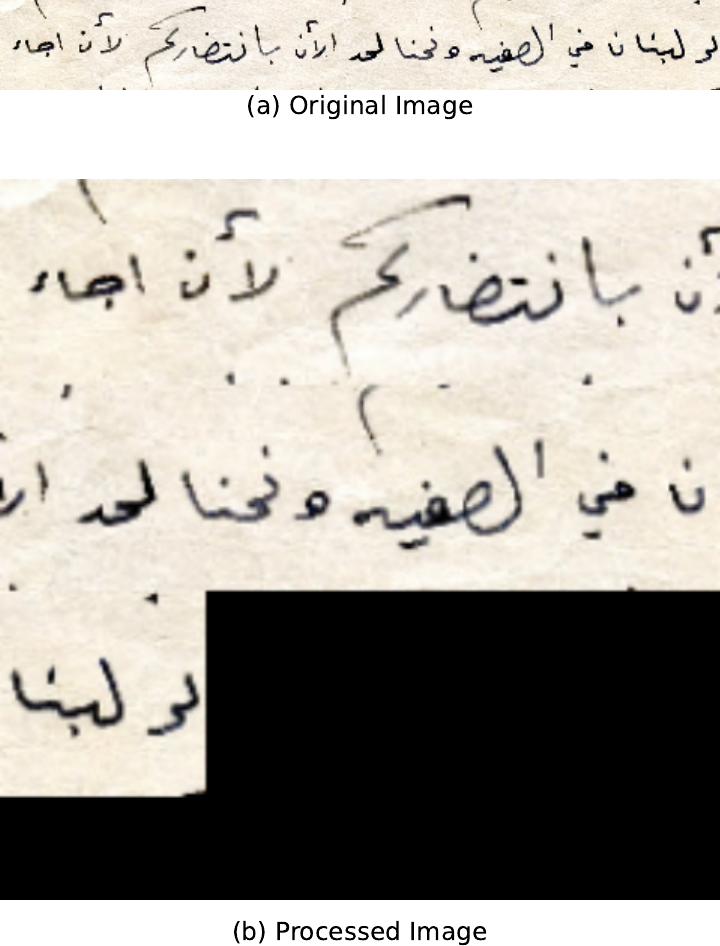}
\caption{BlockProcessor transformation: (a) Original image and (b) Processed image, resized and zero-padded, keeping the aspect ratio and avoiding information loss.}
\label{fig:blockprocessor}
\end{figure}

\vspace{-20pt}

\subsection{Data Preprocessing { and Evaluation Protocols}}
\label{data_preprocessing}

We adopted the BlockProcessor from \cite{chan2024hatformer} to preserve aspect ratios and avoid distortion common in standard resizing pipelines. As illustrated in Figure~\ref{fig:blockprocessor}, the BlockProcessor resizes each image proportionally and zero-pads it into a square block of \(224 \times 224\) pixels. On average, line images were 614 pixels wide after standardizing the heights to 64 pixels. Unlike \cite{saeed2024muharaf} and \cite{chan2024hatformer}, we did not flip images horizontally, as the end-of-line token placement was irrelevant to our approach. Moreover, we did not binarize the images, relying instead on the preprocessing functions of the pre-trained CNNs.

{
\textbf{Closed-set identification.}
We evaluate writer identification under a \emph{closed-set} protocol, formulated as a multi-class classification problem over a stratified fixed set of writer classes. Thus, the same writer identities appear in training, validation, and test splits, and we do not evaluate open-set identification of unseen writers (i.e., writers not present in training), which requires a different protocol and is left for future work.

\textbf{Protocol A: line-level random split.}
Using the filtered labeled set, we perform a 70--15--15 train--validation--test split at the \emph{line} level.
Individual line images are disjoint across splits, and the number of classes 179.
However, since multiple lines originate from the same manuscript page, this protocol may place different lines from the same page across training, validation, and test sets, potentially allowing page-specific cues to influence the measured performance.

\textbf{Protocol B: page-disjoint split.}
We additionally consider a more challenging \emph{page-disjoint} evaluation protocol in which each manuscript page is treated as an indivisible unit: all line images extracted from a given page appear in exactly one split. The manually labeled and filtered dataset contains writers with as few as one or two pages (see Table~\ref{tab:filtered_training}). To preserve the closed-set setting while enforcing a 70--15--15 split at the page level, we therefore restrict Protocol~B to writers with at least three distinct pages, ensuring that at least one page can be allocated to each of the training, validation, and test splits. After applying this constraint, the page-disjoint subset contains 71 writer classes and 16{,}456 line images; a summary comparison of both protocols is provided in Table~\ref{tab:protocol_summary}. A visualization of the per-writer page distribution for these 71 classes is provided in Appendix~\ref{app:protocol-b-dist}, highlighting that the dataset remains skewed.

\begin{table}[htbp]
\caption{{
Summary of Evaluation Protocols}}
\renewcommand{\arraystretch}{1.2}
\label{tab:protocol_summary}
\centering
\footnotesize
\setlength{\tabcolsep}{2pt}
{
\begin{tabular*}{\columnwidth}{@{\extracolsep{\fill}}llll}
\toprule
\textbf{Protocol} & \textbf{No. Writers} & \textbf{No. Pages} & \textbf{No. Lines} \\
\midrule
Line-level     & 179 & 1,015 & 18,987 \\
Page-disjoint  & 71  & 877 & 16,456 \\
\bottomrule
\end{tabular*}
}
\end{table}

\textbf{Data augmentations.}
}
To mitigate the dataset's severe class imbalance, we apply targeted data augmentation strategies to the training split only (see Figure~\ref{fig:augmentations} and Table~\ref{tab:augmentation_parameters}). For both protocols, the split is performed first to prevent \emph{augmentation-induced} leakage, ensuring that no augmented variants of validation or test images can appear during training. During evaluation, shuffling is disabled for validation and test loaders (training is shuffled to prevent order bias). Finally, writer labels are one-hot encoded to be compatible with categorical cross-entropy loss used during training.

\begin{table}[!htbp]
\caption{Data Augmentation Parameters for Training} 
\renewcommand{\arraystretch}{1.2} 
\label{tab:augmentation_parameters}
\centering
\footnotesize 
\setlength{\tabcolsep}{5pt} 
\begin{tabular*}{\columnwidth}{@{\extracolsep{\fill}}ll} 
\toprule 
\textbf{Augmentation Parameter} & \textbf{Value} \\
\midrule 
Rotation Range & ±15° \\ 
Zoom Range & ±30\% \\ 
Shear Range & ±30\% \\ 
Width Shift Range & ±20\% of image width \\ 
Height Shift Range & ±20\% of image height \\ 
Fill Mode & Nearest \\ 
\bottomrule 
\end{tabular*}
\end{table}

\vspace{-10pt}

\section{Proposed Architecture}
    \label{sec:architecture}

    \begin{figure*}[ht]
    \centering
    \includegraphics[width=1.0\textwidth]{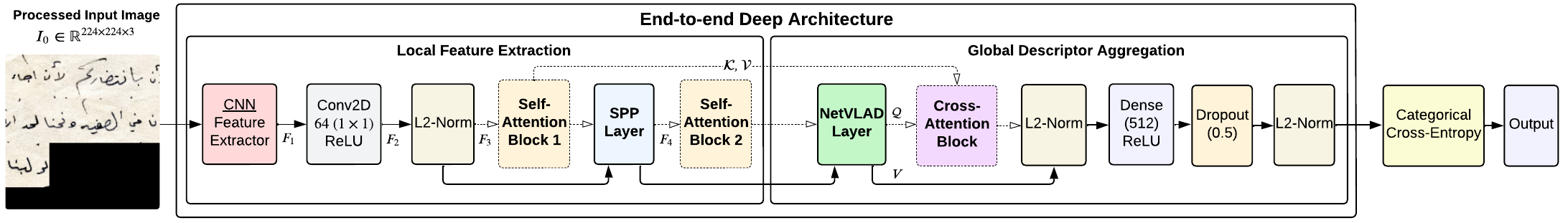}
    \vspace{-1em}
    \caption{Proposed end-to-end architecture illustrating both the attention-based and no-attention variants. The dashed blocks and arrows represent the optional attention path, which is active only in the attention-based version. In this configuration, the queries (\( \mathcal{Q} \)) for the cross-attention block are derived from the NetVLAD layer, while the keys (\( \mathcal{K} \)) and values (\( \mathcal{V} \)) are taken from the layer-normalized features preceding the first multi-head self-attention module.}
    \label{arch}
    \end{figure*}
    
    To address the writer identification problem, we designed an architecture capable of extracting both local and global features that capture the peculiar ways people write. Building upon Chammas et al.'s~\cite{chammas2024} optimized Deep-TEN framework, which is based on the original Deep-TEN architecture~\cite{zhang2017deep}, we propose a set of modifications and refinements that make the architecture better suited to the specific challenges of this task. Figure~\ref{arch} illustrates the two variants of our proposed architecture: one without attention mechanisms (our standard architecture) and one with attention mechanisms (our improved architecture). The overall pipeline comprises three main stages: feature extraction, encoding, and classification, integrated into a single end-to-end system.
    
    For a given input text-line image \( I \) of length \( L \) and width \( W \), intended for writer identification, we first process \( I \) through the BlockProcessor to produce a square RGB image \( I_0 \in \mathbb{R}^{224 \times 224 \times 3} \). Writer classification then begins by extracting local, hierarchical features from \( I_0 \) using a convolutional backbone. In our case, the baseline model configuration uses a ResNet50~\citep{Resnet} pre-trained on ImageNet~\citep{imagenet}, which outputs a deep feature map \( F_1 \in \mathbb{R}^{7 \times 7 \times 2048} \). The \(7 \times 7\) results from the standard downsampling in ResNet50, which reduces the input resolution by a factor of 32 through a series of stride-2 convolutions and pooling layers. This stage captures both low-level details, such as {handwriting} strokes, ligatures, and textures, and higher-level patterns, such as word shapes and stylistic cues. Next, a \(1 \times 1\) convolutional layer reduces the number of channels from 2048 to 64, yielding \( F_2 \in \mathbb{R}^{7 \times 7 \times 64} \), focusing the network on the most salient features while reducing computational cost. An L2-normalization layer is then applied to obtain \( F_3 \), ensuring scale-invariant feature representations that are critical for the stability of later aggregation stages. Before discussing our improved attention-based architecture, we will first consider our standard architecture. Specifically, we will describe the modified Spatial Pyramid Pooling (SPP) \citep{he2015spatial} and NetVLAD \citep{arandjelovic2016netvlad} layers as they are used in our standard architecture, later discussing their integration in the attention-based pipeline.    
    
    \begin{figure}[htbp]
    \centering
    \includegraphics[width=0.85\linewidth]{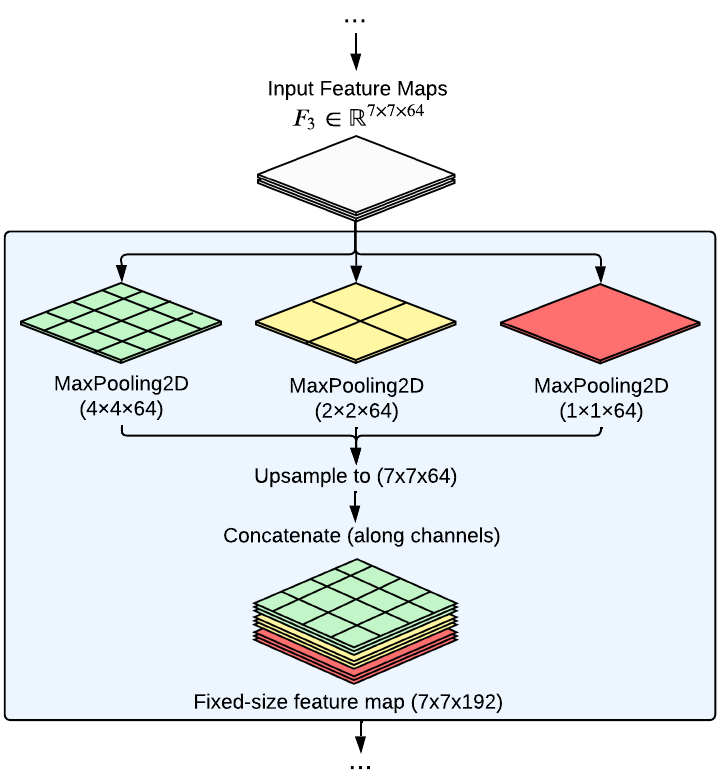}
    \caption{The modified SPP layer.}
    \label{spp}
    \end{figure}
    

    \begin{figure}[htbp]
    \centering
    \includegraphics[width=1.0\linewidth]{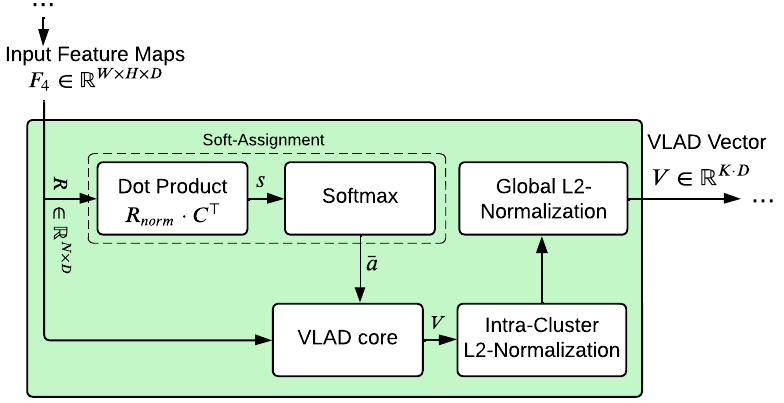}
    \vspace{-1em}
    \caption{The modified NetVLAD layer.}
    \label{vlad}
    \end{figure}

    After L2-normalizing the backbone features \( F_3 \), we apply a modified SPP layer that performs max pooling at multiple spatial scales, using pooling sizes \( n \times n \) where \( n \in \{1, 2, 4\} \). This captures coarse-to-fine contextual information from the normalized feature maps. Unlike the original SPP~\cite{he2015spatial}, which flattens the pooled outputs into a single feature vector, we upsample each pooled output back to the original resolution of \( 7 \times 7 \) and concatenate them channel-wise, producing a multi-scale feature map \( F_4 \in \mathbb{R}^{7 \times 7 \times 192} \) (see Figure~\ref{spp}). This preserves the spatial information in the feature maps and creates fixed-size outputs. More importantly, it supports the attention-based variant by maintaining aligned spatial dimensions required by the attention modules, ensuring seamless integration while retaining multi-scale context.
        
    The local descriptors in \( F_4 \) are then passed to the NetVLAD module, which aggregates local features into a compact global representation. As illustrated in Figure~\ref{vlad}, given an input feature map of size $W = H = 7$ and depth $D = 192$,  the resulting \( N = W \times H = 49 \) spatial descriptors are reshaped into a matrix $R \in \mathbb{R}^{N \times D}$, where each row corresponds to a spatial feature vector. To determine how each descriptor relates to \(K = 64\) learnable cluster centers, we compute similarity scores \(s \in \mathbb{R}^{N \times K}\) using the dot product between each L2-normalized descriptor in \(R\) and the cluster centers \(C \in \mathbb{R}^{K \times D}\). This effectively computes cosine similarity between normalized descriptors \(R_{\text{norm}}\) and cluster centers. Unlike the original NetVLAD~\cite{arandjelovic2016netvlad}, which uses a learnable \( 1 \times 1 \) convolution for assignments, we adopt this fixed cosine similarity scheme—retaining only the cluster centers as trainable weights and enabling a more efficient, deployment-friendly design.
        
    These scores are then softmax-normalized across clusters to produce soft assignments \( \bar{a} \), indicating the degree to which each descriptor belongs to each cluster. Next, for each cluster, the residuals are computed as the difference between every descriptor and the corresponding cluster center. Both the residuals and their corresponding soft assignments \( \bar{a} \) are fed into the VLAD core block, which performs a weighted sum of the residuals across all descriptors for each cluster—producing a residual aggregation matrix \( V \in \mathbb{R}^{K \times D} \). This matrix captures how local descriptors deviate from their assigned cluster centers.
        
    Finally, the matrix \( V \) is L2-normalized first within each cluster (row-wise), then globally (across the entire matrix), and flattened into a fixed-length vector \( V \in \mathbb{R}^{K \cdot D} \), serving as the global handwriting descriptor. This representation captures both the distribution and the spatial deviations of local features relative to learned patterns, enabling a discriminative comparison across writers.
  
\begin{figure}[htbp]
    \centering
    \includegraphics[width=1.0\linewidth]{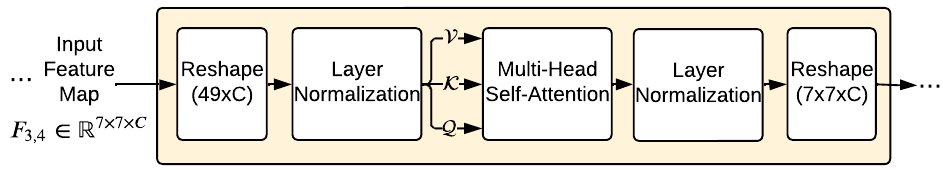}
    
    \caption{Self-Attention Block. Block 1 uses \(C = 64\) while Block 2 uses \(C = 192\). Both apply multi-head self-attention (6 heads, key dimension 32) with \( \mathcal{Q} = \mathcal{K} = \mathcal{V} \). Block 1’s output is reused as \(\mathcal{K}, \mathcal{V}\) in the cross-attention block.}
    \label{self}
    \end{figure}

\begin{figure}[htbp]
    \centering
\includegraphics[width=1.0\linewidth]{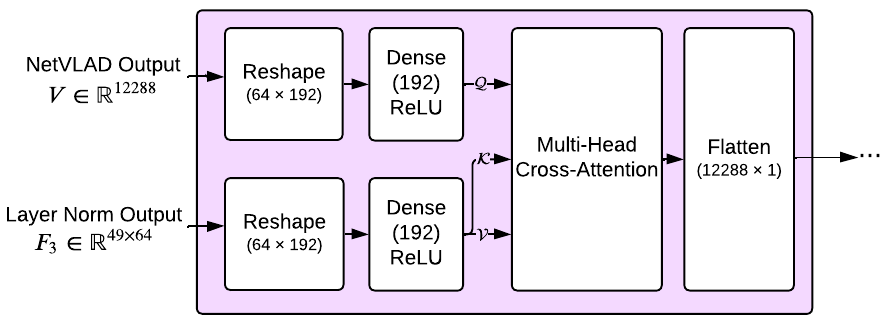}
    \caption{Cross-Attention Block. Multi-head attention (6 heads, key dimension 32) is applied. Here, \( \mathcal{Q} = \) NetVLAD output (global features), and \( \mathcal{K} = \mathcal{V} \) (refined local features from the feature extractors).}
    \label{cross}
    \end{figure}

   In the improved attention-based architecture, three attention blocks are inserted while leaving all tensor sizes unchanged. First, Self-Attention Block~1 operates on a reshaped version of \( F_3 \in \mathbb{R}^{7 \times 7 \times 64} \), which is flattened to \( \mathbb{R}^{49 \times 64} \), enabling the different spatial locations in the output feature maps to attend to one another before reshaping back. The SPP layer then produces \(F_{4}\) exactly as in the standard architecture, after which Self-Attention Block~2 refines these multi-scale features (reshaped as \( \mathbb{R}^{49 \times 192} \)) as seen in Figure~\ref{self}. This forces the model to focus on the most significant patterns across different scales, like how a writer connects the letter "\artext{ل}" (lām) to its adjacent letters to form a word. The NetVLAD module then encodes the refined \(F_{4}\) using the same procedure as before, but its resulting global descriptor $V$ now serves as the query \(\mathcal{Q}\) in a Cross-Attention Block (see Figure~\ref{cross}). Moreover, the keys and values \(\mathcal{K} = \mathcal{V}\) come from the layer-normalized output of Self-Attention Block~1. Then, the dimensions of these two inputs to the multi-head cross-attention module are aligned through two small dense layers. This final attention step fuses the global NetVLAD context with the refined prior local context, generating a richer global descriptor \(V\) that enables the system to balance capturing nuanced characteristics with understanding overarching stylistic patterns.
   


    Regardless of the path taken, each descriptor \(V\) is compacted and regularized by a 512-unit dense layer, dropout (\(p = 0.5\)), and a second L2 normalization. All convolutional and dense layers employ ReLU activations for efficient non-linearity, and both the \(1 \times 1\) convolution and the penultimate dense layer are penalized with L2 regularization \((\lambda = 1 \times 10^{-4})\). The combined effect of dropout and weight decay curbs overfitting, thereby improving the generalizability of the architecture.

    Given that our dataset contains sparse positive pairs, we replace the triplet loss used by Chammas et al.~\cite{chammas2024} with categorical cross-entropy loss, returning to the original Deep-TEN formulation \citep{zhang2017deep}. This change simplifies the training procedure by removing the need for hard triplet mining, while still achieving strong writer identification performance.

\subsection{Experimental Setup}
\label{sec:setup}
    
We set up experiments along two main model-design axes: (1) evaluating alternative feature extractors and (2) assessing the impact of transfer learning via pre-trained weights. { Each configuration is evaluated under both protocols A and B described in Section~\ref{data_preprocessing}. Across protocols, we keep the same pipeline and hyperparameters; only the final classification layer is resized to match the number of writer classes.}

\begin{sloppypar}
\textbf{Alternative feature extractors.} To assess the potential benefits of different backbone architectures beyond ResNet50 \citep{Resnet}, we evaluated DenseNet201 \citep{huang2017densely}, Xception \citep{chollet2017xception}, and \MNVL \citep{9008835} as interchangeable feature extractors within our pipeline. DenseNet201 leverages dense connectivity patterns to improve gradient flow and parameter efficiency, combining fine-grained stroke-level cues with broader handwriting-style features. Xception employs depthwise separable convolutions, factorizing spatial and channel-wise operations to better capture fine-grained spatial structure in a computationally efficient manner, with residual connections throughout. Similarly, \MNVL leverages depthwise separable convolutions within inverted residual blocks, alongside squeeze-and-excitation modules \citep{hu2018squeeze} and hard-swish activations---balancing accuracy and computational cost, and making it particularly attractive for future applications requiring on-device deployment~\citep{abushahla2025quantized}.
\end{sloppypar}
        
Notably, while the original ResNet50 backbone produces a deep feature map \( F_1 \in \mathbb{R}^{7 \times 7 \times 2048} \), which is identical to Xception's, DenseNet201 outputs \( F_1 \in \mathbb{R}^{7 \times 7 \times 1920} \) while \MNVL produces \( F_1 \in \mathbb{R}^{7 \times 7 \times 960} \). All other reported dimensions in the pipeline (e.g., after the \(1 \times 1\) convolution and subsequent layers) remain constant across these architectures.

\textbf{Transfer learning strategies.} We conducted experiments to explore the effects of pre-trained CNN backbones. As described earlier, the baseline configuration uses pre-trained ImageNet weights with frozen backbone layers. We extended this setup by testing fully fine-tuned models, partially fine-tuned models, and models trained entirely from scratch without any pre-trained weights. These experiments clarify the impact of transfer learning on performance and convergence behavior in the writer identification task.

\subsection{Model Training Setup}
\label{sec:training-setup}

Writer identification is approached as a \emph{closed-set} multi-class classification problem with each output neuron representing the probability of a particular writer class. With the preprocessing done in Section~\ref{data_preprocessing}, the specific hyperparameters used in the training process are detailed in Table~\ref{tab:general_training_parameters}. Additionally, we used a learning rate scheduler and an early stopping mechanism, whose parameters are summarized in Tables~\ref{tab:lr_scheduler_parameters} and~\ref{tab:early_stopping_parameters}, respectively. Lastly, each model configuration was trained across three different random seeds ($N{=}3$) to assess run-to-run variability.

\begin{table}[ht]
\caption{General Training Hyperparameters}
\label{tab:general_training_parameters}
\renewcommand{\arraystretch}{1.2} 
\footnotesize
\centering
\setlength{\tabcolsep}{5pt} 
\begin{tabular*}{\columnwidth}{@{\extracolsep{\fill}}ll} 
\toprule 
\textbf{Parameter} & \textbf{Value} \\ 
\midrule 
Optimizer & Adam \\ 
Loss Function & Categorical Cross-Entropy \\ 
Initial Learning Rate & \( 1 \times 10^{-3} \) \\ 
Batch Size & 256 \\ 
Number of Clusters (NetVLAD) & 64 \\ 
Maximum Number of Epochs & 450 \\ 
Dropout Rate & 0.5 \\ 
L2-Regularization & \( 1 \times 10^{-4} \) \\ 
\bottomrule 
\end{tabular*}
\end{table}

\vspace{-20pt}

\begin{table}[htbp]
\caption{Learning Rate Scheduler Parameters}
\label{tab:lr_scheduler_parameters}
\renewcommand{\arraystretch}{1.2} 
\footnotesize
\centering
\setlength{\tabcolsep}{5pt} 
\begin{tabular*}{\columnwidth}{@{\extracolsep{\fill}}ll} 
\toprule 
\textbf{Parameter} & \textbf{Value} \\ 
\midrule 
Learning Rate Scheduler & ReduceLROnPlateau \\ 
Scheduler Reduction Factor & 0.5 \\ 
Scheduler Patience & 10 epochs \\ 
Mode & Max \\ 
Minimum Learning Rate & \( 1 \times 10^{-8} \) \\ 
\bottomrule 
\end{tabular*}
\end{table}

\vspace{-25pt}

\begin{table}[htbp]
\caption{Early Stopping Parameters}
\label{tab:early_stopping_parameters}
\renewcommand{\arraystretch}{1.2} 
\footnotesize
\centering
\setlength{\tabcolsep}{5pt} 
\begin{tabular*}{\columnwidth}{@{\extracolsep{\fill}}ll} 
\toprule 
\textbf{Parameter} & \textbf{Value} \\ 
\midrule 
Early Stopping Metric & Validation Macro F1-Score \\ 
Early Stopping Patience & 50 epochs \\ 
Mode & Max \\ 
\bottomrule 
\end{tabular*}
\end{table}

\vspace{-25pt}

\subsection{Experimental Conditions}
\label{sec:exp-conditions}

{The first set of experiments was} conducted on two NVIDIA A100 (SXM4) Tensor Core GPUs with 80\,GB memory each: one accessed through the AUS AI Lab and one rented via RunPod\footnote{\url{https://www.runpod.io}}. {The second set of experiments was conducted on NVIDIA A10G GPUs via the AUS high-performance computing (HPC) infrastructure, complemented by NVIDIA RTX A6000 GPUs rented via RunPod for configurations requiring additional GPU memory.}

\section{Experimental Results}
\label{sec:results}

We evaluated each model configuration using standard classification metrics. { Results are reported under the two closed-set evaluation protocols defined in Section~\ref{data_preprocessing}. Results for Protocol A are shown in Tables~\ref{tab:top1_accuracy}--\ref{tab:f1_results}, while results for Protocol B are shown in Tables~\ref{tab:top1_accuracy_pagedisj}--\ref{tab:f1_results_pagedisj}. Additional results are provided in Appendix~\ref{app:additional-results}.
}

These results cover all experimental combinations we conducted: with and without attention, different feature extractors, and varying transfer learning setups. All tabulated results are reported as the mean and population standard deviation (computed with denominator $N$) across $N{=}3$ random seeds to illustrate performance variability. For clarity, below we summarize the training configurations shown in the tables:
\begin{itemize}
    \item \textbf{Frozen Pre-trained Feature Extractors:} The baseline configuration employs pre-trained feature extractors with their ImageNet weights fixed during training. These are denoted in the table with "Baseline."
    
    \item \textbf{Fine-tuning Pre-trained Weights:} This configuration involves fine-tuning the pre-trained ImageNet weights, ranging from partial to full fine-tuning, to adapt the model effectively to our handwriting domain. These are displayed in the table with either "Fine-tuned" for full fine-tuning or "Fine-tuned + Last \textit{X} Layers Fine-tuned" for partial fine-tuning, where \textit{X} denotes the number of unfrozen layers starting from the head of the architecture. 

    \item \textbf{Training from Scratch:} In this configuration, all network parameters are randomly initialized, and no pre-trained weights are used. These are shown in the table with "From Scratch."
\end{itemize}

\textbf{Evaluation protocol.} All reported results follow the \emph{closed-set} protocol in Section~\ref{data_preprocessing}, in which writers in the test set are also present during training. Therefore, the results measure discrimination among known writers, not the identification of unseen writers. { We first present results under Protocol A, followed by Protocol B, which is a more challenging setting that evaluates generalization to \emph{unseen pages} of the same writers.
}

\subsection{{Protocol A: Line-level evaluation}}
\label{sec:results_protocolA}

\begin{table*}[!htbp]
\vspace{-10pt}
\centering
\caption{Top-1 Accuracy Test Results. {The \textbf{best result} within Protocol A is \textbf{bolded}.}}
\label{tab:top1_accuracy}
\footnotesize
\renewcommand{\arraystretch}{1.0} 
\setlength{\tabcolsep}{5pt} 
\begin{tabular*}{\textwidth}{@{}l@{\extracolsep{\fill}}cccc@{}}
\toprule
\textbf{Model Configuration} & \multicolumn{4}{c}{\textbf{Top-1 Accuracy -- Mean (Standard Deviation)}} \\
\cmidrule(lr){2-5}
 & \textbf{ResNet50} & \textbf{DenseNet201} & \textbf{Xception} & \textbf{\MNVL} \\
\midrule
Frozen + No Attention (Baseline)  & 0.8652 (0.0028) & 0.8670 (0.0062) & 0.6983 (0.0144) & 0.7217 (0.0129) \\ 
Frozen + Attention                & 0.9114 (0.0041) & 0.9160 (0.0059) & 0.7543 (0.0070) & 0.7805 (0.0084) \\ 

Fine-tuned + Last Layer + No Attention & 0.8724 (0.0037) & 0.8698 (0.0058) & 0.7077 (0.0169) & 0.7343 (0.0050) \\ 
Fine-tuned + Last Layer + Attention    & 0.9121 (0.0045) & 0.9167 (0.0047) & 0.7495 (0.0017) & 0.8003 (0.0134) \\ 
Fine-tuned + Last 5 Layers + No Attention & 0.8974 (0.0050) & 0.8958 (0.0112) & 0.8450 (0.0079) & 0.8602 (0.0041) \\ 
Fine-tuned + Last 5 Layers + Attention    & 0.9320 (0.0033) & 0.9368 (0.0067) & 0.8711 (0.0081) & 0.8726 (0.0075) \\
Fine-tuned + Last 10 Layers + No Attention & 0.9469 (0.0009) & 0.9342 (0.0018) & 0.9383 (0.0063) & 0.8731 (0.0055) \\ 
Fine-tuned + Last 10 Layers + Attention    & 0.9437 (0.0061) & 0.9445 (0.0046) & 0.9433 (0.0031) & 0.8914 (0.0055) \\ 
Fine-tuned + Last 25 Layers + No Attention & 0.9564 (0.0019) & 0.9404 (0.0076) & 0.9488 (0.0047) & 0.9209 (0.0048) \\
Fine-tuned + Last 25 Layers + Attention    & 0.9599 (0.0033) & 0.9479 (0.0091) & 0.9472 (0.0019) & 0.9174 (0.0059) \\ 
Fine-tuned + No Attention         & 0.9802 (0.0048) & 0.9862 (0.0029) & 0.9865 (0.0020) & 0.9844 (0.0029) \\ 
Fine-tuned + Attention            & 0.9828 (0.0039) & \textbf{0.9905 (0.0008)} & 0.9861 (0.0010) & 0.9862 (0.0009) \\ 

From Scratch + No Attention       & 0.9655 (0.0045) & 0.9711 (0.0013) & 0.9792 (0.0011) & 0.4970 (0.3162) \\ 
From Scratch + Attention          & 0.9653 (0.0057) & 0.9674 (0.0038) & 0.9773 (0.0012) & 0.0502 (0.0000) \\ 
\bottomrule
\end{tabular*}
\end{table*}


\begin{table*}[!htbp]
\centering
\caption{Top-5 Accuracy Test Results. {The \textbf{best result} within Protocol A is \textbf{bolded}.}}
\label{tab:top5_accuracy}
\footnotesize
\renewcommand{\arraystretch}{1.0} 
\setlength{\tabcolsep}{5pt} 
\begin{tabular*}{\textwidth}{@{}l@{\extracolsep{\fill}}cccc@{}}
\toprule
\textbf{Model Configuration} & \multicolumn{4}{c}{\textbf{Top-5 Accuracy -- Mean (Standard Deviation)}} \\
\cmidrule(lr){2-5}
 & \textbf{ResNet50} & \textbf{DenseNet201} & \textbf{Xception} & \textbf{\MNVL} \\
\midrule
Frozen + No Attention (Baseline) & 0.9730 (0.0030) & 0.9689 (0.0013) & 0.8843 (0.0139) & 0.9120 (0.0052) \\ 
Frozen + Attention               & 0.9810 (0.0016) & 0.9806 (0.0016) & 0.9170 (0.0066) & 0.9375 (0.0050) \\ 
Fine-tuned + Last Layer + No Attention & 0.9736 (0.0028) & 0.9700 (0.0024) & 0.8945 (0.0103) & 0.9229 (0.0036) \\ 
Fine-tuned + Last Layer + Attention    & 0.9822 (0.0017) & 0.9814 (0.0012) & 0.9138 (0.0023) & 0.9440 (0.0062) \\ 
Fine-tuned + Last 5 Layers + No Attention & 0.9794 (0.0002) & 0.9792 (0.0009) & 0.9471 (0.0006) & 0.9651 (0.0010) \\ 
Fine-tuned + Last 5 Layers + Attention    & 0.9846 (0.0022) & 0.9842 (0.0017) & 0.9547 (0.0040) & 0.9689 (0.0018) \\
Fine-tuned + Last 10 Layers + No Attention & 0.9855 (0.0021) & 0.9851 (0.0012) & 0.9777 (0.0029) & 0.9651 (0.0021) \\ 
Fine-tuned + Last 10 Layers + Attention    & 0.9856 (0.0020) & 0.9864 (0.0007) & 0.9785 (0.0020) & 0.9698 (0.0010) \\ 
Fine-tuned + Last 25 Layers + No Attention & 0.9877 (0.0003) & 0.9846 (0.0000) & 0.9812 (0.0022) & 0.9681 (0.0030) \\
Fine-tuned + Last 25 Layers + Attention    & 0.9898 (0.0010) & 0.9854 (0.0018) & 0.9826 (0.0018) & 0.9704 (0.0042) \\ 
Fine-tuned + No Attention         & 0.9963 (0.0011) & 0.9970 (0.0007) & 0.9966 (0.0009) & 0.9960 (0.0007) \\ 
Fine-tuned + Attention            & 0.9966 (0.0010) & \textbf{0.9973 (0.0002)} & 0.9972 (0.0006) & 0.9960 (0.0004) \\ 
From Scratch + No Attention      & 0.9947 (0.0008) & 0.9946 (0.0007) & 0.9959 (0.0004) & 0.6495 (0.2954) \\ 
From Scratch + Attention         & 0.9940 (0.0010) & 0.9951 (0.0008) & 0.9945 (0.0006) & 0.2186 (0.0086) \\ 
\bottomrule
\end{tabular*}
\end{table*}



\begin{table*}[!htbp]
\centering
\caption{Macro F1-Score Test Results. {The \textbf{best result} within Protocol A is \textbf{bolded}.}}
\label{tab:f1_results}
\footnotesize
\renewcommand{\arraystretch}{0.95}
\setlength{\tabcolsep}{5pt}
\begin{tabular*}{\textwidth}{@{\extracolsep{\fill}}lcccc}
\toprule
\textbf{Model Configuration} & \multicolumn{4}{c}{\textbf{F1-score -- Mean (Standard Deviation)}} \\
\cmidrule(lr){2-5}
 & \textbf{ResNet50} & \textbf{DenseNet201} & \textbf{Xception} & \textbf{\MNVL} \\
\midrule
Frozen + No Attention (Baseline) & 0.6772 (0.0132) & 0.6587 (0.0280) & 0.3158 (0.0256) & 0.3704 (0.0378) \\ 
Frozen + Attention               & 0.7795 (0.0138) & 0.7878 (0.0218) & 0.4517 (0.0241) & 0.5086 (0.0034) \\ 

Fine-tuned + Last Layer + No Attention & 0.6757 (0.0260) & 0.6719 (0.0132) & 0.3525 (0.0442) & 0.3970 (0.0112) \\ 
Fine-tuned + Last Layer + Attention    & 0.7999 (0.0068) & 0.7968 (0.0115) & 0.4465 (0.0387) & 0.5439 (0.0225) \\ 
Fine-tuned + Last 5 Layers + No Attention & 0.7413 (0.0193) & 0.7204 (0.0329) & 0.6448 (0.0117) & 0.6488 (0.0149) \\ 
Fine-tuned + Last 5 Layers + Attention    & 0.8499 (0.0100) & 0.8538 (0.0127) & 0.7205 (0.0156) & 0.7115 (0.0074) \\ 
Fine-tuned + Last 10 Layers + No Attention & 0.8719 (0.0076) & 0.8096 (0.0168) & 0.8496 (0.0135) & 0.6760 (0.0396) \\ 
Fine-tuned + Last 10 Layers + Attention    & 0.8765 (0.0170) & 0.8474 (0.0046) & 0.8688 (0.0135) & 0.7573 (0.0150) \\ 
Fine-tuned + Last 25 Layers + No Attention & 0.9042 (0.0039) & 0.8353 (0.0188) & 0.8659 (0.0244) & 0.8255 (0.0134) \\ 
Fine-tuned + Last 25 Layers + Attention    & 0.9152 (0.0107) & 0.8659 (0.0222) & 0.8708 (0.0183) & 0.8185 (0.0054) \\ 
Fine-tuned + No Attention         & 0.9302 (0.0275) & 0.9557 (0.0185) & 0.9693 (0.0038) & 0.9573 (0.0096) \\ 
Fine-tuned + Attention            & 0.9413 (0.0210) & \textbf{0.9744 (0.0023)} & 0.9701 (0.0029) & 0.9690 (0.0046) \\ 
From Scratch + No Attention      & 0.8821 (0.0336) & 0.9052 (0.0124) & 0.9344 (0.0050) & 0.1187 (0.0839) \\ 
From Scratch + Attention         & 0.8859 (0.0332) & 0.9060 (0.0196) & 0.9451 (0.0017) & 0.0005 (0.0000) \\ 
\bottomrule
\end{tabular*}
\end{table*}

\subsubsection{Overall performance} 
\label{sec:overall}

Across the results reported in Tables~\ref{tab:top1_accuracy}--\ref{tab:f1_results} for all evaluated model configurations, the DenseNet201 + Fine-tuned + Attention configuration emerged as the top performer, achieving a Macro F1-score of \textbf{0.9744}~$\pm$~0.0023 despite the dataset's severe class imbalance. This setup handled both individual and composite author classes with high accuracy---particularly the highlighted two-author class (e.g., \textit{Yousef Hobeiche \& Angele Ellis}), which was perfectly classified despite its scarcity (Appendix~\ref{app:classification-reports}, Table~\ref{tab:app:crA-02}). On the other hand, our strongest baseline without fine-tuning or attention (ResNet50 + Frozen + No Attention) failed to capture such fine-grained distinctions, particularly for composite classes (Appendix~\ref{app:classification-reports}, Table~\ref{tab:app:crA-03}).

\subsubsection{Failure case analysis}
\label{sec:failure}

Despite the near-perfect performance of our best model configuration (DenseNet201 + Fine-tuned + Attention), the observed failure cases are dominated by the \emph{extreme long-tail} class distribution (Appendix~\ref{app:protocol-a-dist}) rather than by a systematic architectural weakness. Writers with only 1--3 test samples (75 out of 179 classes) exhibit the lowest and most volatile class-wise F1-scores (e.g., \textit{Father Elias}: F1-score $= 0.3333 \pm 0.4714$; \textit{Youssef Moussa Sadaqa}: F1-score $= 0.7222 \pm 0.2079$), since a single confusion can collapse recall and inflate variance across seeds. For classes with more test samples (at least 5), failures are comparatively rarer and tend to appear as isolated false negatives affecting a small subset of minority writers (e.g., \textit{Maurice Paul Sarrail}: F1-score $= 0.8329 \pm 0.1177$; \textit{Shukri Kanaan}: F1-score $= 0.9181 \pm 0.0262$; \textit{Mohammed Amin al-Husseini}: F1-score $= 0.9231 \pm 0.1088$).

Composite two-author labels are \emph{not} a dominant failure mode under our formulation, but interpretation is limited by very small \emph{test supports} (i.e., few test samples per class). In particular, the writer-pair classes considered here have only 2 test samples each (\textit{Yahia Mansour \& Asaad Koury}; \textit{Yousef Hobeiche \& Angele Ellis}; \textit{Thomas \& John Oussani}). With this caveat, our best model configuration still classifies these composite labels reliably: it achieves perfect F1-scores for \textit{Yahia Mansour \& Asaad Koury} and \textit{Yousef Hobeiche \& Angele Ellis}, while \textit{Thomas \& John Oussani} remains high (F1-score $= 0.9333 \pm 0.0943$). This indicates that mixed-writer lines can be modeled effectively as composite single-label classes even without explicitly segmenting the line image into writer-specific regions (i.e., without annotating which pixels or {handwriting} strokes belong to the first vs. second writer). Where constituent writers are available as standalone classes, the model can separate the single-writer from mixed-writer lines: \textit{Yousef Hobeiche} (F1-score $= 0.9974 \pm 0.0037$, support $= 65$) and \textit{Angele Ellis} (F1-score $= 1.0000 \pm 0.0000$, support $= 12$) remain near-perfect alongside their composite class. For \textit{Thomas \& John Oussani} and \textit{Yahia Mansour \& Asaad Koury}, the constituent writers do not appear as standalone labels in our test split, so their individual (non-composite) performance cannot be assessed directly.

In contrast, the strongest baseline (ResNet50 + Frozen + No Attention) exposes the limitations of non-adapted representations in this historical setting. Although its overall F1-score ($0.6772 \pm 0.0132$) remains moderate, the performance is disproportionately inflated by frequent writers. At the same time, the \emph{long tail} degrades sharply---the baseline yields 14 classes with zero-mean recall (and zero-mean F1-score). Interestingly, this weakness is not only confined to ultra-rare writers. Even among classes with at least 5 test samples, several remain poorly separated (e.g., \textit{Maurice Paul Sarrail}: F1-score $= 0$ across seeds; \textit{Youhanna Sfeir}: F1-score $= 0.2286 \pm 0.1682$; \textit{Father Tobia al-Issa}: F1-score $= 0.3368 \pm 0.1102$). This is consistent with the expectation that a frozen backbone cannot adapt to the script variability and degradation patterns of historical manuscripts. The composite two-writer labels further highlight this brittleness, but in distinct ways. First, the baseline collapses on \textit{Yahia Mansour \& Asaad Koury} (F1-score $= 0$ across seeds) even though in our test split, the constituent writer names do not appear as standalone labels. This indicates that frozen representations can fail to model mixed-writer lines even in the absence of an explicit constituent-vs-composite writer ambiguity. Second, for \textit{Yousef Hobeiche \& Angele Ellis}, where both constituent writers are present as standalone classes, the baseline attains reasonable single-writer performance (\textit{Yousef Hobeiche}: F1-score $= 0.8787 \pm 0.0290$; \textit{Angele Ellis}: F1-score $= 0.6799 \pm 0.0462$) but only partially recovers on the composite label (F1-score $= 0.1667 \pm 0.2357$). This gap is consistent with a more challenging decision boundary when the model must distinguish three closely related labels (writer A vs writer B vs writers A \textit{\& }B), making it easier for composite examples to be absorbed into a constituent class under a frozen backbone. Finally, the composite label of \textit{Thomas \& John Oussani} achieves a non-trivial F1-score of $0.7778 \pm 0.1571$, but its interpretation is constrained by having only two test samples, making the estimate highly seed-sensitive. Moreover, since its constituent writers do not appear as standalone labels, no conclusions can be drawn about constituent-vs-composite writer separation.


\subsection{Protocol B: Page-disjoint evaluation}
\label{sec:results_protocolB}
\begin{table*}[!htbp]
\vspace{-10pt}
\centering
\caption{{Top-1 Accuracy Test Results. The \textbf{best result} within Protocol B is \textbf{bolded}.}}
\label{tab:top1_accuracy_pagedisj}
\footnotesize
\renewcommand{\arraystretch}{1.0}
\setlength{\tabcolsep}{5pt}
{
\begin{tabular*}{\textwidth}{@{}l@{\extracolsep{\fill}}cccc@{}}
\toprule
\textbf{Model Configuration} & \multicolumn{4}{c}{\textbf{Top-1 Accuracy -- Mean (Standard Deviation)}} \\
\cmidrule(lr){2-5}
 & \textbf{ResNet50} & \textbf{DenseNet201} & \textbf{Xception} & \textbf{\MNVL} \\
\midrule
Frozen + No Attention (Baseline)  & 0.6988 (0.0089) & 0.6725 (0.0182) & 0.4974 (0.0065) & 0.6091 (0.0342) \\
Frozen + Attention                & 0.7036 (0.0189) & 0.7246 (0.0111) & 0.5527 (0.0177) & 0.6413 (0.0181) \\

Fine-tuned + Last Layer + No Attention & 0.6808 (0.0223) & 0.6729 (0.0309) & 0.4930 (0.0159) & 0.6136 (0.0103) \\
Fine-tuned + Last Layer + Attention    & 0.7162 (0.0176) & 0.7202 (0.0150) & 0.5444 (0.0246) & 0.6417 (0.0152) \\
Fine-tuned + Last 5 Layers + No Attention & 0.7121 (0.0215) & 0.6488 (0.0265) & 0.6356 (0.0203) & 0.6751 (0.0135) \\
Fine-tuned + Last 5 Layers + Attention    & 0.7237 (0.0156) & 0.6378 (0.1316) & 0.6333 (0.0481) & 0.6687 (0.0437) \\
Fine-tuned + Last 10 Layers + No Attention & 0.7317 (0.0273) & 0.7034 (0.0248) & 0.7319 (0.0218) & 0.6550 (0.0285) \\
Fine-tuned + Last 10 Layers + Attention    & 0.7293 (0.0289) & 0.6949 (0.0598) & 0.7060 (0.0322) & 0.6579 (0.0425) \\
Fine-tuned + Last 25 Layers + No Attention & 0.7587 (0.0228) & 0.7380 (0.0161) & 0.7590 (0.0295) & 0.7206 (0.0155) \\
Fine-tuned + Last 25 Layers + Attention    & 0.7566 (0.0282) & 0.7230 (0.0320) & 0.7583 (0.0229) & 0.7347 (0.0233) \\
Fine-tuned + No Attention         & 0.7517 (0.0299) & 0.7370 (0.0158) & 0.7675 (0.0063) & \textbf{0.7861 (0.0260)} \\
Fine-tuned + Attention            & 0.7442 (0.0180) & 0.7700 (0.0220) & 0.7706 (0.0171) & 0.7707 (0.0111) \\

From Scratch + No Attention       & 0.6592 (0.0271) & 0.6674 (0.0136) & 0.6913 (0.0514) & 0.0534 (0.0475) \\
From Scratch + Attention          & 0.6766 (0.0294) & 0.6615 (0.0255) & 0.6812 (0.0533) & 0.0156 (0.0078) \\
\bottomrule
\end{tabular*}
}
\end{table*}

\begin{table*}[!htbp]
\centering
\caption{{Top-5 Accuracy Test Results. The \textbf{best result} within Protocol B is \textbf{bolded}.}}
\label{tab:top5_accuracy_pagedisj}
\footnotesize
\renewcommand{\arraystretch}{1.0}
\setlength{\tabcolsep}{5pt}
{
\begin{tabular*}{\textwidth}{@{}l@{\extracolsep{\fill}}cccc@{}}
\toprule
\textbf{Model Configuration} & \multicolumn{4}{c}{\textbf{Top-5 Accuracy -- Mean (Standard Deviation)}} \\
\cmidrule(lr){2-5}
 & \textbf{ResNet50} & \textbf{DenseNet201} & \textbf{Xception} & \textbf{\MNVL} \\
\midrule
Frozen + No Attention (Baseline)  & 0.8735 (0.0147) & 0.8553 (0.0106) & 0.7411 (0.0032) & 0.8102 (0.0184) \\
Frozen + Attention                & 0.8736 (0.0131) & 0.8579 (0.0052) & 0.7722 (0.0104) & 0.8224 (0.0021) \\

Fine-tuned + Last Layer + No Attention & 0.8665 (0.0094) & 0.8585 (0.0147) & 0.7299 (0.0115) & 0.8080 (0.0027) \\
Fine-tuned + Last Layer + Attention    & \textbf{0.8779 (0.0154)} & 0.8523 (0.0032) & 0.7560 (0.0137) & 0.8209 (0.0018) \\
Fine-tuned + Last 5 Layers + No Attention & 0.8721 (0.0191) & 0.8430 (0.0081) & 0.8105 (0.0060) & 0.8489 (0.0122) \\
Fine-tuned + Last 5 Layers + Attention    & 0.8778 (0.0144) & 0.8244 (0.0509) & 0.8026 (0.0285) & 0.8120 (0.0210) \\
Fine-tuned + Last 10 Layers + No Attention & 0.8547 (0.0192) & 0.8556 (0.0029) & 0.8404 (0.0157) & 0.8143 (0.0058) \\
Fine-tuned + Last 10 Layers + Attention    & 0.8646 (0.0174) & 0.8415 (0.0286) & 0.8375 (0.0060) & 0.8140 (0.0164) \\
Fine-tuned + Last 25 Layers + No Attention & 0.8574 (0.0184) & 0.8583 (0.0080) & 0.8520 (0.0137) & 0.8328 (0.0204) \\
Fine-tuned + Last 25 Layers + Attention    & 0.8677 (0.0114) & 0.8543 (0.0088) & 0.8585 (0.0165) & 0.8375 (0.0169) \\
Fine-tuned + No Attention         & 0.8705 (0.0080) & 0.8657 (0.0069) & 0.8523 (0.0060) & 0.8688 (0.0033) \\
Fine-tuned + Attention            & 0.8682 (0.0108) & 0.8723 (0.0148) & 0.8681 (0.0098) & 0.8584 (0.0057) \\

From Scratch + No Attention       & 0.8166 (0.0243) & 0.8174 (0.0202) & 0.8337 (0.0334) & 0.0885 (0.0461) \\
From Scratch + Attention          & 0.8262 (0.0182) & 0.8210 (0.0262) & 0.8159 (0.0369) & 0.0890 (0.0487) \\
\bottomrule
\end{tabular*}}
\end{table*}

\begin{table*}[!htbp]
\centering
\caption{{Macro F1-Score Test Results. The \textbf{best result} within Protocol B is \textbf{bolded}.}}
\label{tab:f1_results_pagedisj}
\footnotesize
\renewcommand{\arraystretch}{0.95}
\setlength{\tabcolsep}{5pt}
{
\begin{tabular*}{\textwidth}{@{\extracolsep{\fill}}lcccc}
\toprule
\textbf{Model Configuration} & \multicolumn{4}{c}{\textbf{F1-score -- Mean (Standard Deviation)}} \\
\cmidrule(lr){2-5}
 & \textbf{ResNet50} & \textbf{DenseNet201} & \textbf{Xception} & \textbf{\MNVL} \\
\midrule
Frozen + No Attention (Baseline) & 0.5706 (0.0254) & 0.5324 (0.0306) & 0.3575 (0.0244) & 0.4715 (0.0116) \\
Frozen + Attention               & 0.5868 (0.0298) & 0.5988 (0.0308) & 0.4266 (0.0398) & 0.5100 (0.0387) \\

Fine-tuned + Last Layer + No Attention & 0.5462 (0.0282) & 0.5363 (0.0130) & 0.3524 (0.0443) & 0.4624 (0.0440) \\
Fine-tuned + Last Layer + Attention    & 0.6018 (0.0300) & 0.5885 (0.0516) & 0.3968 (0.0330) & 0.5125 (0.0442) \\
Fine-tuned + Last 5 Layers + No Attention & 0.5937 (0.0225) & 0.5073 (0.0912) & 0.5056 (0.0428) & 0.5322 (0.0317) \\
Fine-tuned + Last 5 Layers + Attention    & 0.6003 (0.0323) & 0.5123 (0.1760) & 0.4968 (0.0891) & 0.5298 (0.1034) \\
Fine-tuned + Last 10 Layers + No Attention & 0.6082 (0.0305) & 0.5785 (0.0681) & 0.6289 (0.0149) & 0.5115 (0.0882) \\
Fine-tuned + Last 10 Layers + Attention    & 0.6073 (0.0228) & 0.5442 (0.1193) & 0.5882 (0.0617) & 0.5077 (0.0908) \\
Fine-tuned + Last 25 Layers + No Attention & 0.6502 (0.0137) & 0.6053 (0.0564) & 0.6596 (0.0051) & 0.5910 (0.0386) \\
Fine-tuned + Last 25 Layers + Attention    & 0.6626 (0.0104) & 0.5874 (0.0863) & 0.6504 (0.0041) & 0.6203 (0.0198) \\
Fine-tuned + No Attention         & 0.6072 (0.0278) & 0.5991 (0.0288) & 0.6572 (0.0203) & 0.6544 (0.0209) \\
Fine-tuned + Attention            & 0.6009 (0.0458) & 0.6516 (0.0234) & 0.6444 (0.0164) & \textbf{0.6655 (0.0069)} \\

From Scratch + No Attention      & 0.4719 (0.0626) & 0.4895 (0.0266) & 0.5342 (0.1019) & 0.0014 (0.0012) \\
From Scratch + Attention         & 0.5116 (0.0574) & 0.4851 (0.0410) & 0.5070 (0.1185) & 0.0004 (0.0002) \\
\bottomrule
\end{tabular*}}
\end{table*}


\subsubsection{{Overall performance}}
\label{sec:overall_pagedisj}

{
\begin{sloppypar}
Under this stricter page-disjoint setting, performance decreases substantially across all backbones as seen in Tables~\ref{tab:top1_accuracy_pagedisj}--\ref{tab:f1_results_pagedisj}. Relative to Protocol A, for example, the best-performing line-level configuration (DenseNet201 + Fine-tuned + Attention) drops from a \(0.9744 \pm 0.0023\) to a \(0.6516 \pm 0.0234\) F1-score. This sharp degradation indicates that the near-perfect line-level results are partially benefiting from within-page shared cues (e.g., scan artifacts, background texture, and other page-specific characteristics) in addition to writer-invariant handwriting features. Nonetheless, page-disjoint Top-5 accuracy remains relatively high for multiple configurations: ResNet50 + Fine-tuned + Last Layer + Attention achieves the highest Top-5 accuracy at 0.8779~$\pm$~0.0154, while DenseNet201 + Fine-tuned + Attention achieves 0.8723~$\pm$~0.0148. This suggests that the learned representations still capture writer-related information, but exhibit greater ambiguity when generalizing to unseen pages, primarily impacting strict Top-1 decisions.

Across all evaluated configurations in Protocol B, \MNVL becomes the strongest model under F1-score: the Fine-tuned + Attention setting achieves \textbf{0.6655} \(\pm~0.0069\) with comparatively low variance. Interestingly, the best Top-1 accuracy is attained by a \emph{different} \MNVL configuration (Fine-tuned + No Attention) at \textbf{\(0.7861 \pm 0.0260\)} (see Table~\ref{tab:top1_accuracy_pagedisj}), further highlighting the competitiveness of lightweight backbones under stricter page-disjoint generalization.

Finally, unlike Protocol A, where the best-performing setup is consistently aligned across metrics, Protocol B exhibits a clearer decoupling of metrics, with the best Top-1, Top-5, and F1 scores achieved under different configurations. This suggests that, under page-disjoint evaluation, architectural/training choices trade off between strict top-1 correctness and broader top-k separability, and reinforces that conclusions should be drawn using multiple complementary metrics rather than a single best configuration.
\end{sloppypar}
}

\subsubsection{Failure case analysis}
\label{sec:failure_pagedisj}

Inspecting the classification report of the best-performing Protocol B configuration (\MNVL + Fine-tuned + Attention) clarifies why performance is substantially lower than in the line-level split (see Appendix~\ref{app:classification-reports}, Table~\ref{tab:app:crB-mbv3}). When a writer has a sufficiently large number of pages (see Appendix~\ref{app:protocol-b-dist}), the model typically achieves strong per-class performance, indicating that it can learn a stable writing style and correctly identify the writer even from lines originating from a single unseen test page. Representative examples include \textit{Emily Nasrallah} (48 pages; F1-score: \(1.0000 \pm 0.0000\)), \textit{Salah Tizani} (48 pages; F1-score: \(0.9937 \pm 0.0067\)), \textit{Hanna Moussa} (41 pages; F1-score: \(0.9989 \pm 0.0016\)), and even \textit{Hanna Ghayth}, which attains an F1-score of \(0.9096 \pm 0.0108\).

By contrast, many of the weakest classes (including near-zero F1-scores) correspond to writers with only 3--4 pages in total, for whom the training split provides too little \emph{page diversity} to reliably capture writer-invariant features. However, low page count is not the only driver. There are notable exceptions where writers with only 3--4 pages still achieve high scores, such as \textit{Botros Feghaly} (3 pages; F1-score: \(0.9626 \pm 0.0207\)), \textit{Elie Maroun Khalil} (4 pages; F1-score: \(0.9608 \pm 0.0075\)), and \textit{Murshid Habashi} (4 pages; F1-score: \(0.9139 \pm 0.0639\)). In these cases, the pages originate from the same document (or letter), yielding highly consistent page-level characteristics (paper tone, ink density, and overall writing appearance), which makes the class comparatively easy to recognize even under page-disjoint splitting.

\begin{figure*}[htbp]
\centering

\begin{subfigure}[t]{0.32\textwidth}
\centering
\includegraphics[width=\linewidth]{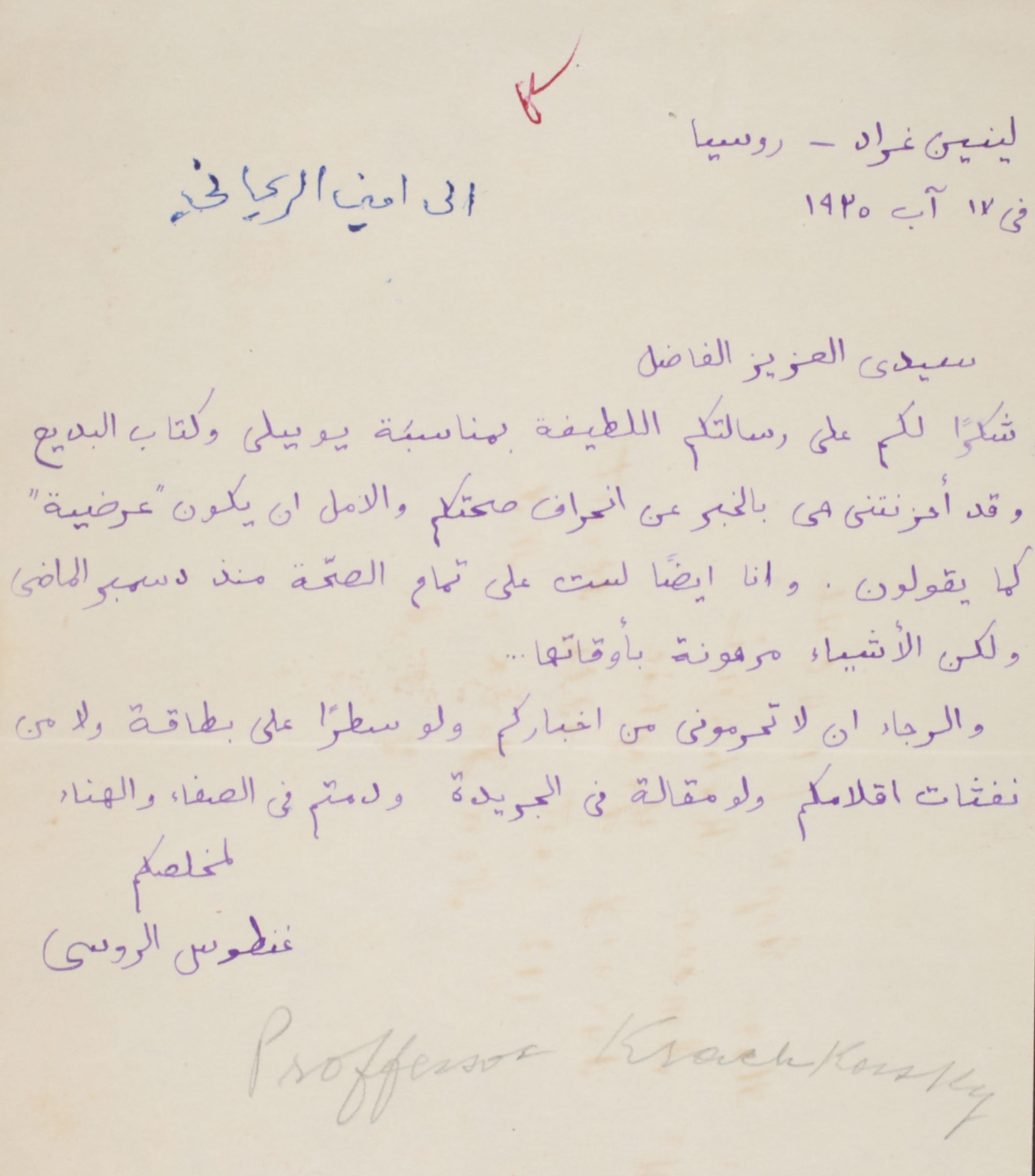}
\caption{{\textit{Prof.\ Ign.\ Kratchkovsky} -- Page 1}}
\label{fig:fail_kratch_p1}
\end{subfigure}\hfill
\begin{subfigure}[t]{0.32\textwidth}
\centering
\includegraphics[width=\linewidth]{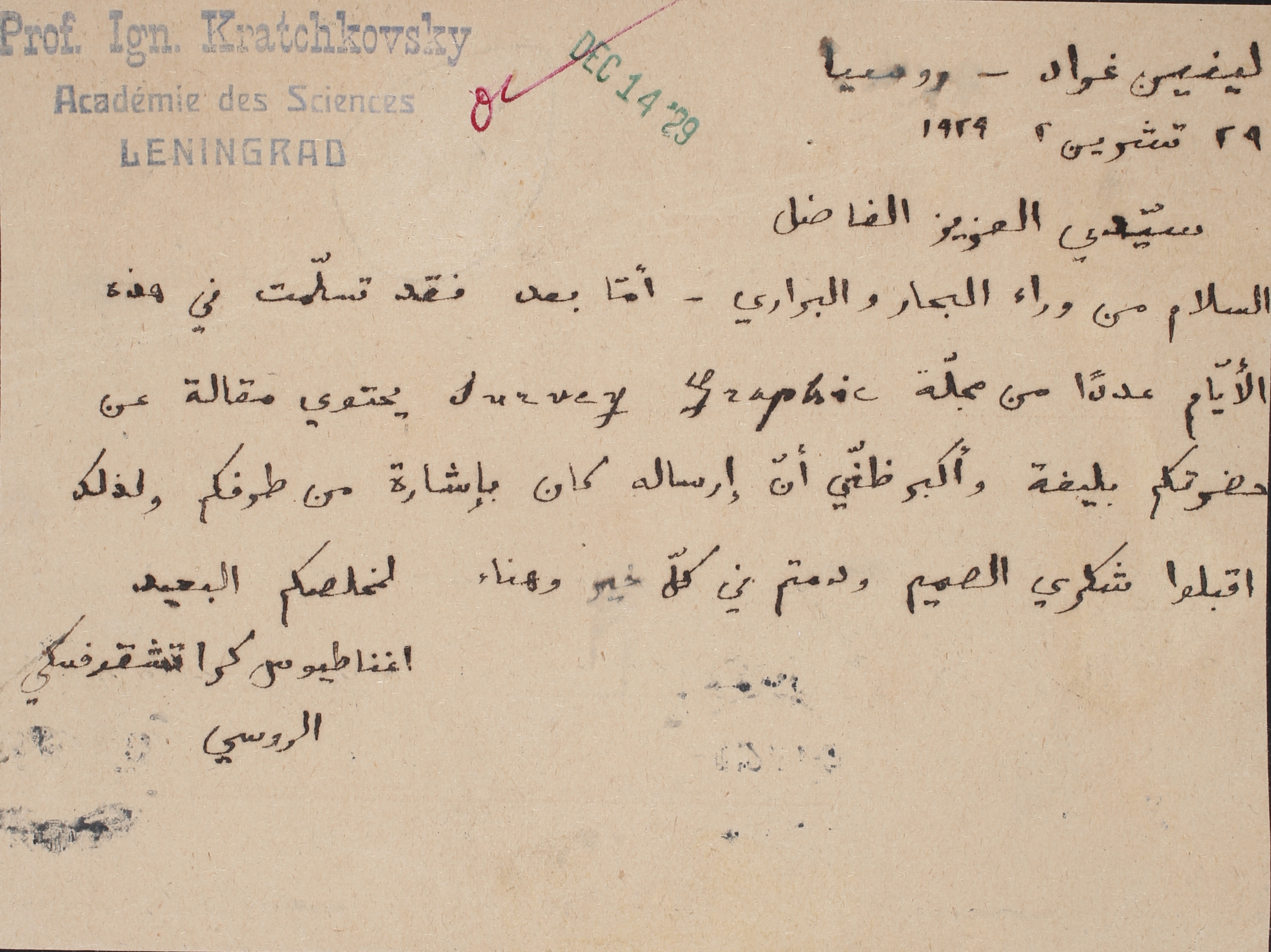}
\caption{{\textit{Prof.\ Ign.\ Kratchkovsky} -- Page 2}}
\label{fig:fail_kratch_p2}
\end{subfigure}\hfill
\begin{subfigure}[t]{0.32\textwidth}
\centering
\includegraphics[width=\linewidth]{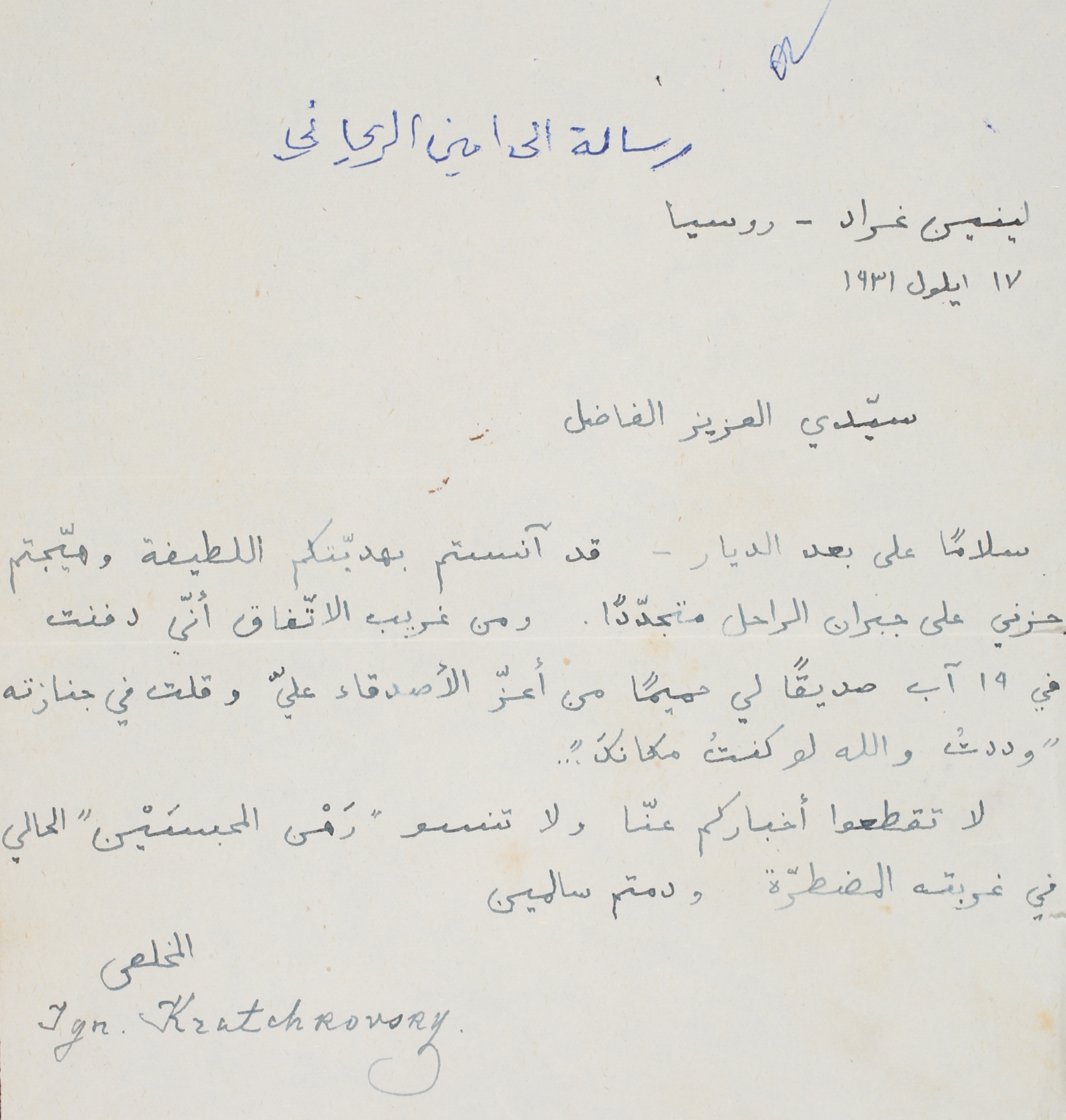}
\caption{{\textit{Prof.\ Ign.\ Kratchkovsky} -- Page 3}}
\label{fig:fail_kratch_p3}
\end{subfigure}

\vspace{2mm}

\begin{subfigure}[t]{0.32\textwidth}
\centering
\includegraphics[width=\linewidth]{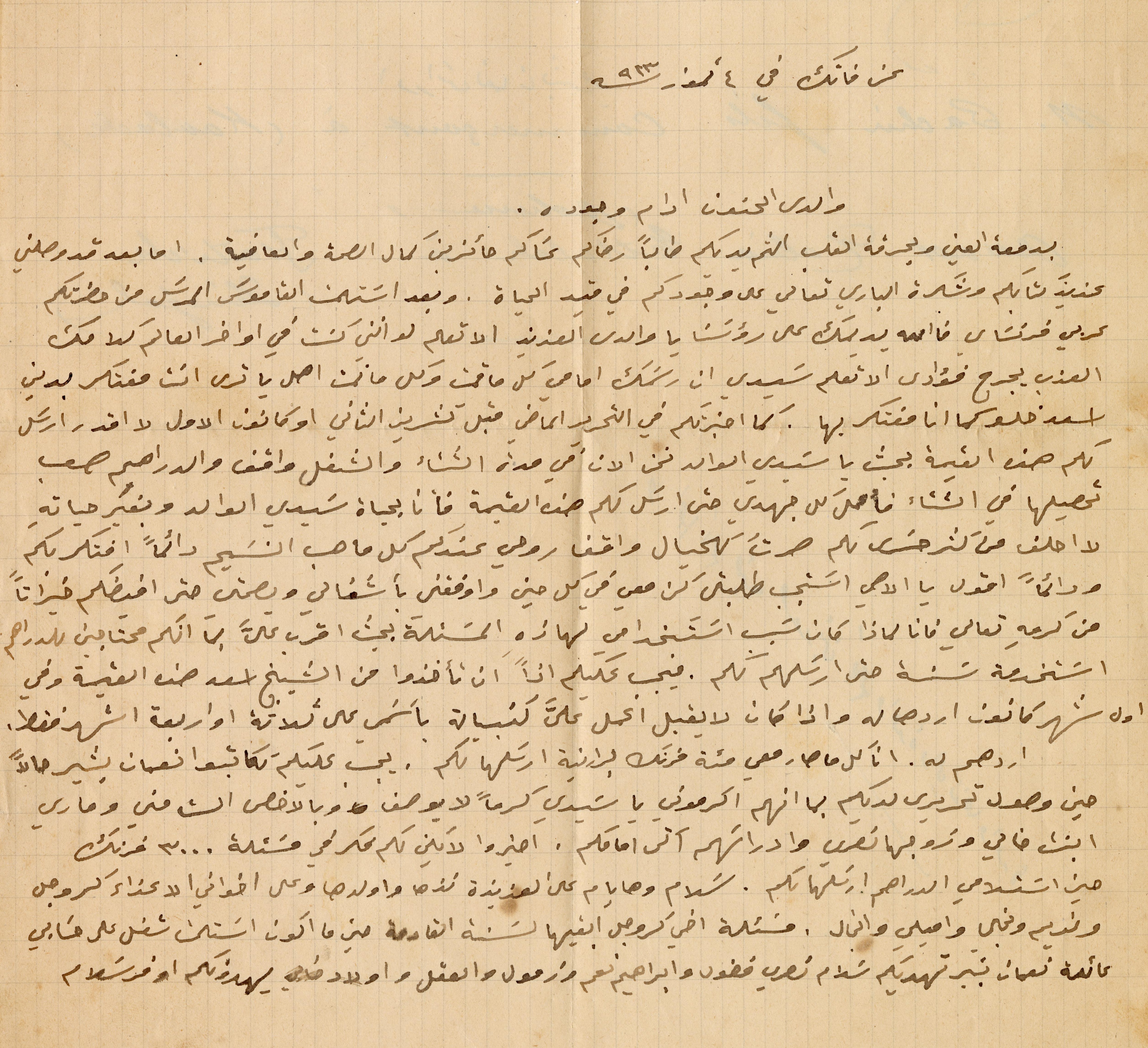}
\caption{{\textit{Asaad Koury Attallah} -- Page 1}}
\label{fig:fail_attallah_p1}
\end{subfigure}\hfill
\begin{subfigure}[t]{0.32\textwidth}
\centering
\includegraphics[width=\linewidth]{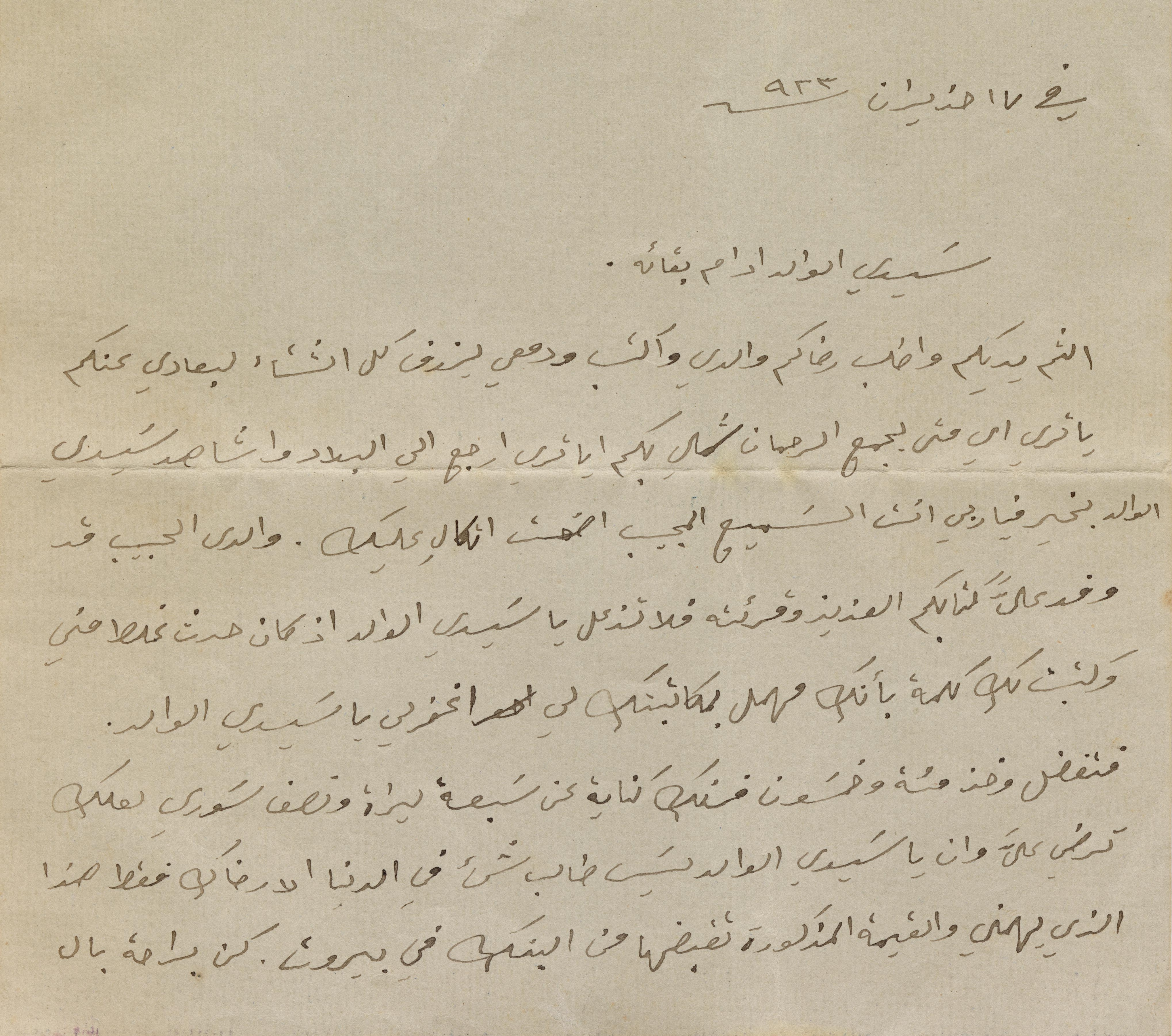}
\caption{{\textit{Asaad Koury Attallah} -- Page 2}}
\label{fig:fail_attallah_p2}
\end{subfigure}\hfill
\begin{subfigure}[t]{0.32\textwidth}
\centering
\includegraphics[width=\linewidth]{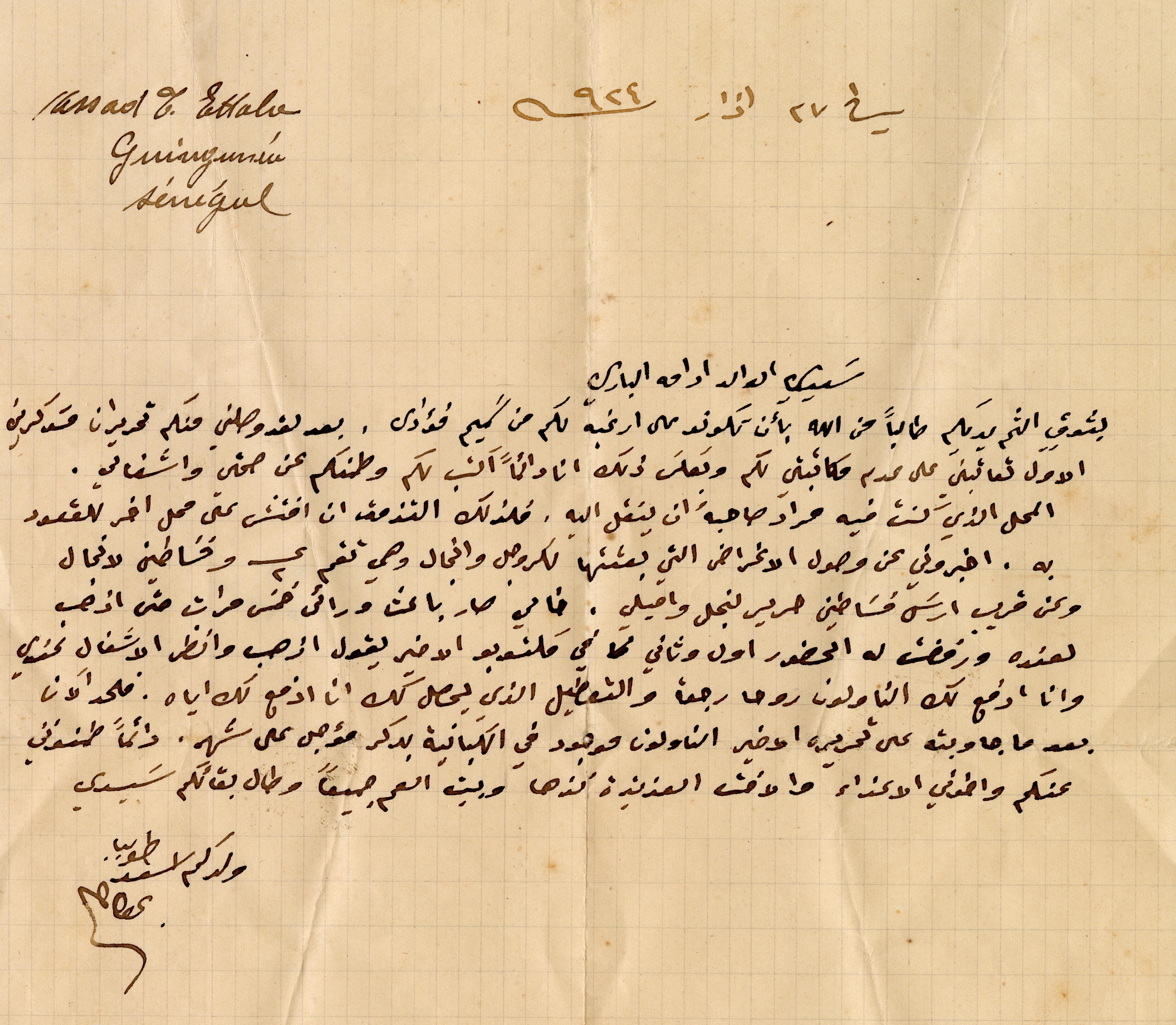}
\caption{{\textit{Asaad Koury Attallah} -- Page 3}}
\label{fig:fail_attallah_p3}
\end{subfigure}

\vspace{2mm}

\begin{subfigure}[t]{0.32\textwidth}
\centering
\includegraphics[width=\linewidth]{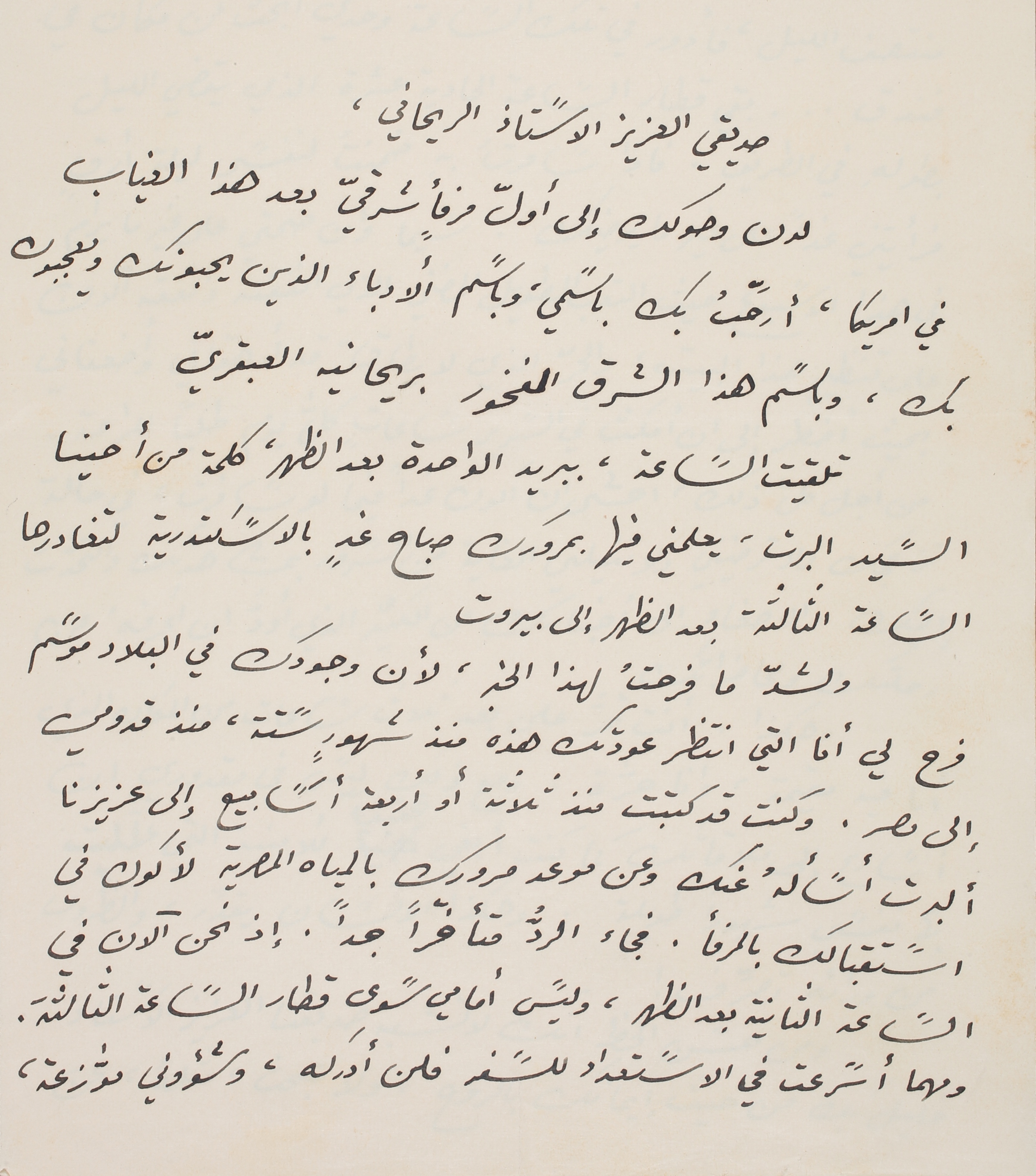}
\caption{{\textit{May Ziadeh} -- Page 1}}
\label{fig:fail_ziadeh_p1}
\end{subfigure}\hfill
\begin{subfigure}[t]{0.32\textwidth}
\centering
\includegraphics[width=\linewidth]{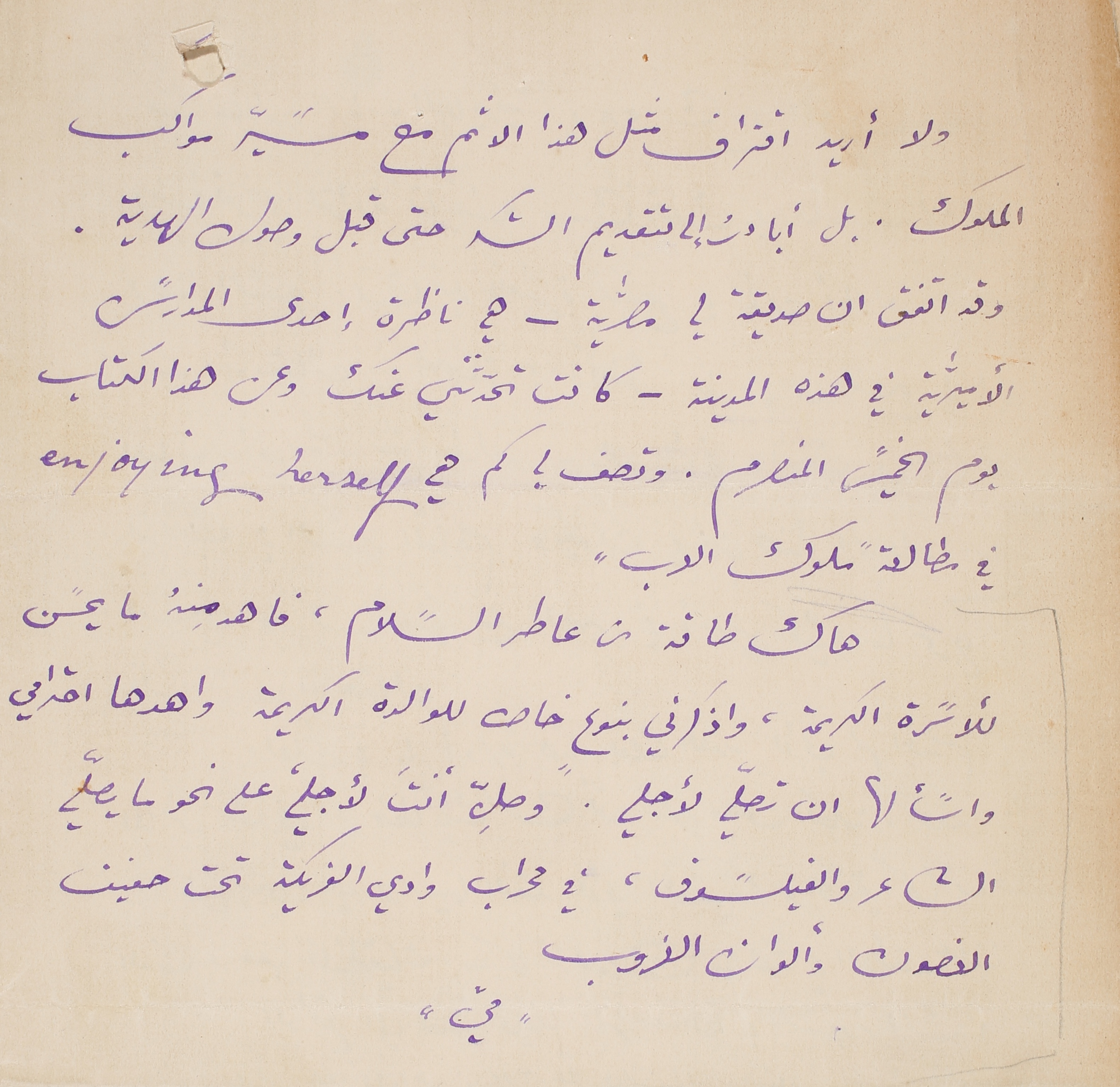}
\caption{{\textit{May Ziadeh} -- Page 2}}
\label{fig:fail_ziadeh_p2}
\end{subfigure}\hfill
\begin{subfigure}[t]{0.32\textwidth}
\centering
\includegraphics[width=\linewidth]{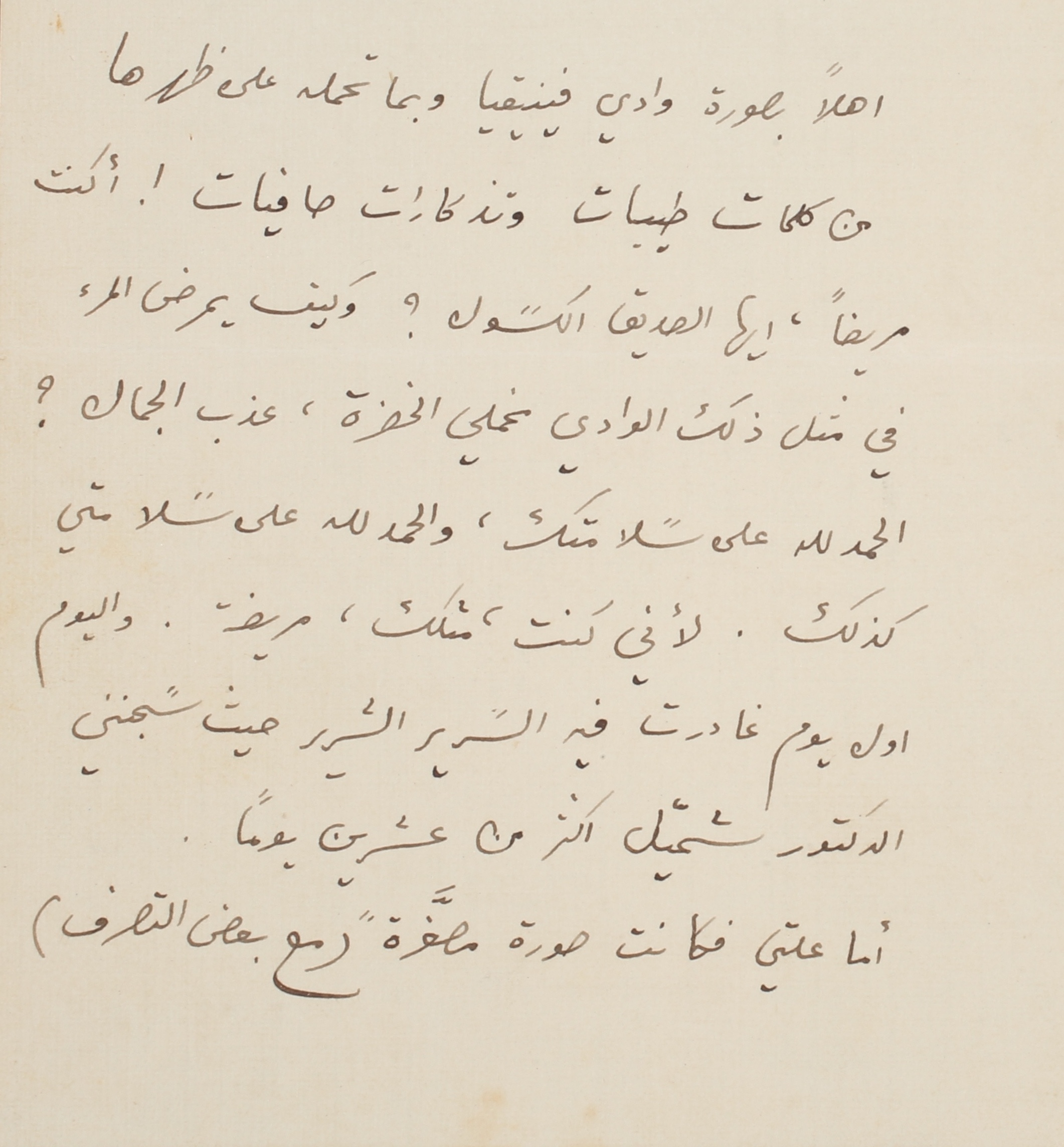}
\caption{{\textit{May Ziadeh} -- Page 3}}
\label{fig:fail_ziadeh_p3}
\end{subfigure}

\caption{{Intra-writer page variability under Protocol B for \textit{Prof.\ Ign.\ Kratchkovsky} (a--c), \textit{Asaad Koury Tobia Attallah} (d--f), and \textit{May Ziadeh} (g--i). Across pages, noticeable changes in paper tone, ink density, and handwriting appearance highlight a key source of error under page-disjoint evaluation.}}
\label{fig:failure_intra_writer_variability}
\end{figure*}

Composite two-author labels are even more constrained under Protocol B. Since our page-disjoint splitting requires at least 3 pages per class, this form of evaluation includes fewer writer classes than Protocol A, thereby removing many ultra-rare writers (see Appendix~\ref{app:protocol-b-dist}). Most composite classes fall below this page threshold and are therefore absent from this protocol. The main exception is \textit{Father Youssef Hanna \& Father Botros Hasan} with 3 pages in total, where the single test page exhibits noticeably different page appearance from the training pages, leading to a low and highly seed-sensitive F1-score of \(0.2821 \pm 0.3989\). Moreover, among its constituent writers, only \textit{Father Botros Hasan} appears as a standalone class under Protocol B. Despite having 6 pages overall, it still attains a low F1-score of \(0.3416 \pm 0.2150\), consistent with moderate-to-high intra-writer variability across pages. Given these small supports, we avoid over-interpreting per-class fluctuations for composite labels and treat them as illustrative rather than definitive.

Conversely, the most severe failures often arise when a writer’s pages exhibit pronounced intra-writer variability in handwriting appearance, ink, and page tone (see Figure~\ref{fig:failure_intra_writer_variability}). This is particularly evident for \textit{Prof.\ Ign.\ Kratchkovsky} (6 pages), where the pages are largely single-page letters written in different years and the line images differ substantially (Figures~\ref{fig:fail_kratch_p1}--\ref{fig:fail_kratch_p3}); consequently, the test-page lines deviate sharply from the writer representation learned from the training pages, causing the class to collapse to an \(\mathrm{F1\text{-}score}=0\) across seeds. A similar pattern appears for \textit{Asaad Koury Tobia Attallah} (3 pages), where all pages are visually distinct (Figures~\ref{fig:fail_attallah_p1}--\ref{fig:fail_attallah_p3}) and performance again drops to an \(\mathrm{F1\text{-}score}=0\) across seeds. Another illustrative case is \textit{May Ziadeh}, with 12 pages in total. While some pages come from the same letter (consistent with the intermediate \(\mathrm{F1\text{-}score~of~} 0.3591 \pm 0.2549\)), other pages differ noticeably across letters in page tone and ink properties and even in handwriting style (e.g., faster writing and less consistent letter shaping; Figures~\ref{fig:fail_ziadeh_p1}--\ref{fig:fail_ziadeh_p3}), which increases confusion under page-disjoint evaluation.

Furthermore, the strongest baseline under Protocol B (ResNet50 + Frozen + No Attention) exposes the limitations of non-adapted representations when generalizing across unseen pages. While its Top-5 accuracy remains relatively high (\(0.8735 \pm 0.0147\)), its F1-score drops to \(0.5706 \pm 0.0254\), indicating that many writers remain only weakly separable once page-specific cues change.

Overall, Protocol B exposes substantial within-writer variability across pages, and this effect is amplified for minority writers. While fewer pages generally correlate with lower performance (with the exceptions noted above), the dominant failure mode is the combination of limited page coverage and high intra-writer variation. This confirms that page-disjoint evaluation is markedly more challenging and highlights that Muharaf remains a non-trivial dataset under realistic generalization constraints.

\subsection{{Cross-protocol analysis}}
\label{sec:cross_protocol}

\subsubsection{Impact of attention mechanisms}
\label{sec:attn_cross}

{ Across both protocols, attention generally improves performance, but the gain is not uniform and depends on the quality of the underlying feature representation.

Under the line-level split,} attention mechanisms consistently boosted performance across architectures (see Tables~\ref{tab:top1_accuracy}--\ref{tab:f1_results} and Appendix~\ref{app:additional-results}). However, attention alone was insufficient when paired with weak or frozen feature representations. In particular, attention improved results on fine-tuned backbones but offered marginal benefit or even harm when representations lacked discriminative power, especially for two-author classes (see Appendix~\ref{app:classification-reports}). Overall, attention was most effective when applied to refined embeddings adapted to the handwriting domain. This indicates that the combination of deeper fine-tuning and attention is critical for robust performance in Protocol A, as neither component alone reliably captured both local and global handwriting cues. Additional evidence is provided by the DenseNet201 classification reports in Tables~\ref{tab:app:crA-01}--\ref{tab:app:crA-02} and~\ref{tab:app:crA-07}--\ref{tab:app:crA-10} of Appendix~\ref{app:classification-reports}, particularly for composite author classes.

{\begin{sloppypar}
In contrast, under the page-disjoint split, the effect of attention becomes more configuration-dependent (see Tables~\ref{tab:top1_accuracy_pagedisj}--\ref{tab:f1_results_pagedisj}). With frozen backbones, attention yields consistent gains in F1-score across all architectures, suggesting that attention can partially compensate for non-adapted features in this harder setting. However, under full fine-tuning, the trend is mixed: attention improves DenseNet201 ($0.5991$ $\to$ $0.6516$) and \MNVL ($0.6544$ $\to$ $0.6655$), but slightly degrades ResNet50 ($0.6072$ $\to$ $0.6009$) and Xception ($0.6572$ $\to$ $0.6444$). This inconsistency also appears in Top-1 and Top-5, where the best entries come from different configurations (e.g., Top-5 peaks at ResNet50 with last-layer fine-tuning + attention). Overall, unlike Protocol A, where attention is reliably beneficial once the backbone is sufficiently adapted, Protocol B shows that attention is not uniformly advantageous and should be carefully considered alongside the fine-tuning depth and metric.
\end{sloppypar}}

\subsubsection{Effect of fine-tuning depth}
\label{sec:ftdepth_cross}

{ Deeper fine-tuning is a key driver of performance in both evaluation protocols, but it behaves more predictably in Protocol A than in Protocol B. Under Protocol A, progressively unfreezing layers yields clear gains:} unfreezing only the last layer already improved \MNVL's F1-score to 0.5439~$\pm$~0.0225. Unfreezing the last 25 layers in DenseNet201 \emph{with attention} yielded 0.8659~$\pm$~0.0222---significant, but still inferior to full fine-tuning. Interestingly, when limited fine-tuning was combined with attention, performance sometimes declined, suggesting that suboptimal representations may misguide attention modules. Ultimately, fully fine-tuning all layers led to the best results across all models.

{Under Protocol B, the same general pattern holds---fine-tuning deeper tends to help---but the improvements are less monotonic and sometimes accompanied by higher variability, reflecting the added difficulty of page-level domain shift. This indicates that partial adaptation may be insufficient to learn page-invariant handwriting cues, and that stronger domain adaptation is typically required to generalize to unseen pages.}

{Meanwhile, training from scratch consistently underperformed in both protocols} even with attention, reinforcing the advantage of transfer learning from ImageNet initialization. These weights provided essential feature-detection capabilities, accelerating convergence and enabling the extraction of nuanced handwriting features, especially in limited-data scenarios such as ours.

\subsubsection{Lightweight model behavior} \MNVL exhibited highly variable performance across training strategies. {Under Protocol A}, when fine-tuned with attention, it achieved an F1-score of \textbf{0.9690}~$\pm$~0.0046---surpassing ResNet50 under the same conditions and approaching DenseNet201. However, \MNVL collapsed when trained from scratch, yielding the worst overall scores (0.0005~$\pm$~0.0000 with attention; 0.1187~$\pm$~0.0839 without attention). This discrepancy can be attributed to the shared training settings with the heavier models and the absence of meaningful initial weights, which likely caused convergence to a poor solution or triggered early stopping before learning useful representations.

{The same sensitivity to initialization is even more apparent under Protocol B. When fine-tuned from ImageNet weights, \MNVL remains competitive with heavier backbones and achieves the highest F1-score across all page-disjoint configurations (Fine-tuned + Attention: \textbf{0.6655}~$\pm$~0.0069), while also attaining the best Top-1 accuracy in Protocol B (Fine-tuned + No Attention: \textbf{0.7861}~$\pm$~0.0260). In contrast, training from scratch again collapses (F1-score of $0.0014~\pm~0.0012$ without attention and $0.0004~\pm~0.0002$ with attention), confirming that, in this setting, lightweight models rely critically on transferable pre-trained features and are unlikely to be trainable reliably under the same training settings used for the heavier architectures.
}

\subsubsection{Architectural considerations} 
\label{sec:arch_cross}

Lastly, the channel dimensionality of each backbone’s output tensors provides useful architectural context but does not reliably predict downstream performance. DenseNet201 produces a \(\mathbb{R}^{7 \times 7 \times 1920}\) tensor, ResNet50 and Xception a \(\mathbb{R}^{7 \times 7 \times 2048}\) tensor, and \MNVL a more compact \(\mathbb{R}^{7 \times 7 \times 960}\) tensor. Although a higher number of channels is often associated with richer feature representations, our results show that channel count alone is not a dependable indicator of writer identification performance.

{Under Protocol A, DenseNet201 achieves the strongest results when fully fine-tuned with attention, despite having fewer channels than ResNet50 and Xception. However, under Protocol B, the strongest backbone is not unanimous across the metrics. While different \MNVL configurations attain the best Top-1 accuracy and F1-score, the peak Top-5 accuracy arises from a different feature extractor (ResNet50 with last-layer fine-tuning + attention). This divisive result indicates that the most effective inductive biases can change once the model is forced to predict the author of a page not seen in the training set from that particular writer.
}

Conversely, under training-from-scratch conditions, Xception tends to be the most resilient among the larger models, whereas both DenseNet201 and ResNet50 exhibit noticeable drops. The extreme sensitivity of \MNVL to initialization---collapsing entirely without pre-training---further emphasizes that success may not simply be a matter of channel depth. Instead, these outcomes highlight the importance of architectural design choices, such as Xception’s depth-wise separable filters, the presence of transferable pre-trained features, and the integration of attention mechanisms---all of which interact to enhance representation quality and ultimately enable more effective writer identification.

\section{Limitations and Future Work}
\label{sec:limitations}

Our data-driven approach to writer identification highlights the importance of high-quality, diverse handwriting data. While our system achieved strong results, several limitations present opportunities for future research.

\subsection{Dataset Limitations}

\textbf{Manual Labeling Challenges.} The ambiguities in authorship attribution due to the lack of distinctive features or metadata of some handwritten samples significantly affected further labeling efforts. For instance, some that could be labeled were either too difficult to do so, like the Ottoman Turkish texts (see Figure~\ref{fig:ottoman-lines}), while others were not handwritten (see Figure~\ref{fig:typewritten-lines}).

\textbf{Dataset Imbalance.} The significant class imbalance across both protocols (see Appendices~\ref{app:protocol-a-dist} and~\ref{app:protocol-b-dist}) posed difficulties in model training and generalization by favoring overrepresented writers and, under Protocol~B, pages that are more similar to those observed during training. Moreover, not all composite writer classes have individual handwritten text for each author. This limitation further restricts the model’s ability to effectively distinguish between certain writers, especially when their handwriting samples are found on the same page or manuscript.

\subsection{Future Research Directions}
\label{sec:future-directions}

\textbf{Dataset Refinement and Augmentation.} Extending manual labeling, possibly using historical metadata, would enhance data quality. Furthermore, given the difficulty of collecting large numbers of authentic samples, synthetic data generation using generative models~\citep{mayr2025zero} could help improve minority-class performance and support few-shot learning scenarios. {In light of the performance gap under page-disjoint evaluation, a practical next step is to refine preprocessing to better normalize page appearance and emphasize ink strokes, for example, via binarization (e.g., Otsu thresholding) and contrast normalization. Complementarily, stronger augmentation tailored to historical documents (background/texture perturbations, illumination shifts, stain/noise simulation) could improve robustness to page-specific variations.} Finally, semi- and self-supervised learning strategies, including pseudo-labeling and iterative refinement, offer promising avenues for leveraging unlabeled data more effectively in this context.

{\textbf{Architectural Improvements and Deployment Studies.} Architectural improvements that explicitly de-emphasize background and reinforce stroke-driven discrimination are warranted, for example, through gating or improved attention mechanisms that suppress background noise and prioritize handwriting strokes.} Moreover, exploring alternative models (e.g., transformer-based architectures like ViTs~\citep{ViT}) with positional encodings may further enhance performance, particularly for mixed-script handwriting. {Overall, closing the gap between the near-perfect line-level results and the page-disjoint setting remains an important future goal.} Finally, these advances can be paired with quantization and lightweight design choices~\citep{abushahla2025quantized} to enable practical, real-time on-device writer identification.

\textbf{Open-set and Out-of-distribution Generalization.} Since this work uses a \emph{closed-set} protocol, an important next step is to evaluate \emph{open-set} identification with writer-disjoint splits and protocols that allow an \emph{unknown-writer} outcome. Alongside this direction, out-of-distribution (OOD) robustness can be studied via cross-collection testing (e.g., different archives, acquisition conditions, or script mixtures) under distribution shifts. Both directions naturally motivate embedding-based verification/retrieval (query--gallery matching), where samples are compared by distances in a learned embedding space, and unseen writers can be supported by enrolling new gallery examples. Recent work on open writer identification via distance-based matching provides a valuable starting point for designing such open-set protocols and exploring cross-domain generalization trends~\citep{briber2024open}.

\vspace{-10pt}

\section{Conclusion}
\label{sec:conclusion}

This work strengthens the Muharaf dataset as a benchmark for \emph{closed-set} writer identification by substantially expanding its labeled subset from 6,858 (28.00\%) to 21,249 (86.75\%) lines via manual verification and retaining 18,987 (77.51\%) lines after filtering. Building on this refined data, we developed an end-to-end CNN-based system augmented with attention mechanisms to classify handwritten text-line images from historical Arabic manuscripts and {evaluated it across fourteen model configurations under two complementary protocols: (i) a line-level split and (ii) a stricter page-disjoint split that assigns all lines from each page to a single split. To the best of our knowledge, this is the first study to report writer-identification baselines on Muharaf under both protocols.} Under the line-level protocol, DenseNet201 + Attention achieves the best results (Top-1 accuracy: \textbf{99.05\%}, Top-5 accuracy: \textbf{99.73\%}, F1-score: \textbf{97.44\%}). {In contrast, the stricter page-disjoint evaluation yields lower performance (Top-1 accuracy: \textbf{78.61\%}, Top-5 accuracy: \textbf{87.79\%}, F1-score: \textbf{66.55\%}), thus quantifying the influence of page-level cues and the difficulty of generalizing to unseen pages.}

These findings reinforce the practical value of transfer learning and layer-wise fine-tuning for writer identification when labeled data are limited{, while also showing that near-ceiling line-level performance does not directly translate to page-level generalization.} {Accordingly, by pairing our expanded labeled subset with benchmarks reported under both line-level and page-disjoint protocols, we provide a clearer and more realistic reference point for future work on page-invariant modeling, robustness to acquisition artifacts, and broader evaluation settings (e.g., open-set identification).} This study not only contributes to a stronger experimental foundation for Arabic writer identification but also emphasizes the potential of combining refined datasets with DL techniques. By leveraging these methods, we can better utilize the vast amounts of remaining unlabeled data, thereby advancing the field of writer identification in Arabic manuscripts even further and unlocking the hidden narratives preserved within these timeless documents.

\vspace{-10pt}
\section*{Author Contributions}
\vspace{-5pt}
\noindent
\textbf{Hamza A. Abushahla}: Conceptualization, Data curation, Investigation, Writing – original draft, Writing – review \& editing, Validation, Visualization. \textbf{Ariel Justine N. Panopio}: Conceptualization, Methodology, Investigation, Software, Writing – original draft, Writing – review \& editing, Validation. \textbf{Layth Al-Khairulla}: Data curation, Investigation, Writing – review \& editing, Validation. \textbf{Mohamed I. AlHajri}: Supervision, Resources, Writing – review \& editing.

\vspace{-10pt}
\section*{Declaration of competing interest}
\vspace{-5pt}
The authors declare that they have no known competing financial interests or personal relationships that could have appeared to influence the work reported in this article.

\vspace{-10pt}
\section*{Data availability}
\vspace{-5pt}
The data, code, and supplementary materials for this study are available on \href{https://github.com/7abushahla/Muharaf-Writer-Identification}{GitHub}. The original public Muharaf dataset is available on \href{https://doi.org/10.5281/zenodo.11492215}{Zenodo}. 

\bibliographystyle{IEEEtran}
\bibliography{sn-bibliography}

@article{saeed2024muharaf,
  title={Muharaf: Manuscripts of handwritten Arabic dataset for cursive text recognition},
  author={Saeed, Mehreen and Chan, Adrian and Mijar, Anupam and Habchi, Gerges and Younes, Carlos and Wong, Chau-Wai and Khater, Akram and others},
  journal={Advances in Neural Information Processing Systems},
  volume={37},
  pages={58525--58538},
  year={2024}
}

@article{rehman2019writer,
  title={Writer identification using machine learning approaches: a comprehensive review},
  author={Rehman, Arshia and Naz, Saeeda and Razzak, Muhammad Imran},
  journal={Multimedia Tools and Applications},
  volume={78},
  pages={10889--10931},
  year={2019},
  publisher={Springer}
}

@inproceedings{chammas2020,
  title={Writer identification for historical handwritten documents using a single feature extraction method},
  author={Chammas, Michel and Makhoul, Abdallah and Demerjian, Jacques},
  booktitle={2020 19th IEEE International Conference on Machine Learning and Applications (ICMLA)},
  pages={1--6},
  year={2020},
  organization={IEEE}
}

@article{levenshtein1966binary,
  title={Binary codes capable of correcting deletions, insertions, and reversals},
  author={Levenshtein, VI},
  journal={Proceedings of the Soviet physics doklady},
  year={1966}
}

@INPROCEEDINGS{9008835,
  author={Howard, Andrew and Sandler, Mark and Chen, Bo and Wang, Weijun and Chen, Liang-Chieh and Tan, Mingxing and Chu, Grace and Vasudevan, Vijay and Zhu, Yukun and Pang, Ruoming and Adam, Hartwig and Le, Quoc},
  booktitle={2019 IEEE/CVF International Conference on Computer Vision (ICCV)}, 
  title={Searching for MobileNetV3}, 
  year={2019},
  volume={},
  number={},
  pages={1314-1324},
  doi={10.1109/ICCV.2019.00140}}

@inproceedings{hu2018squeeze,
  title={Squeeze-and-excitation networks},
  author={Hu, Jie and Shen, Li and Sun, Gang},
  booktitle={Proceedings of the IEEE conference on computer vision and pattern recognition},
  pages={7132--7141},
  year={2018}
}

@article{chammas2024,
  title={An End-to-End deep learning system for writer identification in handwritten Arabic manuscripts},
  author={Chammas, Michel and Makhoul, Abdallah and Demerjian, Jacques and Dannaoui, Elie},
  journal={Multimedia Tools and Applications},
  volume={83},
  number={18},
  pages={54569--54589},
  year={2024},
  publisher={Springer}
}

@inproceedings{ngo2021vlad,
  title={A-VLAD: An end-to-end attention-based neural network for writer identification in historical documents},
  author={Ngo, Trung Tan and Nguyen, Hung Tuan and Nakagawa, Masaki},
  booktitle={International Conference on Document Analysis and Recognition},
  pages={396--409},
  year={2021},
  organization={Springer}
}

@article{boyi2024evolution,
  title={The Evolution and History of the Arabic Language},
  author={Boyi, Ahmad Muhammad and Yusuf, Muhammad Badamasi and Isa, Mustapha Muhammad},
  journal={Journal of African Resilience and Advancement Research},
  year={2024}
}

@article{asi2017,
  title={On writer identification for Arabic historical manuscripts},
  author={Asi, Abedelkadir and Abdalhaleem, Alaa and Fecker, Daniel and M{\"a}rgner, Volker and El-Sana, Jihad},
  journal={International Journal on Document Analysis and Recognition (IJDAR)},
  volume={20},
  pages={173--187},
  year={2017},
  publisher={Springer}
}

@article{cilia2020end,
  title={An end-to-end deep learning system for medieval writer identification},
  author={Cilia, Nicole Dalia and De Stefano, Claudio and Fontanella, Francesco and Marrocco, Claudio and Molinara, Mario and Di Freca, A Scotto},
  journal={Pattern Recognition Letters},
  volume={129},
  pages={137--143},
  year={2020},
  publisher={Elsevier}
}

@article{cilia2020experimental,
  title={An experimental comparison between deep learning and classical machine learning approaches for writer identification in medieval documents},
  author={Cilia, Nicole Dalia and De Stefano, Claudio and Fontanella, Francesco and Marrocco, Claudio and Molinara, Mario and Freca, Alessandra Scotto di},
  journal={Journal of Imaging},
  volume={6},
  number={9},
  pages={89},
  year={2020},
  publisher={MDPI}
}

@article{kumar2024attention,
  title={Attention based End to end network for Offline Writer Identification on Word level data},
  author={Kumar, Vineet and Sundaram, Suresh},
  journal={arXiv preprint arXiv:2404.07602},
  year={2024}
}

@article{chan2024hatformer,
  title={HATFormer: Historic Handwritten Arabic Text Recognition with Transformers},
  author={Chan, Adrian and Mijar, Anupam and Saeed, Mehreen and Wong, Chau-Wai and Khater, Akram},
  journal={arXiv preprint arXiv:2410.02179},
  year={2024}
}

@article{mahmoud2014khatt,
  title={KHATT: An open Arabic offline handwritten text database},
  author={Mahmoud, Sabri A and Ahmad, Irfan and Al-Khatib, Wasfi G and Alshayeb, Mohammad and Parvez, Mohammad Tanvir and M{\"a}rgner, Volker and Fink, Gernot A},
  journal={Pattern Recognition},
  volume={47},
  number={3},
  pages={1096--1112},
  year={2014},
  publisher={Elsevier}
}

@inproceedings{abdelhaleem2017wahd,
  title={Wahd: a database for writer identification of arabic historical documents},
  author={Abdelhaleem, Alaa and Droby, Ahmed and Asi, Abedelkader and Kassis, Majeed and Al Asam, Reem and El-sanaa, Jihad},
  booktitle={2017 1st International workshop on arabic script analysis and recognition (ASAR)},
  pages={64--68},
  year={2017},
  organization={IEEE}
}

@inproceedings{zhang2017deep,
  title={Deep ten: Texture encoding network},
  author={Zhang, Hang and Xue, Jia and Dana, Kristin},
  booktitle={Proceedings of the IEEE conference on computer vision and pattern recognition},
  pages={708--717},
  year={2017}
}

@inproceedings{arandjelovic2016netvlad,
  title={NetVLAD: CNN architecture for weakly supervised place recognition},
  author={Arandjelovic, Relja and Gronat, Petr and Torii, Akihiko and Pajdla, Tomas and Sivic, Josef},
  booktitle={Proceedings of the IEEE conference on computer vision and pattern recognition},
  pages={5297--5307},
  year={2016}
}

@inproceedings{srivastava2021exploiting,
  title={Exploiting multi-scale fusion, spatial attention and patch interaction techniques for text-independent writer identification},
  author={Srivastava, Abhishek and Chanda, Sukalpa and Pal, Umapada},
  booktitle={Asian Conference on Pattern Recognition},
  pages={203--217},
  year={2021},
  organization={Springer}
}

@inproceedings{koepf2022writer,
  title={Writer identification and writer retrieval using vision transformer for forensic documents},
  author={Koepf, Michael and Kleber, Florian and Sablatnig, Robert},
  booktitle={International Workshop on Document Analysis Systems},
  pages={352--366},
  year={2022},
  organization={Springer}
}

@inproceedings{xing2016deepwriter,
  title={Deepwriter: A multi-stream deep CNN for text-independent writer identification},
  author={Xing, Linjie and Qiao, Yu},
  booktitle={2016 15th international conference on frontiers in handwriting recognition (ICFHR)},
  pages={584--589},
  year={2016},
  organization={IEEE}
}

@article{chahi2023effective,
  title={An effective DeepWINet CNN model for off-line text-independent writer identification},
  author={Chahi, Abderrazak and El-Merabet, Youssef and Ruichek, Yassine and Touahni, Raja},
  journal={Pattern Analysis and Applications},
  volume={26},
  number={3},
  pages={1539--1556},
  year={2023},
  publisher={Springer}
}

@article{srihari2002individuality,
  title={Individuality of handwriting},
  author={Srihari, Sargur N and Cha, Sung-Hyuk and Arora, Hina and Lee, Sangjik},
  journal={Journal of Forensic Sciences},
  volume={47},
  number={4},
  pages={856--872},
  year={2002},
  publisher={ASTM AMERICAN SOCIETY FOR TESTING AND MATERIAL}
}

@article{djeddi2013text,
  title={Text-independent writer recognition using multi-script handwritten texts},
  author={Djeddi, Chawki and Siddiqi, Imran and Souici-Meslati, Labiba and Ennaji, Abdellatif},
  journal={Pattern Recognition Letters},
  volume={34},
  number={10},
  pages={1196--1202},
  year={2013},
  publisher={Elsevier}
}

@inproceedings{bertolini2016multi,
  title={Multi-script writer identification using dissimilarity},
  author={Bertolini, Diego and Oliveira, Luiz S and Sabourin, Robert},
  booktitle={2016 23rd International Conference on Pattern Recognition (ICPR)},
  pages={3025--3030},
  year={2016},
  organization={IEEE}
}

@article{vaswani2017attention,
  title={Attention is all you need},
  author={Vaswani, Ashish and Shazeer, Noam and Parmar, Niki and Uszkoreit, Jakob and Jones, Llion and Gomez, Aidan N and Kaiser, {\L}ukasz and Polosukhin, Illia},
  journal={Advances in neural information processing systems},
  volume={30},
  year={2017}
}

@article{VGG,
  title={Very deep convolutional networks for large-scale image recognition},
  author={Simonyan, Karen and Zisserman, Andrew},
  journal={arXiv preprint arXiv:1409.1556},
  year={2014}
}

@inproceedings{Resnet,
  title={Deep residual learning for image recognition},
  author={He, Kaiming and Zhang, Xiangyu and Ren, Shaoqing and Sun, Jian},
  booktitle={Proceedings of the IEEE conference on computer vision and pattern recognition},
  pages={770--778},
  year={2016}
}

@article{ViT,
  title={An image is worth 16x16 words: Transformers for image recognition at scale},
  author={Dosovitskiy, Alexey},
  journal={arXiv preprint arXiv:2010.11929},
  year={2020}
}

@article{he2015spatial,
  title={Spatial pyramid pooling in deep convolutional networks for visual recognition},
  author={He, Kaiming and Zhang, Xiangyu and Ren, Shaoqing and Sun, Jian},
  journal={IEEE transactions on pattern analysis and machine intelligence},
  volume={37},
  number={9},
  pages={1904--1916},
  year={2015},
  publisher={IEEE}
}

@inproceedings{imagenet,
  title={Imagenet: A large-scale hierarchical image database},
  author={Deng, Jia and Dong, Wei and Socher, Richard and Li, Li-Jia and Li, Kai and Fei-Fei, Li},
  booktitle={2009 IEEE conference on computer vision and pattern recognition},
  pages={248--255},
  year={2009},
  organization={Ieee}
}

@inproceedings{huang2017densely,
  title={Densely connected convolutional networks},
  author={Huang, Gao and Liu, Zhuang and Van Der Maaten, Laurens and Weinberger, Kilian Q},
  booktitle={Proceedings of the IEEE conference on computer vision and pattern recognition},
  pages={4700--4708},
  year={2017}
}

@inproceedings{pechwitz2002ifn,
  title={IFN/ENIT-database of handwritten Arabic words},
  author={Pechwitz, Mario and Maddouri, S Snoussi and M{\"a}rgner, Volker and Ellouze, Noureddine and Amiri, Hamid and others},
  booktitle={Proc. of CIFED},
  volume={2},
  pages={127--136},
  year={2002},
  organization={Citeseer}
}

@article{he2017writer,
  title={Writer identification using curvature-free features},
  author={He, Sheng and Schomaker, Lambert},
  journal={Pattern Recognition},
  volume={63},
  pages={451--464},
  year={2017},
  publisher={Elsevier}
}

@inproceedings{chollet2017xception,
  title={Xception: Deep learning with depthwise separable convolutions},
  author={Chollet, Fran{\c{c}}ois},
  booktitle={Proceedings of the IEEE conference on computer vision and pattern recognition},
  pages={1251--1258},
  year={2017}
}

@article{briber2024open,
  title={Open writer identification from handwritten text fragments using lite convolutional neural network},
  author={Briber, Amina and Chibani, Youcef},
  journal={International Journal on Document Analysis and Recognition (IJDAR)},
  volume={27},
  number={4},
  pages={529--551},
  year={2024},
  publisher={Springer}
}

@article{yang2024dt2f,
  title={DT2F-TLNet: A novel text-independent writer identification and verification model using a combination of deep type-2 fuzzy architecture and Transfer Learning networks based on handwriting data},
  author={Yang, Jing and Shokouhifar, Mohammad and Yee, Lip and Khan, Abdullah Ayub and Awais, Muhammad and Mousavi, Zohreh},
  journal={Expert Systems with Applications},
  volume={242},
  pages={122704},
  year={2024},
  publisher={Elsevier}
}

@article{mayr2025zero,
  title={Zero-shot paragraph-level handwriting imitation with latent diffusion models},
  author={Mayr, Martin and Dreier, Marcel and Kordon, Florian and Seuret, Mathias and Z{\"o}llner, Jochen and Wu, Fei and Maier, Andreas and Christlein, Vincent},
  journal={International Journal of Computer Vision},
  pages={1--22},
  year={2025},
  publisher={Springer}
}

@article{fatnassi2025st,
  title={ST-WID: Self-supervised transformer for writer identification in arabic handwritten scripts: I. Fatnassi et al.},
  author={Fatnassi, Islem and Khamekhem Jemni, Sana and Ammar, Sourour and Kessentini, Yousri},
  journal={Signal, Image and Video Processing},
  volume={19},
  number={14},
  pages={1190},
  year={2025},
  publisher={Springer}
}

@article{okawa2025multistage,
  title={Multistage Convolutional Neural Network With Deformable Attention for Word-Level Offline Text-Independent Writer Identification},
  author={Okawa, Manabu},
  journal={IEEE Access},
  year={2025},
  publisher={IEEE}
}

@inproceedings{majithia2025integrated,
  title={An Integrated Convolutional and Transformer Architecture for Word-Based Handwriter Identification},
  author={Majithia, Aditya and Pedersen, Arthur Paul and Grossberg, Michael},
  booktitle={IFIP International Conference on Artificial Intelligence Applications and Innovations},
  pages={44--59},
  year={2025},
  organization={Springer}
}

@article{maitra2025decorrelation,
  title={Decorrelation-Based Self-Supervised Visual Representation Learning for Writer Identification},
  author={Maitra, Arkadip and Mitra, Shree and Manna, Siladittya and Bhattacharya, Saumik and Pal, Umapada},
  journal={ACM Transactions on Asian and Low-Resource Language Information Processing},
  volume={24},
  number={7},
  pages={1--17},
  year={2025},
  publisher={ACM New York, NY}
}

@inproceedings{khalaif2025enhanced,
  title={An Enhanced Deep Learning Approach for Writer Identification and Verification Using Corner Detection},
  author={Khalaif, Mays Zeedan and Younis, Muhanad Tahrir},
  booktitle={International Conference on Cybersecurity and Artificial Intelligence Strategies},
  pages={232--245},
  year={2025},
  organization={Springer}
}

@article{IAM_dataset,
  title={The IAM-database: an English sentence database for offline handwriting recognition},
  author={Marti, U-V and Bunke, Horst},
  journal={International journal on document analysis and recognition},
  volume={5},
  pages={39--46},
  year={2002},
  publisher={Springer}
}

@inproceedings{CVL_dataset,
  title={Cvl-database: An off-line database for writer retrieval, writer identification and word spotting},
  author={Kleber, Florian and Fiel, Stefan and Diem, Markus and Sablatnig, Robert},
  booktitle={2013 12th international conference on document analysis and recognition},
  pages={560--564},
  year={2013},
  organization={IEEE}
}

@article{he2015junction,
  title={Junction detection in handwritten documents and its application to writer identification},
  author={He, Sheng and Wiering, Marco and Schomaker, Lambert},
  journal={Pattern Recognition},
  volume={48},
  number={12},
  pages={4036--4048},
  year={2015},
  publisher={Elsevier}
}

@inproceedings{mezghani2012database,
  title={A database for arabic handwritten text image recognition and writer identification},
  author={Mezghani, Anis and Kanoun, Slim and Khemakhem, Maher and El Abed, Haikal},
  booktitle={2012 international conference on frontiers in handwriting recognition},
  pages={399--402},
  year={2012},
  organization={IEEE}
}

@inproceedings{louloudis2013icdar,
  title={ICDAR 2013 competition on writer identification},
  author={Louloudis, Georgios and Gatos, Basilios and Stamatopoulos, Nikolaos and Papandreou, A},
  booktitle={2013 12th International conference on document analysis and recognition},
  pages={1397--1401},
  year={2013},
  organization={IEEE}
}

@article{AHAWP_dataset,
  title={Arabic handwritten alphabets, words and paragraphs per user (AHAWP) dataset},
  author={Khan, Majid Ali},
  journal={Data in Brief},
  volume={41},
  pages={107947},
  year={2022},
  publisher={Elsevier}
}

@article{abushahla2025quantized,
  title={Neural Network Quantization for Microcontrollers: A Comprehensive Survey of Methods, Platforms, and Applications},
  author={Abushahla, Hamza A and Varam, Dara and Panopio, Ariel Justine N and AlHajri, Mohamed I},
  journal={arXiv preprint arXiv:2508.15008},
  year={2025}
}

\clearpage
\appendix
\setcounter{table}{0}
\counterwithin{table}{section}
\setcounter{figure}{0}
\counterwithin{figure}{section}

\makeatletter
\def\@seccntformat#1{\appendixname~\csname the#1\endcsname\sectcounterend\hskip\betweenumberspace}
\makeatother

\onecolumn
\begingroup
\renewcommand{\figurename}{Fig.}%

\section{Dataset Distribution Under Protocol A}\label{app:protocol-a-dist}
\vspace{-20pt}
\begin{figure}[!htbp]
  \centering
  \includegraphics[scale=0.75]{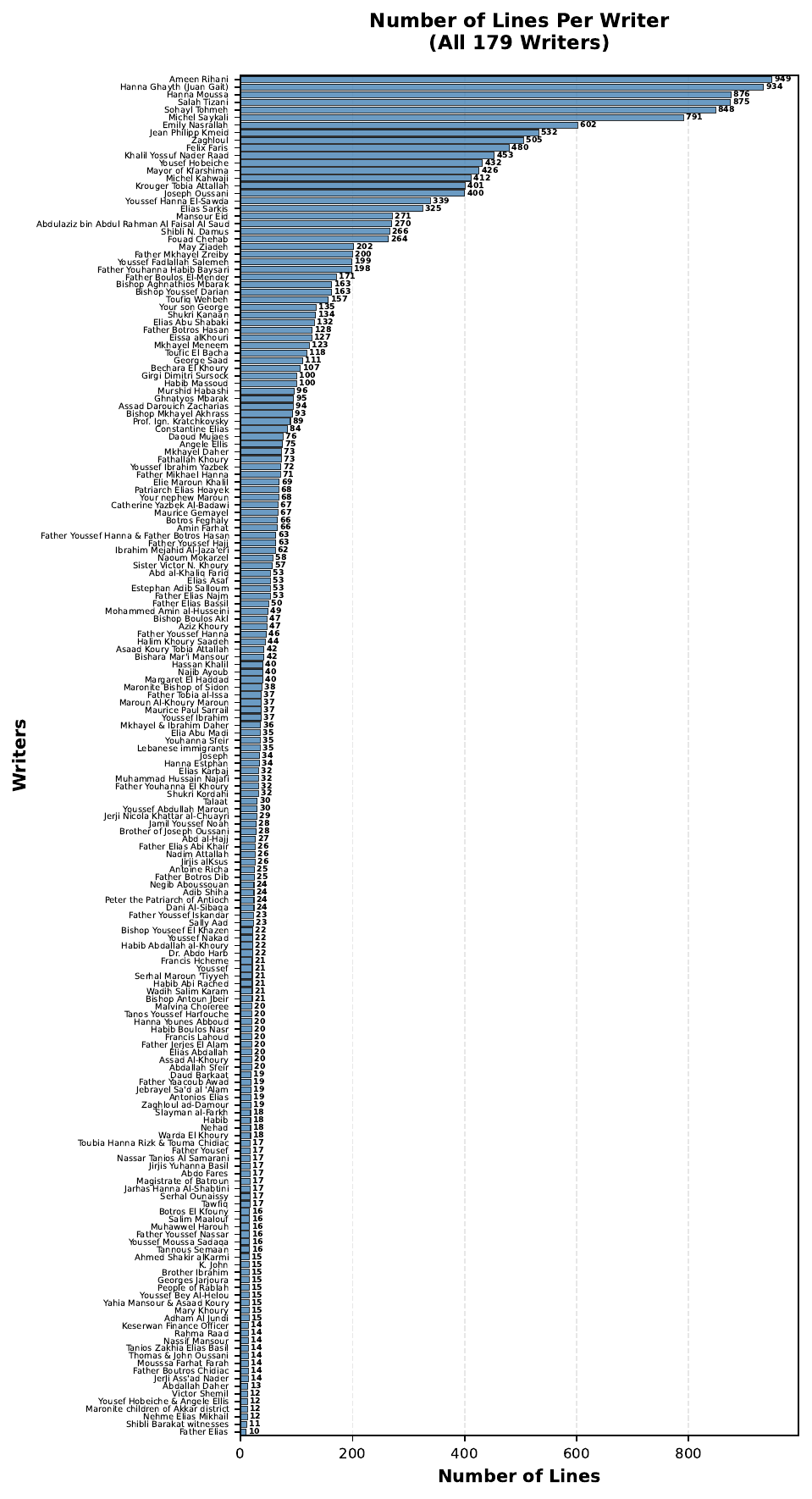}
  \caption{Number of labeled line images per writer under Protocol~A, illustrating the long-tail imbalance of the filtered training set.}
  \label{fig:app:protocol-a-lines}
\end{figure}
\clearpage

\section{Dataset Distribution Under Protocol B}\label{app:protocol-b-dist}
\vspace{-20pt}
\begin{figure}[!htbp]
  \centering
  \includegraphics[scale=0.75]{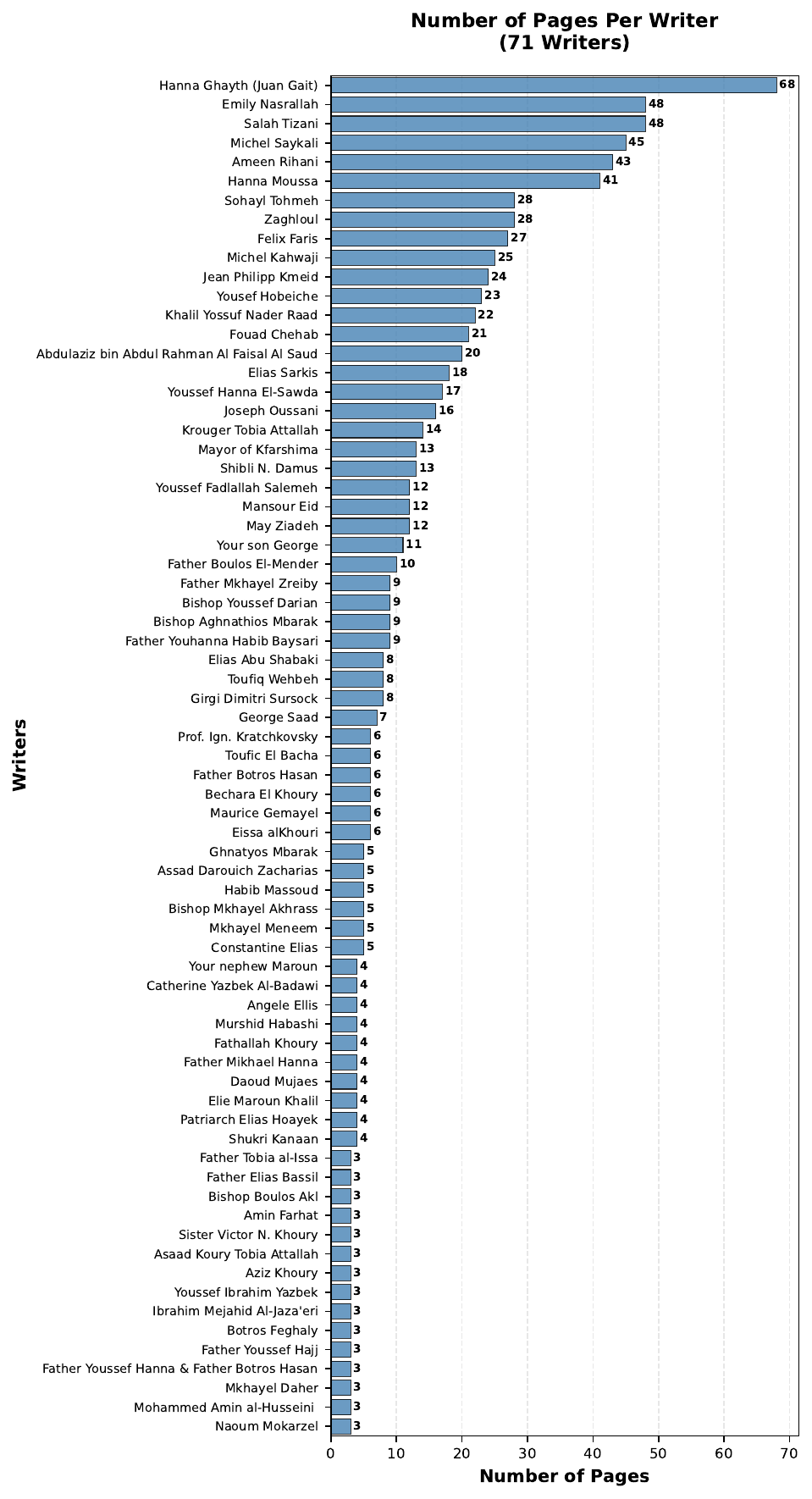}
  \caption{Number of manuscript pages per writer under Protocol~B, illustrating skewed per-writer supports.}
  \label{fig:app:protocol-b-pages}
\end{figure}
\clearpage
\endgroup
\renewcommand{\figurename}{Figure}%

\section{Additional Results}\label{app:additional-results}
\begin{table}[!htbp]
\centering
\caption{Recall Test Results. The \textbf{best result} within Protocol A is \textbf{bolded}.}
\label{tab:app:recall-a}
\footnotesize
\renewcommand{\arraystretch}{1.0}
\setlength{\tabcolsep}{5pt}
\begin{tabular*}{\textwidth}{@{\extracolsep{\fill}}lcccc}
\toprule
\textbf{Model Configuration} & \multicolumn{4}{c}{\textbf{Recall -- Mean (Standard Deviation)}} \\
\cmidrule(lr){2-5}
 & \textbf{ResNet50} & \textbf{DenseNet201} & \textbf{Xception} & \textbf{\MNVL} \\
\midrule
Frozen + No Attention (Baseline)            & 0.6648 (0.0137) & 0.6489 (0.0291) & 0.3147 (0.0232) & 0.3595 (0.0325) \\
Frozen + Attention                           & 0.7792 (0.0140) & 0.7880 (0.0192) & 0.4463 (0.0219) & 0.4978 (0.0028) \\

Fine-tuned + Last Layer + No Attention         & 0.6658 (0.0280) & 0.6605 (0.0141) & 0.3492 (0.0407) & 0.3855 (0.0121) \\
Fine-tuned + Last Layer + Attention            & 0.7966 (0.0072) & 0.7957 (0.0155) & 0.4422 (0.0345) & 0.5327 (0.0228) \\
Fine-tuned + Last 5 Layers + No Attention        & 0.7371 (0.0177) & 0.7090 (0.0335) & 0.6328 (0.0146) & 0.6366 (0.0147) \\
Fine-tuned + Last 5 Layers + Attention           & 0.8472 (0.0085) & 0.8557 (0.0113) & 0.7195 (0.0157) & 0.7064 (0.0089) \\
Fine-tuned + Last 10 Layers + No Attention       & 0.8740 (0.0069) & 0.8101 (0.0152) & 0.8453 (0.0129) & 0.6670 (0.0393) \\
Fine-tuned + Last 10 Layers + Attention          & 0.8757 (0.0200) & 0.8511 (0.0057) & 0.8691 (0.0147) & 0.7524 (0.0181) \\
Fine-tuned + Last 25 Layers + No Attention       & 0.9069 (0.0070) & 0.8362 (0.0240) & 0.8665 (0.0204) & 0.8305 (0.0108) \\
Fine-tuned + Last 25 Layers + Attention          & 0.9178 (0.0100) & 0.8703 (0.0217) & 0.8738 (0.0146) & 0.8292 (0.0004) \\
Fine-tuned + No Attention                      & 0.9348 (0.0232) & 0.9561 (0.0175) & 0.9727 (0.0039) & 0.9595 (0.0094) \\
Fine-tuned + Attention                         & 0.9448 (0.0190) & \textbf{0.9730 (0.0033)} & 0.9697 (0.0041) & 0.9703 (0.0051) \\
From Scratch + No Attention                  & 0.8828 (0.0345) & 0.9042 (0.0121) & 0.9322 (0.0040) & 0.1418 (0.0967) \\
From Scratch + Attention                     & 0.8848 (0.0310) & 0.9030 (0.0231) & 0.9438 (0.0015) & 0.0056 (0.0000) \\
\bottomrule
\end{tabular*}
\end{table}

\vspace{-10pt}

\begin{table}[!htbp]
\centering
\caption{Precision Test Results. The \textbf{best result} within Protocol A is \textbf{bolded}.}
\label{tab:app:precision-a}
\footnotesize
\renewcommand{\arraystretch}{1.0}
\setlength{\tabcolsep}{5pt}
\begin{tabular*}{\textwidth}{@{\extracolsep{\fill}}lcccc}
\toprule
\textbf{Model Configuration} & \multicolumn{4}{c}{\textbf{Precision -- Mean (Standard Deviation)}} \\
\cmidrule(lr){2-5}
 & \textbf{ResNet50} & \textbf{DenseNet201} & \textbf{Xception} & \textbf{\MNVL} \\
\midrule
Frozen + No Attention (Baseline)            & 0.7387 (0.0155) & 0.7128 (0.0236) & 0.3645 (0.0251) & 0.4479 (0.0453) \\
Frozen + Attention                           & 0.8146 (0.0183) & 0.8199 (0.0257) & 0.5017 (0.0236) & 0.5850 (0.0068) \\

Fine-tuned + Last Layer + No Attention         & 0.7265 (0.0254) & 0.7293 (0.0106) & 0.4017 (0.0464) & 0.4800 (0.0068) \\
Fine-tuned + Last Layer + Attention            & 0.8350 (0.0086) & 0.8352 (0.0085) & 0.5025 (0.0449) & 0.6166 (0.0192) \\
Fine-tuned + Last 5 Layers + No Attention        & 0.7867 (0.0281) & 0.7759 (0.0264) & 0.7032 (0.0072) & 0.7107 (0.0179) \\
Fine-tuned + Last 5 Layers + Attention           & 0.8811 (0.0159) & 0.8819 (0.0131) & 0.7602 (0.0152) & 0.7661 (0.0051) \\
Fine-tuned + Last 10 Layers + No Attention       & 0.8984 (0.0104) & 0.8400 (0.0207) & 0.8781 (0.0151) & 0.7339 (0.0428) \\
Fine-tuned + Last 10 Layers + Attention          & 0.9025 (0.0093) & 0.8753 (0.0029) & 0.8977 (0.0099) & 0.8129 (0.0166) \\
Fine-tuned + Last 25 Layers + No Attention       & 0.9233 (0.0027) & 0.8659 (0.0135) & 0.8948 (0.0274) & 0.8526 (0.0189) \\
Fine-tuned + Last 25 Layers + Attention          & 0.9352 (0.0087) & 0.8897 (0.0193) & 0.8990 (0.0205) & 0.8463 (0.0040) \\
Fine-tuned + No Attention                    & 0.9378 (0.0294) & 0.9638 (0.0168) & 0.9739 (0.0036) & 0.9631 (0.0094) \\
Fine-tuned + Attention                       & 0.9500 (0.0213) & \textbf{0.9819 (0.0032)} & 0.9803 (0.0010) & 0.9765 (0.0038) \\
From Scratch + No Attention                  & 0.9006 (0.0323) & 0.9242 (0.0120) & 0.9503 (0.0053) & 0.1141 (0.0808) \\
From Scratch + Attention                     & 0.9058 (0.0327) & 0.9298 (0.0125) & 0.9606 (0.0041) & 0.0003 (0.0000) \\
\bottomrule
\end{tabular*}
\end{table}

\vspace{-10pt}

\begin{table}[!htbp]
\centering
\caption{Test Loss Results. The \textbf{best result} within Protocol A is \textbf{bolded}.}
\label{tab:app:loss-a}
\footnotesize
\renewcommand{\arraystretch}{1.0}
\setlength{\tabcolsep}{5pt}
\begin{tabular*}{\textwidth}{@{\extracolsep{\fill}}lcccc}
\toprule
\textbf{Model Configuration} & \multicolumn{4}{c}{\textbf{Test Loss -- Mean (Standard Deviation)}} \\
\cmidrule(lr){2-5}
 & \textbf{ResNet50} & \textbf{DenseNet201} & \textbf{Xception} & \textbf{\MNVL} \\
\midrule
Frozen + No Attention (Baseline)            & 0.5517 (0.0088) & 0.5415 (0.0235) & 1.2822 (0.0502) & 1.1310 (0.0181) \\
Frozen + Attention                           & 0.3995 (0.0145) & 0.3747 (0.0138) & 1.0810 (0.0227) & 0.9310 (0.0429) \\

Fine-tuned + Last Layer + No Attention         & 0.5230 (0.0072) & 0.5379 (0.0225) & 1.2475 (0.0528) & 1.0789 (0.0143) \\
Fine-tuned + Last Layer + Attention            & 0.3998 (0.0178) & 0.3735 (0.0155) & 1.1140 (0.0109) & 0.8551 (0.0403) \\
Fine-tuned + Last 5 Layers + No Attention        & 0.4553 (0.0024) & 0.4321 (0.0246) & 0.7992 (0.0418) & 0.6212 (0.0149) \\
Fine-tuned + Last 5 Layers + Attention           & 0.3364 (0.0196) & 0.3125 (0.0309) & 0.7460 (0.0176) & 0.5684 (0.0215) \\
Fine-tuned + Last 10 Layers + No Attention       & 0.3037 (0.0072) & 0.3056 (0.0097) & 0.3988 (0.0433) & 0.5769 (0.0171) \\
Fine-tuned + Last 10 Layers + Attention          & 0.3377 (0.0294) & 0.2786 (0.0150) & 0.3691 (0.0195) & 0.5454 (0.0218) \\
Fine-tuned + Last 25 Layers + No Attention       & 0.2779 (0.0063) & 0.3093 (0.0227) & 0.3443 (0.0130) & 0.5018 (0.0452) \\
Fine-tuned + Last 25 Layers + Attention          & 0.2489 (0.0288) & 0.2784 (0.0264) & 0.3497 (0.0163) & 0.5051 (0.0342) \\
Fine-tuned + No Attention                      & 0.1161 (0.0147) & 0.0876 (0.0135) & 0.0958 (0.0134) & 0.1046 (0.0186) \\
Fine-tuned + Attention                         & 0.1052 (0.0226) & \textbf{0.0694 (0.0061)} & 0.0902 (0.0035) & 0.0947 (0.0067) \\
From Scratch + No Attention                  & 0.1597 (0.0132) & 0.1455 (0.0056) & 0.1119 (0.0093) & 2.2607 (1.5662) \\
From Scratch + Attention                     & 0.1613 (0.0231) & 0.1600 (0.0123) & 0.1386 (0.0068) & 4.6212 (0.1239) \\
\bottomrule
\end{tabular*}
\end{table}

\begin{table}[!htbp]
\centering
\caption{Recall Test Results. The \textbf{best result} within Protocol B is \textbf{bolded}.}
\label{tab:app:recall-b}
\footnotesize
\renewcommand{\arraystretch}{1.0}
\setlength{\tabcolsep}{5pt}
\begin{tabular*}{\textwidth}{@{\extracolsep{\fill}}lcccc}
\toprule
\textbf{Model Configuration} & \multicolumn{4}{c}{\textbf{Recall -- Mean (Standard Deviation)}} \\
\cmidrule(lr){2-5}
 & \textbf{ResNet50} & \textbf{DenseNet201} & \textbf{Xception} & \textbf{\MNVL} \\
\midrule
Frozen + No Attention (Baseline)            & 0.5949 (0.0183) & 0.5530 (0.0199) & 0.3827 (0.0295) & 0.4839 (0.0078) \\
Frozen + Attention                         & 0.6046 (0.0289) & 0.6194 (0.0388) & 0.4456 (0.0471) & 0.5228 (0.0401) \\

Fine-tuned + Last Layer + No Attention     & 0.5679 (0.0164) & 0.5578 (0.0181) & 0.3752 (0.0407) & 0.4782 (0.0409) \\
Fine-tuned + Last Layer + Attention        & 0.6180 (0.0251) & 0.6089 (0.0510) & 0.4265 (0.0373) & 0.5234 (0.0417) \\
Fine-tuned + Last 5 Layers + No Attention  & 0.6128 (0.0193) & 0.5273 (0.0809) & 0.5334 (0.0433) & 0.5592 (0.0325) \\
Fine-tuned + Last 5 Layers + Attention     & 0.6235 (0.0275) & 0.5420 (0.1681) & 0.5248 (0.0877) & 0.5611 (0.1007) \\
Fine-tuned + Last 10 Layers + No Attention & 0.6345 (0.0278) & 0.6034 (0.0667) & 0.6438 (0.0181) & 0.5412 (0.0829) \\
Fine-tuned + Last 10 Layers + Attention    & 0.6388 (0.0196) & 0.5794 (0.1089) & 0.6164 (0.0549) & 0.5381 (0.0889) \\
Fine-tuned + Last 25 Layers + No Attention & 0.6783 (0.0155) & 0.6383 (0.0486) & 0.6780 (0.0031) & 0.6179 (0.0311) \\
Fine-tuned + Last 25 Layers + Attention    & 0.6787 (0.0144) & 0.6170 (0.0758) & 0.6688 (0.0107) & 0.6445 (0.0266) \\
Fine-tuned + No Attention                  & 0.6438 (0.0138) & 0.6333 (0.0102) & 0.6816 (0.0282) & \textbf{0.6885 (0.0108)} \\
Fine-tuned + Attention                     & 0.6333 (0.0294) & 0.6832 (0.0089) & 0.6777 (0.0125) & 0.6865 (0.0135) \\
From Scratch + No Attention                & 0.5381 (0.0435) & 0.5438 (0.0224) & 0.5780 (0.0926) & 0.0141 (0.0000) \\
From Scratch + Attention                   & 0.5567 (0.0381) & 0.5376 (0.0290) & 0.5627 (0.0954) & 0.0141 (0.0000) \\
\bottomrule
\end{tabular*}
\end{table}

\vspace{-10pt}

\begin{table}[!htbp]
\centering
\caption{Precision Test Results. The \textbf{best result} within Protocol B is \textbf{bolded}.}
\label{tab:app:precision-b}
\footnotesize
\renewcommand{\arraystretch}{1.0}
\setlength{\tabcolsep}{5pt}
\begin{tabular*}{\textwidth}{@{\extracolsep{\fill}}lcccc}
\toprule
\textbf{Model Configuration} & \multicolumn{4}{c}{\textbf{Precision -- Mean (Standard Deviation)}} \\
\cmidrule(lr){2-5}
 & \textbf{ResNet50} & \textbf{DenseNet201} & \textbf{Xception} & \textbf{\MNVL} \\
\midrule
Frozen + No Attention (Baseline)            & 0.5484 (0.0312) & 0.5139 (0.0403) & 0.3355 (0.0205) & 0.4606 (0.0239) \\
Frozen + Attention                         & 0.5700 (0.0307) & 0.5796 (0.0238) & 0.4094 (0.0334) & 0.4984 (0.0421) \\

Fine-tuned + Last Layer + No Attention     & 0.5267 (0.0387) & 0.5165 (0.0089) & 0.3325 (0.0476) & 0.4482 (0.0492) \\
Fine-tuned + Last Layer + Attention        & 0.5867 (0.0355) & 0.5695 (0.0521) & 0.3720 (0.0344) & 0.5021 (0.0465) \\
Fine-tuned + Last 5 Layers + No Attention  & 0.5761 (0.0281) & 0.4895 (0.0994) & 0.4812 (0.0459) & 0.5077 (0.0322) \\
Fine-tuned + Last 5 Layers + Attention     & 0.5789 (0.0364) & 0.4870 (0.1801) & 0.4719 (0.0900) & 0.5024 (0.1058) \\
Fine-tuned + Last 10 Layers + No Attention & 0.5841 (0.0330) & 0.5558 (0.0698) & 0.6148 (0.0124) & 0.4855 (0.0924) \\
Fine-tuned + Last 10 Layers + Attention    & 0.5788 (0.0256) & 0.5140 (0.1259) & 0.5627 (0.0670) & 0.4808 (0.0917) \\
Fine-tuned + Last 25 Layers + No Attention & 0.6244 (0.0124) & 0.5759 (0.0623) & 0.6422 (0.0076) & 0.5670 (0.0470) \\
Fine-tuned + Last 25 Layers + Attention    & \textbf{0.6474 (0.0105)} & 0.5611 (0.0941) & 0.6332 (0.0070) & 0.5979 (0.0142) \\
Fine-tuned + No Attention                  & 0.5754 (0.0402) & 0.5694 (0.0432) & 0.6347 (0.0145) & 0.6238 (0.0291) \\
Fine-tuned + Attention                     & 0.5726 (0.0590) & 0.6235 (0.0358) & 0.6142 (0.0198) & 0.6460 (0.0065) \\
From Scratch + No Attention                & 0.4218 (0.0719) & 0.4452 (0.0290) & 0.4974 (0.1076) & 0.0008 (0.0007) \\
From Scratch + Attention                   & 0.4747 (0.0705) & 0.4425 (0.0484) & 0.4644 (0.1298) & 0.0002 (0.0001) \\
\bottomrule
\end{tabular*}
\end{table}

\vspace{-10pt}

\begin{table}[!htbp]
\centering
\caption{Test Loss Results. The \textbf{best result} within Protocol B is \textbf{bolded}.}
\label{tab:app:loss-b}
\footnotesize
\renewcommand{\arraystretch}{1.0}
\setlength{\tabcolsep}{5pt}
\begin{tabular*}{\textwidth}{@{\extracolsep{\fill}}lcccc}
\toprule
\textbf{Model Configuration} & \multicolumn{4}{c}{\textbf{Test Loss -- Mean (Standard Deviation)}} \\
\cmidrule(lr){2-5}
 & \textbf{ResNet50} & \textbf{DenseNet201} & \textbf{Xception} & \textbf{\MNVL} \\
\midrule
Frozen + No Attention (Baseline)            & 1.5836 (0.0897) & 1.6863 (0.0748) & 2.4914 (0.0505) & 2.0683 (0.1450) \\
Frozen + Attention                         & 1.6172 (0.1343) & 1.7234 (0.0900) & 2.4198 (0.0357) & 2.1090 (0.1062) \\

Fine-tuned + Last Layer + No Attention     & 1.6338 (0.0890) & 1.6851 (0.1179) & 2.5855 (0.0939) & 2.0822 (0.1324) \\
Fine-tuned + Last Layer + Attention        & 1.6498 (0.1410) & 1.6924 (0.0744) & 2.4164 (0.0172) & 2.0582 (0.1078) \\
Fine-tuned + Last 5 Layers + No Attention  & 1.7123 (0.1504) & 1.7525 (0.0914) & 2.2157 (0.0450) & 1.9285 (0.1250) \\
Fine-tuned + Last 5 Layers + Attention     & 1.6055 (0.0387) & 1.9828 (0.3119) & 2.2562 (0.0729) & 2.1362 (0.0703) \\
Fine-tuned + Last 10 Layers + No Attention & 1.8789 (0.1659) & 1.7603 (0.1170) & 2.1344 (0.2775) & 2.1522 (0.2272) \\
Fine-tuned + Last 10 Layers + Attention    & 1.7746 (0.1674) & 1.8089 (0.1878) & 2.0095 (0.1133) & 2.0209 (0.0517) \\
Fine-tuned + Last 25 Layers + No Attention & 1.9069 (0.2826) & 1.6937 (0.1449) & 2.0380 (0.2488) & 2.1044 (0.3570) \\
Fine-tuned + Last 25 Layers + Attention    & 1.7816 (0.1589) & 1.7279 (0.1172) & 1.9592 (0.1395) & 2.2494 (0.2569) \\
Fine-tuned + No Attention                  & 1.7742 (0.0613) & 1.7317 (0.1221) & 2.0275 (0.0528) & \textbf{1.5808 (0.1075)} \\
Fine-tuned + Attention                     & 1.7843 (0.1066) & 1.7852 (0.1248) & 1.9128 (0.2054) & 1.7630 (0.0797) \\
From Scratch + No Attention                & 1.9663 (0.1899) & 1.9407 (0.1358) & 1.9831 (0.1813) & 4.3754 (0.0656) \\
From Scratch + Attention                   & 2.1376 (0.2081) & 2.0616 (0.1529) & 2.2160 (0.0523) & 4.4023 (0.1125) \\
\bottomrule
\end{tabular*}
\end{table}

\clearpage

\section{Classification Reports}\label{app:classification-reports}

\noindent Tables~\ref{tab:app:crA-01}--\ref{tab:app:crA-10} present partial classification reports for selected configurations under \textbf{Protocol A}, including the best-performing model (DenseNet201 + Fine-tuned + Attention) and the best baseline model (ResNet50 + Frozen + No Attention). We additionally include Table~\ref{tab:app:crB-mbv3}, which presents a partial classification report for the best-performing model under \textbf{Protocol B} (\MNVL + Fine-tuned + Attention). Across all tables, we \textbf{bold} specific rows to emphasize key results, with ellipses (\ldots) indicating that additional rows exist before and after the displayed subset.

\begin{table}[ht]
\caption{Classification Report for DenseNet201 + Fine-tuned + No Attention (Protocol A)}
\label{tab:app:crA-01}
\footnotesize
\renewcommand{\arraystretch}{1.3} 
\centering
\setlength{\tabcolsep}{5pt} 
\begin{tabular*}{\columnwidth}{@{\extracolsep{\fill}}lcccc} 
\toprule 
\textbf{Class Name} & \multicolumn{4}{c}{\textbf{Mean (Standard Deviation)}} \\
\cmidrule(lr){2-5}
 & \textbf{Precision} & \textbf{Recall} & \textbf{F1-Score} & \textbf{Support} \\
\midrule 
\vdots & \vdots & \vdots & \vdots & \vdots \\ 
\textbf{Yousef Hobeiche} & \textbf{1.0000 (0.0000)} & \textbf{0.9949 (0.0089)} & \textbf{0.9974 (0.0045)} & \textbf{65.0000 (0.0000)} \\
\textbf{Angele Ellis} & \textbf{0.9744 (0.0444)} & \textbf{1.0000 (0.0000)} & \textbf{0.9867 (0.0231)} & \textbf{12.0000 (0.0000)} \\
\textbf{Yousef Hobeiche \& Angele Ellis} & \textbf{0.6667 (0.5774)} & \textbf{0.6667 (0.5774)} & \textbf{0.6667 (0.5774)} & \textbf{2.0000 (0.0000)} \\
\vdots & \vdots & \vdots & \vdots & \vdots \\ 
Elias Abdallah & 1.0000 (0.0000) & 1.0000 (0.0000) & 1.0000 (0.0000) & 3.0000 (0.0000) \\
Youssef Bey Al-Helou  & 1.0000 (0.0000) & 0.8333 (0.2887) & 0.8889 (0.1925) & 2.0000 (0.0000) \\
Father Elias  & 0.6667 (0.5774) & 0.6667 (0.5774) & 0.6667 (0.5774) & 1.0000 (0.0000) \\
Magistrate of Batroun  & 1.0000 (0.0000) & 1.0000 (0.0000) & 1.0000 (0.0000) & 2.0000 (0.0000) \\
\vdots & \vdots & \vdots & \vdots & \vdots \\ 
\hline
accuracy & 0.9862 (0.0035) & 0.9862 (0.0035) & 0.9862 (0.0035) & 0.9862 (0.0035) \\
macro avg & 0.9638 (0.0205) & 0.9561 (0.0215) & 0.9557 (0.0226) & 2849.0000 (0.0000) \\
weighted avg & 0.9863 (0.0041) & 0.9862 (0.0035) & 0.9852 (0.0041) & 2849.0000 (0.0000) \\
\bottomrule 
\end{tabular*}
\end{table}

\begin{table}[ht]
\caption{Classification Report for the Best Model (DenseNet201 + Fine-tuned + Attention) (Protocol A)}
\label{tab:app:crA-02}
\footnotesize
\renewcommand{\arraystretch}{1.3} 
\centering
\setlength{\tabcolsep}{5pt} 
\begin{tabular*}{\columnwidth}{@{\extracolsep{\fill}}lcccc} 
\toprule 
\textbf{Class Name} & \multicolumn{4}{c}{\textbf{Mean (Standard Deviation)}} \\
\cmidrule(lr){2-5}
 & \textbf{Precision} & \textbf{Recall} & \textbf{F1-Score} & \textbf{Support} \\
\midrule 
\vdots & \vdots & \vdots & \vdots & \vdots \\ 
\textbf{Yousef Hobeiche} & \textbf{1.0000 (0.0000)} & \textbf{0.9949 (0.0089)} & \textbf{0.9974 (0.0045)} & \textbf{65.0000 (0.0000)} \\
\textbf{Angele Ellis} & \textbf{1.0000 (0.0000)} & \textbf{1.0000 (0.0000)} & \textbf{1.0000 (0.0000)} & \textbf{12.0000 (0.0000)} \\
\textbf{Yousef Hobeiche \& Angele Ellis} & \textbf{1.0000 (0.0000)} & \textbf{1.0000 (0.0000)} & \textbf{1.0000 (0.0000)} & \textbf{2.0000 (0.0000)} \\
\vdots & \vdots & \vdots & \vdots & \vdots \\ 
Elias Abdallah & 0.9167 (0.1443) & 1.0000 (0.0000) & 0.9524 (0.0825) & 3.0000 (0.0000) \\
Youssef Bey Al-Helou & 1.0000 (0.0000) & 0.8333 (0.2887) & 0.8889 (0.1925) & 2.0000 (0.0000) \\
Father Elias & 0.3333 (0.5774) & 0.3333 (0.5774) & 0.3333 (0.5774) & 1.0000 (0.0000) \\
Magistrate of Batroun & 1.0000 (0.0000) & 1.0000 (0.0000) & 1.0000 (0.0000) & 2.0000 (0.0000) \\
\vdots & \vdots & \vdots & \vdots & \vdots \\ 
\hline
accuracy & 0.9905 (0.0009) & 0.9905 (0.0009) & 0.9905 (0.0009) & 0.9905 (0.0009) \\
macro avg & 0.9819 (0.0039) & 0.9730 (0.0040) & 0.9744 (0.0028) & 2849.0000 (0.0000) \\
weighted avg & 0.9912 (0.0006) & 0.9905 (0.0009) & 0.9902 (0.0007) & 2849.0000 (0.0000) \\
\bottomrule 
\end{tabular*}
\end{table}

\clearpage

\begin{table}[ht]
\caption{Classification Report for the Best Baseline Model (ResNet50 + Frozen + No Attention) (Protocol A)}
\label{tab:app:crA-03}
\footnotesize
\renewcommand{\arraystretch}{1.3} 
\centering
\setlength{\tabcolsep}{5pt} 
\begin{tabular*}{\columnwidth}{@{\extracolsep{\fill}}lcccc}
\toprule
\textbf{Class Name} & \multicolumn{4}{c}{\textbf{Mean (Standard Deviation)}} \\
\cmidrule(lr){2-5}
 & \textbf{Precision} & \textbf{Recall} & \textbf{F1-Score} & \textbf{Support} \\
\midrule
\vdots & \vdots & \vdots & \vdots & \vdots \\ 
\textbf{Yousef Hobeiche} & \textbf{0.8490 (0.0410)} & \textbf{0.9128 (0.0622)} & \textbf{0.8787 (0.0355)} & \textbf{65.0000 (0.0000)} \\
\textbf{Angele Ellis} & \textbf{0.5541 (0.0495)} & \textbf{0.8889 (0.1273)} & \textbf{0.6799 (0.0565)} & \textbf{12.0000 (0.0000)} \\
\textbf{Yousef Hobeiche \& Angele Ellis} & \textbf{0.1667 (0.2887)} & \textbf{0.1667 (0.2887)} & \textbf{0.1667 (0.2887)} & \textbf{2.0000 (0.0000)} \\
\vdots & \vdots & \vdots & \vdots & \vdots \\
Elias Abdallah & 0.3333 (0.5774) & 0.1111 (0.1925) & 0.1667 (0.2887) & 3.0000 (0.0000) \\
Youssef Bey Al-Helou  & 0.0000 (0.0000) & 0.0000 (0.0000) & 0.0000 (0.0000) & 2.0000 (0.0000) \\
Father Elias  & 0.3333 (0.5774) & 0.3333 (0.5774) & 0.3333 (0.5774) & 1.0000 (0.0000) \\
Magistrate of Batroun  & 0.0000 (0.0000) & 0.0000 (0.0000) & 0.0000 (0.0000) & 2.0000 (0.0000) \\
\vdots & \vdots & \vdots & \vdots & \vdots \\
\hline
accuracy & 0.8652 (0.0035) & 0.8652 (0.0035) & 0.8652 (0.0035) & 0.8652 (0.0035) \\
macro avg & 0.7387 (0.0190) & 0.6648 (0.0168) & 0.6772 (0.0162) & 2849.0000 (0.0000) \\
weighted avg & 0.8701 (0.0079) & 0.8652 (0.0035) & 0.8572 (0.0052) & 2849.0000 (0.0000) \\
\bottomrule 
\end{tabular*}
\end{table}

\noindent The classification report for the best baseline model shows that the model struggles with classes having few samples, yielding very low precision, recall, and F1-scores. For example, "Youssef Bey Al-Helou," "Father Elias," "Magistrate of Batroun," and "Elias Abdallah" suffer from limited samples and sparse training representation. This data scarcity likely prevents the model from learning meaningful writer-specific patterns, resulting in poor generalization and misidentification during evaluation.

\begin{table}[htbp]
\caption{Classification Report for ResNet50 + Frozen + Attention (Protocol A)}
\label{tab:app:crA-04}
\footnotesize
\renewcommand{\arraystretch}{1.3} 
\centering
\setlength{\tabcolsep}{5pt} 
\begin{tabular*}{\columnwidth}{@{\extracolsep{\fill}}lcccc} 
\toprule
\textbf{Class Name} & \multicolumn{4}{c}{\textbf{Mean (Standard Deviation)}} \\
\cmidrule(lr){2-5}
 & \textbf{Precision} & \textbf{Recall} & \textbf{F1-Score} & \textbf{Support} \\
\midrule
\vdots & \vdots & \vdots & \vdots & \vdots \\ 
\textbf{Yousef Hobeiche} & \textbf{0.9393 (0.0266)} & \textbf{0.9538 (0.0533)} & \textbf{0.9462 (0.0363)} & \textbf{65.0000 (0.0000)} \\
\textbf{Angele Ellis} & \textbf{0.7075 (0.0417)} & \textbf{1.0000 (0.0000)} & \textbf{0.8282 (0.0286)} & \textbf{12.0000 (0.0000)} \\
\textbf{Yousef Hobeiche \& Angele Ellis} & \textbf{0.0000 (0.0000)} & \textbf{0.0000 (0.0000)} & \textbf{0.0000 (0.0000)} & \textbf{2.0000 (0.0000)} \\
\vdots & \vdots & \vdots & \vdots & \vdots \\ 
Elias Abdallah & 1.0000 (0.0000) & 1.0000 (0.0000) & 1.0000 (0.0000) & 3.0000 (0.0000) \\
Youssef Bey Al-Helou  & 0.3333 (0.5774) & 0.3333 (0.5774) & 0.3333 (0.5774) & 2.0000 (0.0000) \\
Father Elias  & 0.3333 (0.5774) & 0.3333 (0.5774) & 0.3333 (0.5774) & 1.0000 (0.0000) \\
Magistrate of Batroun  & 0.0000 (0.0000) & 0.0000 (0.0000) & 0.0000 (0.0000) & 2.0000 (0.0000) \\
\vdots & \vdots & \vdots & \vdots & \vdots \\ 
\hline
accuracy & 0.9114 (0.0050) & 0.9114 (0.0050) & 0.9114 (0.0050) & 0.9114 (0.0050) \\
macro avg & 0.8146 (0.0225) & 0.7792 (0.0171) & 0.7795 (0.0169) & 2849.0000 (0.0000) \\
weighted avg & 0.9135 (0.0066) & 0.9114 (0.0050) & 0.9069 (0.0043) & 2849.0000 (0.0000) \\
\bottomrule 
\end{tabular*}
\end{table}

\begin{table}[htbp]
\caption{Classification Report for DenseNet201 Fine-tuned + Last 25 Layers + No Attention (Protocol A)}
\label{tab:app:crA-05}
\footnotesize
\renewcommand{\arraystretch}{1.3} 
\centering
\setlength{\tabcolsep}{5pt} 
\begin{tabular*}{\columnwidth}{@{\extracolsep{\fill}}lcccc} 
\toprule 
\textbf{Class Name} & \multicolumn{4}{c}{\textbf{Mean (Standard Deviation)}} \\
\cmidrule(lr){2-5}
 & \textbf{Precision} & \textbf{Recall} & \textbf{F1-Score} & \textbf{Support} \\
\midrule 
\vdots & \vdots & \vdots & \vdots & \vdots \\ 
\textbf{Yousef Hobeiche} & \textbf{0.9800 (0.0228)} & \textbf{0.9846 (0.0154)} & \textbf{0.9821 (0.0115)} & \textbf{65.0000 (0.0000)} \\
\textbf{Angele Ellis} & \textbf{0.8654 (0.0449)} & \textbf{0.8889 (0.0481)} & \textbf{0.8767 (0.0418)} & \textbf{12.0000 (0.0000)} \\
\textbf{Yousef Hobeiche \& Angele Ellis} & \textbf{0.6667 (0.5774)} & \textbf{0.3333 (0.2887)} & \textbf{0.4444 (0.3849)} & \textbf{2.0000 (0.0000)} \\
\vdots & \vdots & \vdots & \vdots & \vdots \\ 
Elias Abdallah & 1.0000 (0.0000) & 0.4444 (0.1925) & 0.6000 (0.1732) & 3.0000 (0.0000) \\
Youssef Bey Al-Helou & 0.6667 (0.5774) & 0.6667 (0.5774) & 0.6667 (0.5774) & 2.0000 (0.0000) \\
Father Elias & 0.0000 (0.0000) & 0.0000 (0.0000) & 0.0000 (0.0000) & 1.0000 (0.0000) \\
Magistrate of Batroun & 0.3333 (0.5774) & 0.3333 (0.5774) & 0.3333 (0.5774) & 2.0000 (0.0000) \\
\vdots & \vdots & \vdots & \vdots & \vdots \\ 
\hline
accuracy & 0.9404 (0.0093) & 0.9404 (0.0093) & 0.9404 (0.0093) & 0.9404 (0.0093) \\
macro avg & 0.8659 (0.0165) & 0.8362 (0.0293) & 0.8353 (0.0230) & 2849.0000 (0.0000) \\
weighted avg & 0.9430 (0.0085) & 0.9404 (0.0093) & 0.9375 (0.0096) & 2849.0000 (0.0000) \\
\bottomrule 
\end{tabular*}
\end{table}

\FloatBarrier

\begin{table}[htbp]
\caption{Classification Report for DenseNet201 Fine-tuned + Last 25 Layers + Attention (Protocol A)}
\label{tab:app:crA-06}
\footnotesize
\renewcommand{\arraystretch}{1.3} 
\centering
\setlength{\tabcolsep}{5pt} 
\begin{tabular*}{\columnwidth}{@{\extracolsep{\fill}}lcccc} 
\toprule 
\textbf{Class Name} & \multicolumn{4}{c}{\textbf{Mean (Standard Deviation)}} \\
\cmidrule(lr){2-5}
 & \textbf{Precision} & \textbf{Recall} & \textbf{F1-Score} & \textbf{Support} \\
\midrule 
\vdots & \vdots & \vdots & \vdots & \vdots \\ 
\textbf{Yousef Hobeiche} & \textbf{0.9750 (0.0228)} & \textbf{0.9897 (0.0089)} & \textbf{0.9823 (0.0158)} & \textbf{65.0000 (0.0000)} \\
\textbf{Angele Ellis} & \textbf{0.7951 (0.0776)} & \textbf{0.9167 (0.0833)} & \textbf{0.8470 (0.0287)} & \textbf{12.0000 (0.0000)} \\
\textbf{Yousef Hobeiche \& Angele Ellis} & \textbf{0.5000 (0.5000)} & \textbf{0.3333 (0.2887)} & \textbf{0.3889 (0.3469)} & \textbf{2.0000 (0.0000)} \\
\vdots & \vdots & \vdots & \vdots & \vdots \\ 
Elias Abdallah & 0.8000 (0.3464) & 0.6667 (0.0000) & 0.7000 (0.1732) & 3.0000 (0.0000) \\
Youssef Bey Al-Helou & 0.6667 (0.3333) & 0.6667 (0.2887) & 0.6222 (0.2037) & 2.0000 (0.0000) \\
Father Elias & 0.3333 (0.5774) & 0.3333 (0.5774) & 0.3333 (0.5774) & 1.0000 (0.0000) \\
Magistrate of Batroun & 0.0000 (0.0000) & 0.0000 (0.0000) & 0.0000 (0.0000) & 2.0000 (0.0000) \\
\vdots & \vdots & \vdots & \vdots & \vdots \\ 
\hline
accuracy & 0.9479 (0.0112) & 0.9479 (0.0112) & 0.9479 (0.0112) & 0.9479 (0.0112) \\
macro avg & 0.8897 (0.0237) & 0.8703 (0.0266) & 0.8659 (0.0272) & 2849.0000 (0.0000) \\
weighted avg & 0.9520 (0.0106) & 0.9479 (0.0112) & 0.9462 (0.0116) & 2849.0000 (0.0000) \\
\bottomrule 
\end{tabular*}
\end{table}

\begin{table}[htbp]
\caption{Classification Report for DenseNet201 + Frozen + No Attention (Protocol A)}
\label{tab:app:crA-07}
\footnotesize
\renewcommand{\arraystretch}{1.3} 
\centering
\setlength{\tabcolsep}{5pt} 
\begin{tabular*}{\columnwidth}{@{\extracolsep{\fill}}lcccc} 
\toprule 
\textbf{Class Name} & \multicolumn{4}{c}{\textbf{Mean (Standard Deviation)}} \\
\cmidrule(lr){2-5}
 & \textbf{Precision} & \textbf{Recall} & \textbf{F1-Score} & \textbf{Support} \\
\midrule 
\vdots & \vdots & \vdots & \vdots & \vdots \\ 
\textbf{Yousef Hobeiche} & \textbf{0.9233 (0.0281)} & \textbf{0.9795 (0.0178)} & \textbf{0.9503 (0.0150)} & \textbf{65.0000 (0.0000)} \\
\textbf{Angele Ellis} & \textbf{0.6825 (0.1689)} & \textbf{0.7222 (0.1925)} & \textbf{0.7009 (0.1769)} & \textbf{12.0000 (0.0000)} \\
\textbf{Yousef Hobeiche \& Angele Ellis} & \textbf{0.0000 (0.0000)} & \textbf{0.0000 (0.0000)} & \textbf{0.0000 (0.0000)} & \textbf{2.0000 (0.0000)} \\
\vdots & \vdots & \vdots & \vdots & \vdots \\ 
Elias Abdallah & 1.0000 (0.0000) & 0.5556 (0.1925) & 0.7000 (0.1732) & 3.0000 (0.0000) \\
Youssef Bey Al-Helou & 0.3333 (0.5774) & 0.3333 (0.5774) & 0.3333 (0.5774) & 2.0000 (0.0000) \\
Father Elias & 0.0000 (0.0000) & 0.0000 (0.0000) & 0.0000 (0.0000) & 1.0000 (0.0000) \\
Magistrate of Batroun & 0.0000 (0.0000) & 0.0000 (0.0000) & 0.0000 (0.0000) & 2.0000 (0.0000) \\
\vdots & \vdots & \vdots & \vdots & \vdots \\ 
\hline
accuracy & 0.8705 (0.0043) & 0.8705 (0.0043) & 0.8705 (0.0043) & 0.8705 (0.0043) \\
macro avg & 0.6346 (0.1136) & 0.6179 (0.1043) & 0.6092 (0.1060) & 2849.0000 (0.0000) \\
weighted avg & 0.8660 (0.0062) & 0.8705 (0.0043) & 0.8644 (0.0064) & 2849.0000 (0.0000) \\
\bottomrule 
\end{tabular*}
\end{table}

\FloatBarrier

\begin{table}[htbp]
\caption{Classification Report for DenseNet201 + Frozen + Attention (Protocol A)}
\label{tab:app:crA-08}
\footnotesize
\renewcommand{\arraystretch}{1.3} 
\centering
\setlength{\tabcolsep}{5pt} 
\begin{tabular*}{\columnwidth}{@{\extracolsep{\fill}}lcccc}
\toprule
\textbf{Class Name} & \multicolumn{4}{c}{\textbf{Mean (Standard Deviation)}} \\
\cmidrule(lr){2-5}
 & \textbf{Precision} & \textbf{Recall} & \textbf{F1-Score} & \textbf{Support} \\
\midrule
\vdots & \vdots & \vdots & \vdots & \vdots \\ 
\textbf{Yousef Hobeiche} & \textbf{0.9919 (0.0045)} & \textbf{0.9851 (0.0076)} & \textbf{0.9884 (0.0033)} & \textbf{65.0000 (0.0000)} \\
\textbf{Angele Ellis} & \textbf{0.9111 (0.0244)} & \textbf{0.9444 (0.0824)} & \textbf{0.9222 (0.0434)} & \textbf{12.0000 (0.0000)} \\
\textbf{Yousef Hobeiche \& Angele Ellis} & \textbf{1.0000 (0.0000)} & \textbf{0.6667 (0.2887)} & \textbf{0.7778 (0.1925)} & \textbf{2.0000 (0.0000)} \\
\vdots & \vdots & \vdots & \vdots & \vdots \\ 
Elias Abdallah & 0.9167 (0.1443) & 0.7778 (0.1925) & 0.8190 (0.0330) & 3.0000 (0.0000) \\
Youssef Bey Al-Helou & 0.8889 (0.1925) & 0.6667 (0.2887) & 0.7111 (0.0770) & 2.0000 (0.0000) \\
Father Elias & 0.3333 (0.5774) & 0.3333 (0.5774) & 0.3333 (0.5774) & 1.0000 (0.0000) \\
Magistrate of Batroun & 1.0000 (0.0000) & 0.6667 (0.2887) & 0.7778 (0.1925) & 2.0000 (0.0000) \\
\vdots & \vdots & \vdots & \vdots & \vdots \\ 
\hline
accuracy & 0.9686 (0.0041) & 0.9686 (0.0041) & 0.9686 (0.0041) & 0.9686 (0.0041) \\
macro avg & 0.9251 (0.0297) & 0.8920 (0.0364) & 0.9013 (0.0327) & 2849.0000 (0.0000) \\
weighted avg & 0.9683 (0.0044) & 0.9686 (0.0041) & 0.9662 (0.0049) & 2849.0000 (0.0000) \\
\bottomrule
\end{tabular*}
\end{table}


\begin{table}[htbp]
\caption{Classification Report for DenseNet201 + From Scratch + No Attention (Protocol A)}
\label{tab:app:crA-09}
\footnotesize
\renewcommand{\arraystretch}{1.3} 
\centering
\setlength{\tabcolsep}{5pt} 
\begin{tabular*}{\columnwidth}{@{\extracolsep{\fill}}lcccc} 
\toprule 
\textbf{Class Name} & \multicolumn{4}{c}{\textbf{Mean (Standard Deviation)}} \\
\cmidrule(lr){2-5}
 & \textbf{Precision} & \textbf{Recall} & \textbf{F1-Score} & \textbf{Support} \\
\midrule 
\vdots & \vdots & \vdots & \vdots & \vdots \\ 
\textbf{Yousef Hobeiche} & \textbf{0.9850 (0.0149)} & \textbf{0.9949 (0.0089)} & \textbf{0.9898 (0.0043)} & \textbf{65.0000 (0.0000)} \\
\textbf{Angele Ellis} & \textbf{0.8990 (0.0364)} & \textbf{0.9722 (0.0481)} & \textbf{0.9332 (0.0234)} & \textbf{12.0000 (0.0000)} \\
\textbf{Yousef Hobeiche \& Angele Ellis} & \textbf{0.6667 (0.5774)} & \textbf{0.5000 (0.5000)} & \textbf{0.5556 (0.5092)} & \textbf{2.0000 (0.0000)} \\
\vdots & \vdots & \vdots & \vdots & \vdots \\ 
Elias Abdallah & 1.0000 (0.0000) & 0.6667 (0.3333) & 0.7667 (0.2517) & 3.0000 (0.0000) \\
Youssef Bey Al-Helou & 0.6667 (0.5774) & 0.5000 (0.5000) & 0.5556 (0.5092) & 2.0000 (0.0000) \\
Father Elias & 0.3333 (0.5774) & 0.3333 (0.5774) & 0.3333 (0.5774) & 1.0000 (0.0000) \\
Magistrate of Batroun & 1.0000 (0.0000) & 0.6667 (0.2887) & 0.7778 (0.1925) & 2.0000 (0.0000) \\
\vdots & \vdots & \vdots & \vdots & \vdots \\ 
\hline
accuracy & 0.9711 (0.0016) & 0.9711 (0.0016) & 0.9711 (0.0016) & 0.9711 (0.0016) \\
macro avg & 0.9242 (0.0147) & 0.9042 (0.0149) & 0.9052 (0.0152) & 2849.0000 (0.0000) \\
weighted avg & 0.9702 (0.0033) & 0.9711 (0.0016) & 0.9687 (0.0026) & 2849.0000 (0.0000) \\
\bottomrule 
\end{tabular*}
\end{table}

\FloatBarrier

\begin{table}[htbp]
\caption{Classification Report for DenseNet201 + From Scratch + Attention (Protocol A)}
\label{tab:app:crA-10}
\footnotesize
\renewcommand{\arraystretch}{1.3} 
\centering
\setlength{\tabcolsep}{5pt} 
\begin{tabular*}{\columnwidth}{@{\extracolsep{\fill}}lcccc} 
\toprule 
\textbf{Class Name} & \multicolumn{4}{c}{\textbf{Mean (Standard Deviation)}} \\
\cmidrule(lr){2-5}
 & \textbf{Precision} & \textbf{Recall} & \textbf{F1-Score} & \textbf{Support} \\
\midrule 
\vdots & \vdots & \vdots & \vdots & \vdots \\ 
\textbf{Yousef Hobeiche} & \textbf{0.9949 (0.0087)} & \textbf{0.9897 (0.0089)} & \textbf{0.9923 (0.0001)} & \textbf{65.0000 (0.0000)} \\
\textbf{Angele Ellis} & \textbf{0.9209 (0.0037)} & \textbf{0.9722 (0.0481)} & \textbf{0.9456 (0.0250)} & \textbf{12.0000 (0.0000)} \\
\textbf{Yousef Hobeiche \& Angele Ellis} & \textbf{1.0000 (0.0000)} & \textbf{0.6667 (0.2887)} & \textbf{0.7778 (0.1925)} & \textbf{2.0000 (0.0000)} \\
\vdots & \vdots & \vdots & \vdots & \vdots \\ 
Elias Abdallah & 0.9167 (0.1443) & 0.7778 (0.1925) & 0.8190 (0.0330) & 3.0000 (0.0000) \\
Youssef Bey Al-Helou & 0.8889 (0.1925) & 0.6667 (0.2887) & 0.7111 (0.0770) & 2.0000 (0.0000) \\
Father Elias & 0.3333 (0.5774) & 0.3333 (0.5774) & 0.3333 (0.5774) & 1.0000 (0.0000) \\
Magistrate of Batroun & 1.0000 (0.0000) & 0.6667 (0.2887) & 0.7778 (0.1925) & 2.0000 (0.0000) \\
\vdots & \vdots & \vdots & \vdots & \vdots \\ 
\hline
accuracy & 0.9674 (0.0046) & 0.9674 (0.0046) & 0.9674 (0.0046) & 0.9674 (0.0046) \\
macro avg & 0.9298 (0.0153) & 0.9030 (0.0283) & 0.9060 (0.0240) & 2849.0000 (0.0000) \\
weighted avg & 0.9679 (0.0049) & 0.9674 (0.0046) & 0.9653 (0.0052) & 2849.0000 (0.0000) \\
\bottomrule 
\end{tabular*}
\end{table}

\clearpage

\begin{table}[htbp]
\caption{Classification Report for \MNVL + Fine-tuned + Attention (Protocol B)}
\label{tab:app:crB-mbv3}
\footnotesize
\renewcommand{\arraystretch}{1.3} 
\centering
\setlength{\tabcolsep}{5pt} 
\begin{tabular*}{\columnwidth}{@{\extracolsep{\fill}}lcccc} 
\toprule 
\textbf{Class Name} & \multicolumn{4}{c}{\textbf{Mean (Standard Deviation)}} \\
\cmidrule(lr){2-5}
 & \textbf{Precision} & \textbf{Recall} & \textbf{F1-Score} & \textbf{Support} \\
\midrule 
\vdots & \vdots & \vdots & \vdots & \vdots \\ 

Emily Nasrallah & 1.0000 (0.0000) & 1.0000 (0.0000) & 1.0000 (0.0000) & 11.0000 (0.0000) \\
Jean Philipp Kmeid & 0.9982 (0.0025) & 1.0000 (0.0000) & 0.9991 (0.0013) & 27.0000 (0.0000) \\
Hanna Moussa & 1.0000 (0.0000) & 0.9978 (0.0031) & 0.9989 (0.0016) & 25.0000 (0.0000) \\
Salah Tizani & 0.9876 (0.0131) & 1.0000 (0.0000) & 0.9937 (0.0067) & 297.0000 (0.0000) \\
Hanna Ghayth (Juan Gait) & 0.8343 (0.0182) & 1.0000 (0.0000) & 0.9096 (0.0108) & 13.0000 (0.0000) \\

\vdots & \vdots & \vdots & \vdots & \vdots \\ 

Botros Feghaly & 0.9583 (0.0589) & 0.9702 (0.0212) & 0.9626 (0.0207) & 24.0000 (0.0000) \\
Elie Maroun Khalil & 0.9247 (0.0140) & 1.0000 (0.0000) & 0.9608 (0.0075) & 22.0000 (0.0000) \\
Murshid Habashi & 0.8674 (0.1030) & 0.9720 (0.0200) & 0.9139 (0.0639) & 26.0000 (0.0000) \\

\vdots & \vdots & \vdots & \vdots & \vdots \\

Father Botros Hasan & 0.2428 (0.1358) & 0.6667 (0.4423) & 0.3416 (0.2150) & 19.0000 (0.0000) \\
Father Youssef Hanna \& Father Botros Hasan & 0.2529 (0.3576) & 0.3188 (0.4509) & 0.2821 (0.3989) & 23.0000 (0.0000) \\

\vdots & \vdots & \vdots & \vdots & \vdots \\

May Ziadeh & 0.2480 (0.1763) & 0.6515 (0.4611) & 0.3591 (0.2549) & 22.0000 (0.0000) \\

\vdots & \vdots & \vdots & \vdots & \vdots \\

Prof.\ Ign.\ Kratchkovsky & 0.0000 (0.0000) & 0.0000 (0.0000) & 0.0000 (0.0000) & 13.0000 (0.0000) \\
Asaad Koury Tobia Attallah & 0.0000 (0.0000) & 0.0000 (0.0000) & 0.0000 (0.0000) & 10.0000 (0.0000) \\
Amin Farhat & 0.0000 (0.0000) & 0.0000 (0.0000) & 0.0000 (0.0000) & 21.0000 (0.0000) \\

\vdots & \vdots & \vdots & \vdots & \vdots \\ 
\hline
accuracy & 0.7707 (0.0111) & 0.7707 (0.0111) & 0.7707 (0.0111) & 0.7788 (0.0000) \\
macro avg & 0.6460 (0.0065) & 0.6865 (0.0135) & 0.6317 (0.0141) & 2473.0000 (0.0000) \\
weighted avg & 0.7712 (0.0180) & 0.7707 (0.0111) & 0.7468 (0.0136) & 2473.0000 (0.0000) \\
\bottomrule 
\end{tabular*}
\end{table}

\end{document}